\documentclass[journal]{IEEEtran}
\usepackage[colorlinks=true,linkcolor=blue]{hyperref}
\hypersetup{
	colorlinks,
	citecolor=blue,
	filecolor=blue,
	linkcolor=blue,
	urlcolor=blue
}
\usepackage{graphicx}
\usepackage[utf8]{inputenc}
\usepackage{mathtools}
\usepackage{supertabular} 
\usepackage{array}     
\usepackage{float}
\usepackage{lscape}    
\usepackage{rotating}    

\usepackage{tabularx}
\usepackage{booktabs} 

\usepackage{amsmath}
\usepackage{amssymb} 
\usepackage{textcomp}
\usepackage{cite}
\usepackage{multirow}
\usepackage{booktabs}
\usepackage{tabularx}
\usepackage{adjustbox}
\usepackage{algorithmic}
\usepackage{makecell}
\usepackage[ruled,vlined]{algorithm2e}
\graphicspath{{images/},{img/}}
\DeclareGraphicsExtensions{.pdf,.jpeg,.png,.jpg,.png}
\def\BibTeX{{\rm B\kern-.05em{\sc i\kern-.025em b}\kern-.08em
		T\kern-.1667em\lower.7ex\hbox{E}\kern-.125emX}}
\begin{document}
	
	\title{AI-Driven HSI: Multimodality, Fusion, Challenges, and the Deep Learning Revolution}
	\author{David~S.~Bhatti,~\IEEEmembership{}
		Yougin~Choi,~\IEEEmembership{}
		Rahman~S~M~Wahidur,~\IEEEmembership{}
		Maleeka~Bakhtawar,~\IEEEmembership{}
		Sumin~Kim,~\IEEEmembership{}
		Surin Lee,~\IEEEmembership{}
		Yongtae Lee,~\IEEEmembership{}
		and~Heung-No~Lee,~\IEEEmembership{Senior~Member,~IEEE}
		\thanks{D. S. Bhatti, R. S. M. Wahidur, M. Bakhtawar, Y. Lee,  and H.-N. Lee are with the School of Electrical Engineering and Computer Science, Gwangju Institute of Science and Technology (GIST), Gwangju 61005, South Korea (e-mail:  heungno@gist.ac.kr, david.bhatti@seecs.edu.pk;).} 
					\thanks{S. Kim, and Y. Choi are with the Artificial Intelligence Graduate School, GIST, Gwangju 61005, South Korea.}
		\thanks{All authors are members of the INFONET research lab at GIST (Korea), specializing in the convergence of cutting-edge technologies, including HSI, Blockchain, and AI.\url{https://heungno.net/}}
		\thanks{Manuscript received xxxx xx, 2025; revised xxxx xx, 2025.}}
	
	\markboth{Bhatti \MakeLowercase{\textit{et. al.}}: AI-Driven HSI: Multimodality, Fusion, Challenges, and the Deep Learning Revolution}
	{Bhatti \MakeLowercase{\textit{et. al.}}: AI-Driven HSI: Multimodality, Fusion, Challenges, and the Deep Learning Revolution}
	
	
	
	\maketitle
	
	
	\begin{abstract}
	Hyperspectral Imaging (HSI) is a cutting-edge technology that captures comprehensive spatial and spectral data, enabling the analysis of features invisible to conventional imaging systems. This capability supports in-depth investigations of object composition, condition, and transformation, making HSI a critical tool across a wide range of applications, including weather monitoring, food quality control, counterfeit detection, healthcare diagnostics, face anti-spoofing, and extending into fields such as defense, agriculture, and industrial automation. HSI has evolved significantly due to advancements in spectral resolution, device miniaturization, and computational methods, broadening its application scope. This study offers a comprehensive overview of HSI, detailing its fundamental principles, diverse applications, challenges in data fusion, and the pivotal role of deep learning models in HSI processing. We explore how the integration of multimodal HSI with advanced AI models, particularly deep learning frameworks, has substantially improved classification accuracy and operational efficiency. Furthermore, we discuss how deep learning has enhanced multiple aspects of HSI analysis, such as feature extraction and reduction, target and change detection, denoising, unmixing, spatial-spectral attention, dimensionality reduction, land cover mapping, data augmentation, domain adaptation, spectral reconstruction, compression, and super-resolution. An emerging area of focus is the innovative fusion of hyperspectral cameras with large language models (LLMs), referred to as the "high-brain LLM." This integration enables advanced applications, including low-visibility crash detection and face anti-spoofing, while empowering LLMs to generate actionable human-oriented alerts and insights. Additionally, we highlight key players in the HSI industry, its Compound Annual Growth Rate (CAGR), and the rapid expansion and industrial significance of the technology. This work aims to serve both technical and non-technical audiences, offering valuable insights for multidisciplinary researchers and industry practitioners. By addressing open research challenges, emerging trends, and the integration of HSI with deep learning and LLMs, this study provides a thorough understanding of HSI’s current landscape and promising future directions. Presented in a tutorial format, this study equips readers to apply these insights to their respective fields, while also providing information on HSI datasets and software libraries, allowing researchers to effectively engage with this domain.
	\end{abstract}
	
	
	
	\section{Introduction}\label{Sec:Introduction}
	
	HSI captures detailed spatial and spectral data, enabling the analysis of features that traditional imaging systems cannot detect. This makes HSI invaluable across diverse fields such as weather monitoring, crop protection, food quality control, and counterterrorism. Stored as a three-dimensional hypercube, HSI combines spatial and spectral dimensions (height, width, wavelength), allowing in-depth analysis of materials, composition, and transformations. Since its inception in the 1960s, HSI has evolved from multispectral sensors in remote sensing to advanced applications across environmental monitoring, healthcare, defense, agriculture, and more. The term "hyperspectral" was first defined in 1983 as involving over 100 spectral bands, enabling material identification \cite{Lee2019HyperspectralImaging}. However, HSI still faces challenges such as spectral unmixing and data fusion, requiring sophisticated approaches. The integration of AI, Machine Learning (ML), and Deep Learning (DL) has transformed HSI analysis, significantly improving object detection and classification accuracy. The combination of multimodal HSI and advanced AI models is driving the development of more effective systems for defense, technology, and societal applications.
	
\begin{figure*}[!htbp]
		\centering
		\includegraphics[width=0.65\linewidth, keepaspectratio]{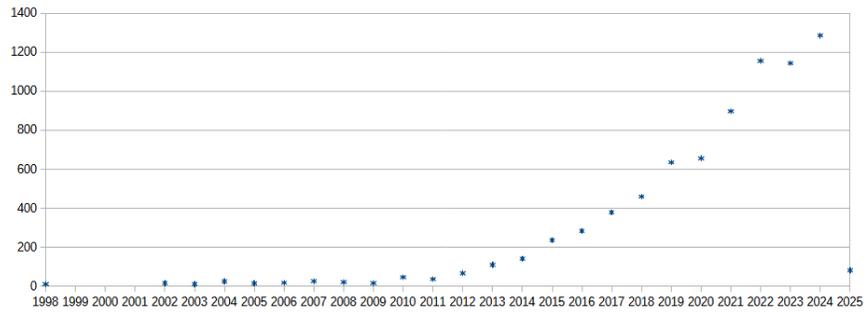}
		\caption{HSI Yealy Publications (1998-2025)}
		\label{fig:HSIPubYearWise}
	\end{figure*}

\begin{figure*}[!htbp]
		\centering
		\includegraphics[width=0.75\linewidth, keepaspectratio]{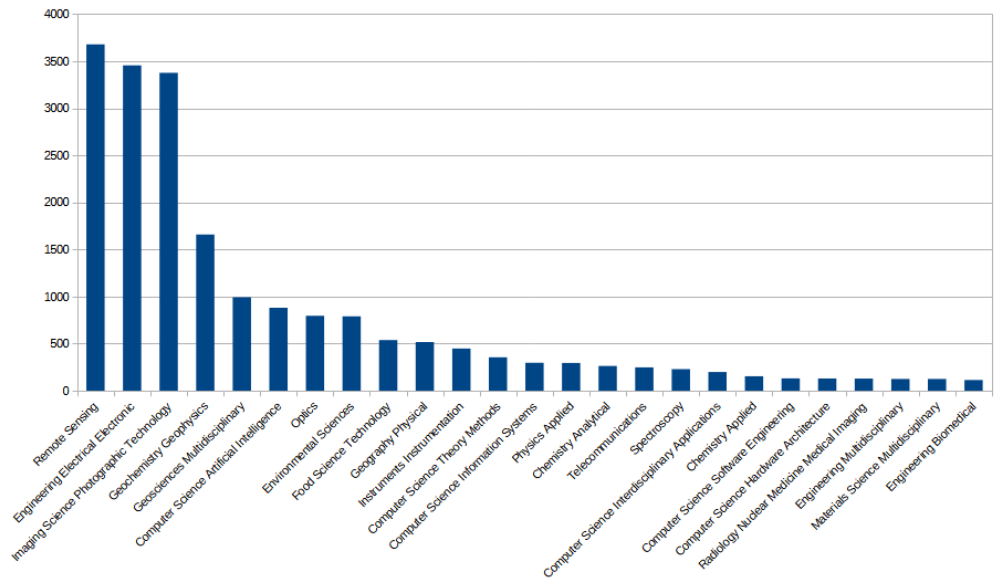}
		\caption{HSI Publication Based on Different WOS Categories (1998-2025)}
		\label{fig:HSIPubCatWise}
	\end{figure*}

\begin{figure*}[!htbp]
		\centering
		\includegraphics[width=0.9\linewidth, keepaspectratio]{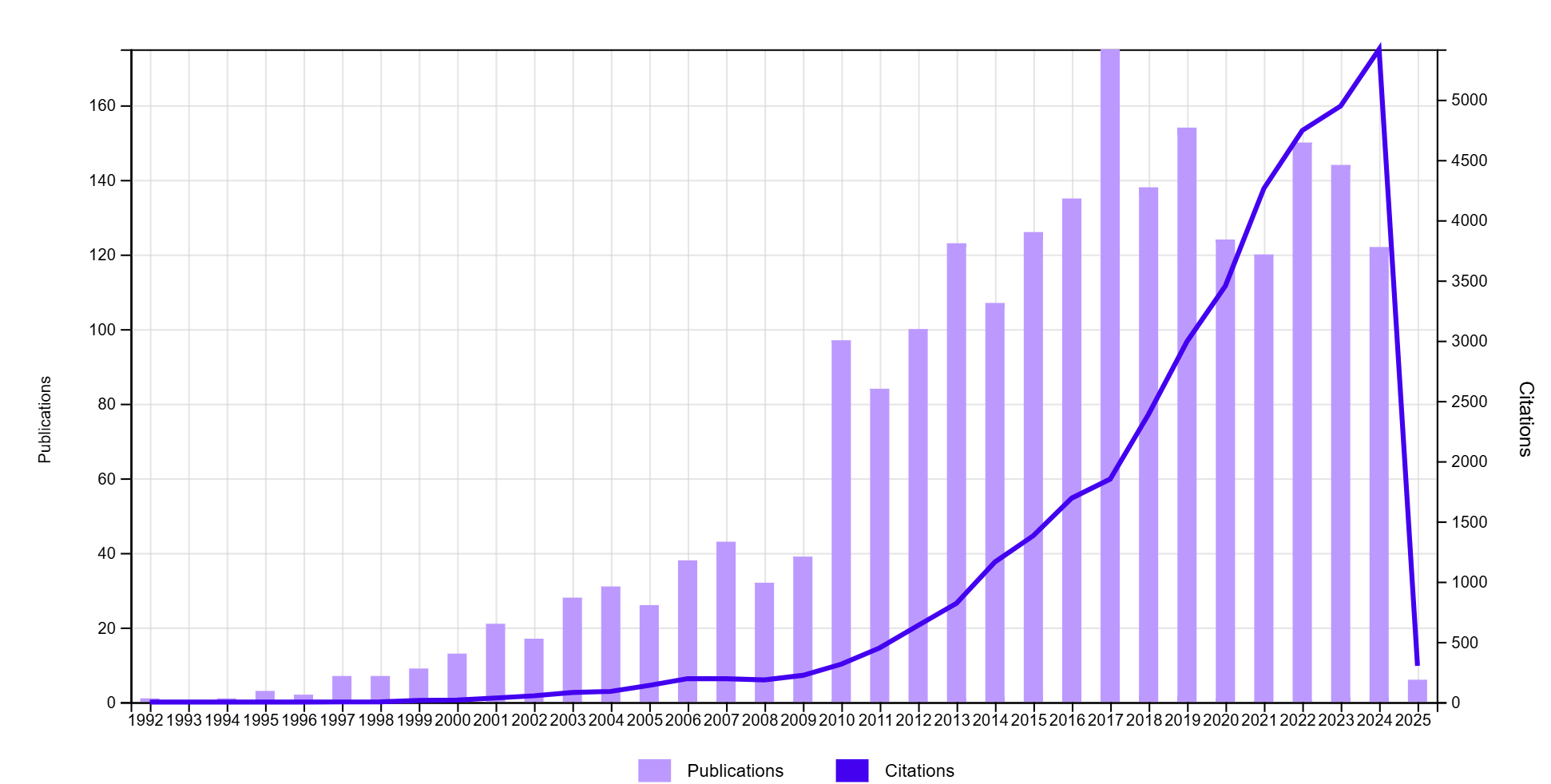}
		\caption{HSI Publication Citations VS Publications (1992-2025)}
		\label{fig:HSIPubAndCite}
	\end{figure*}

	Our analysis of the Web of Science (WOS) database reveals the continued growth of HSI research, reflecting its expanding application range from environmental monitoring to medical diagnostics. Figures \ref{fig:HSIPubYearWise}, \ref{fig:HSIPubCatWise}, and \ref{fig:HSIPubAndCite} highlight this trend, demonstrating HSI's increasing significance in both scientific and industrial domains. 
	
 We have observed that most surveys on HSI focus on specific areas and do not provide a comprehensive overview, leaving readers to search for multiple articles on the same topic. For example, Zhang et al. \cite{HSI:Survey:Zhang:2022} only address image reconstruction from RGB (Red, Green, Blue) to HSI and its associated challenges. Sethi et al. \cite{HSI:Survey:Zhang:Sethy:2021} focus solely on HSI applications in agriculture, with little attention paid to the outdated discussion on HSI's Compound Annual Growth Rate (CAGR). Khan et al. \cite{2018:HSI-ImageAnalysis:Review:IEEEAccess} and Bhargava et al. \cite{HSIAndApplications;BHARGAVA:2024:ElsvrJ} provide detailed reviews on HSI applications. Gu et al. \cite{gu2021multimodal} focus exclusively on multimodal data acquisition, processing, and challenges. Kumar et al. \cite{HSI:DBN:Class:Kumar:2024} provide an excellent survey on deep learning for HSI classification, critically discussing the pros and cons of various studies and the challenges of small data samples. However, this survey is limited to deep learning applications in HSI classification alone. Other reviews include \cite{HSIImageRestoration:Liu:2023}, which highlights image restoration methods; \cite{HSIinmedical:Karim:2022}, which reviews trends in HSI for medical imaging; \cite{HSIforNanoscale:Dong:2019}, which explores HSI for nanoscale materials research; \cite{HSI:AttentionBased:Classification:Vats:2023}, which reviews attention-based HSI classification; and \cite{HSITextureWavelengthSelection:Rogers:2023}, which reviews methods for texture and wavelength feature selection. However, these studies cover only a small fraction of the overall HSI landscape and lack a tutorial format.

Our study overlaps with Kumar et al. \cite{HSI:DBN:Class:Kumar:2024} but differs significantly in terms of the applications of deep learning in HSI. While Kumar et al. focus exclusively on classification, we provide insights into which deep learning models are suitable for various HSI tasks, such as feature extraction and reduction, target and change detection, denoising, unmixing, spatial-spectral attention, dimensionality reduction, land cover mapping, data augmentation, domain adaptation, spectral reconstruction, compression, and super-resolution. Our emphasis on HSI multimodality, DL-based fusion of HSI data, recent challenges, potential solutions, trends, CAGR, and a tutorial approach to image pre-processing, DL-based image reconstruction, feature extraction, and classification offers a more comprehensive perspective than Kumar et al. \cite{HSI:DBN:Class:Kumar:2024}.

Based on a thorough review of existing surveys and reviews, we are confident that this write-up is more comprehensive, well-composed, and presented in a tutorial format. It effectively serves both technical and non-technical readers, catering to multidisciplinary researchers and those keen to explore new research dimensions through the integration of HSI, deep learning, and LLMs.

	
	\subsection{Motivation}
	Our motivation for conducting this article is multifaceted:
	\begin{enumerate}
		\item To analyze the role of HSI in detecting and analyzing objects not detectable by conventional imaging systems.
		\item To explore how deep learning can enhance HSI’s capabilities, addressing its challenges more efficiently and with greater accuracy.
		\item To investigate the opportunities of merging LLMs with hyperspectral cameras for generating human-language-oriented prompts and alerts.
		\item To examine the growth of the HSI industry, including a study of the CAGR predicted by leading business analysis and statistics organizations.
		\item To complete a manuscript that may serve the purpose of learning for new researchers in DL-based HSI.
	\end{enumerate}
	
	The manuscript is structured into the following sections:\newline
	\ref{sec:researchmethod}-\textbf{Research Methodology}: This section details the research methodology followed by the researchers in this study.\newline
	\ref{sec:UnderstandingHSI}-\textbf{Understanding HSI}: This section provides a comprehensive background and understanding of HSI.\newline
	\ref{sec:HSIPreprocessing}-\textbf{HSI Pre-Processing}: Building on the previous section, this part discusses preprocessing techniques essential for computational analysis in HSI.\newline
	\ref{sec:AugmentingHSIWithDeepLearning}-\textbf{Augmenting HSI with Deep Learning}: Various deep learning models, including CNNs (Convolutional Neural Network), DBNs (Deep Belief Networks), RNNs (Recurrent Neural Networks), LSTMs (Long Short-Term Memory), GRUs (Gated Recurrent Units), GANs (GANs), transformers, autoencoders, and stacked autoencoders, are explored for their applications in HSI tasks such as feature extraction, dimensionality reduction, classification, object and change detection, data fusion, augmentation, denoising, and spectral unmixing.\newline
	\ref{sec:MultimodalHSIFusionChallengesLLMs}-\textbf{Multimodal HSI, Fusion, and Challenges}: This section introduces Multimodal HSI, data fusion challenges, and the role of LLMs in HSI applications, presenting new research dimensions.\newline
	\ref{sec:CAGR}-\textbf{CAGR}: This section examines the annual growth rate of the HSI industry.\newline
	\ref{sec:HSIApplications}-\textbf{Applications of HSI}: A broad range of HSI applications in fields such as healthcare, environmental monitoring, defense, forensics, and microscopy are discussed.\newline
	\ref{sec:ChallengesandTrends}-\textbf{Challenges and Recent Trends}: Key challenges and their solutions in HSI and emerging trends shaping the field are highlighted.\newline
	\ref{sec:ConclusionFuturework}-\textbf{Conclusion and Future Work}: The study concludes with a summary of findings and directions for future research in HSI.
	\section {Research Methodology}\label{sec:researchmethod}
	For writing this article, the guidelines laid down by PRISMA \cite{Pagen160:DefsElabs:PRISMA:2021} were followed, with significant help from \cite{DukeLibrary:SysReview:2024}. The PRISMA guidelines, set by medical and health science experts, are also applicable for systematic literature surveys, reviews, and tutorials in other domains. A summary of the key steps followed in this project is briefly discussed below.\newline  
	\textbf{Step-1 Conception \& Research Questions:} Driven by AI growth and HSI's potential to address social, industrial, and technological challenges, the idea was conceived by the researchers working on HSI, AI and LLMs to explore and review true potential of AI-driven HSI. Then to effectively address it, following questions were formulated, which this study can address.
	\begin{enumerate}  
		\item What is HSI, HSI image, signature, HSI Pixel, and hypercube?  
		\item How do single-modal and multimodal HSI systems differ in data acquisition, fusion, challenges, and AI tools for processing HSI images?  
		\item What is the role of AI in enhancing HSI systems, particularly in learning, classification, and challenges?  
		\item What is the CAGR of the HSI industry, and what factors contribute to its growth?  
		\item What are the current challenges, open research issues, and emerging trends in HSI technology.
	\end{enumerate}  
	\textbf{Step-2 Research Team Formation:} A team of researchers specializing in AI, HSI, and communication was formed.\newline  
	\textbf{Step-3 Research Method (articles and journals selection):} The team decided on a unified approach and specific criteria for selecting articles and journals. The keywords used in the search included HSI (Hyperspectral Imaging), HSI basics, signature, hypercube, spectral unmixing, acquisition methods, electromagnetic spectrum, preprocessing, multimodal HSI, data fusion, LLMs, AI, machine learning, deep learning, CNNs, DBNs, GANs, RNNs, transformers in HSI, HSI CAGR, key players, and trends in HSI. Only articles from 2015-2025 were considered, with a few exceptions. Journals from publishers like IEEE (Institute of Electrical and Electronics
	Engineers), ACM (Association for Computing Machin
	ery), Springer, Elsevier, Wiley-Blackwell, and Taylor \& Francis were included. Articles from databases like Web of Science, Scopus, and SpringerLink were also considered. Only articles with an impact factor greater than zero were included.\newline  
	\textbf{Step-4 Documentation \& Reporting:} Selected articles were reviewed and their findings documented in paragraphs, images, graphs, and tables by the research team. Research gaps, challenges, and unaddressed issues were highlighted, and the CAGR of HSI in the international market was reported.\newline  
	\textbf{Step-5 Repetitive Review, Updating, and Submission:} This study was continuously reviewed through periodic meetings. The final report, after revisions, was submitted for publication.  

		\section{Understanding Hyperspectral Imaging}\label{sec:UnderstandingHSI}
\begin{figure}[!htbp]
			\centering
			\includegraphics[width=0.70\linewidth, keepaspectratio]{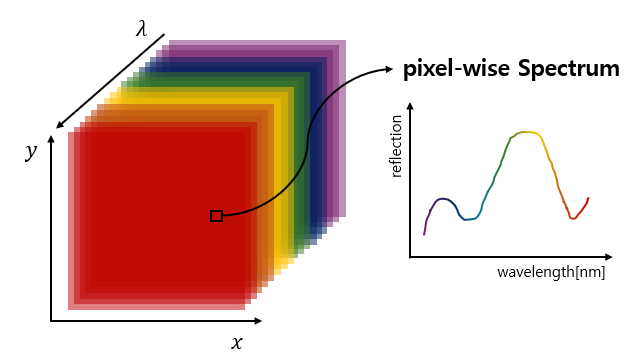}
			\caption{Spectral Information}
			\label{fig:SpectralInformation}
		\end{figure}

		HSI generates high-resolution, multi-spectral images by capturing a continuum of spectral lines across a wide range of wavelengths. Unlike multispectral sensors, which capture images in broad wavelength bands, hyperspectral sensors collect data across dozens to hundreds of narrow, contiguous spectral bands, enabling the extraction of a continuous spectrum for each pixel, as depicted in Figure \ref{fig:SpectralInformation}.  
		
\begin{figure}[!htbp]
			\centering
			\includegraphics[width=0.9\linewidth, keepaspectratio]{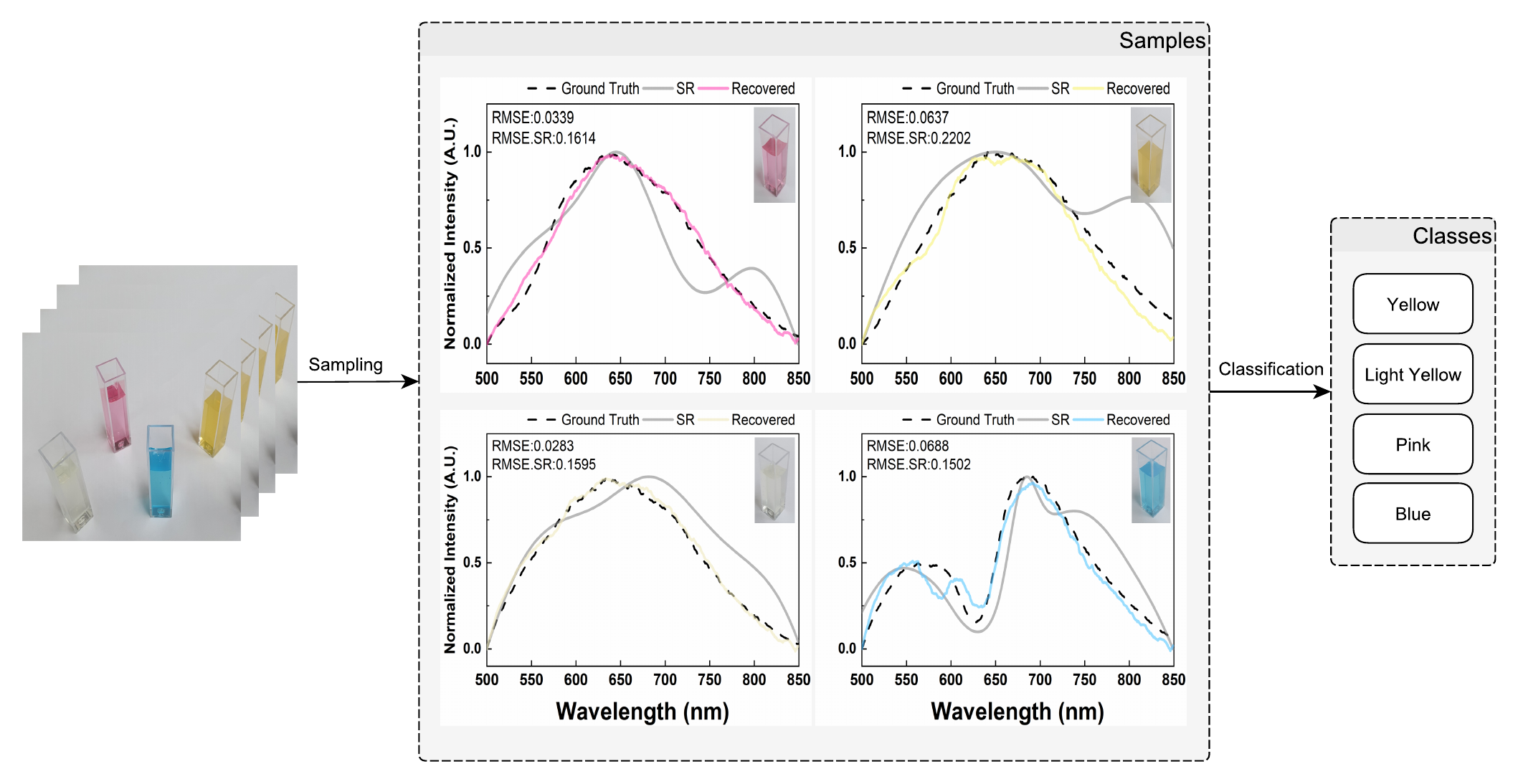}
			\caption{HSI Cube, Sampling, and Signature}
			\label{fig:HSICube}
		\end{figure}

		Hyperspectral cameras capture contiguous wavelengths across visible, near-infrared (NIR), short-wave infrared (SWIR), and mid-wave infrared (MIR) spectra, achieving spectral resolutions as fine as a nanometer. Light interacting with the sensor is dispersed into numerous spectral bands, forming a \textbf{hypercube}, which is a three-dimensional dataset with two spatial and one spectral dimension, as shown in Figure \ref{fig:HSICube} and Figure \ref{fig:SpectralInformation}. This allows for the identification of materials based on their \textbf{spectral signatures}, such as vegetation, soil, and minerals. Slight variations in these signatures help differentiate plant species.  After applying corrections for sensor characteristics, atmospheric conditions, and terrain effects, extracted spectra can be compared with field or laboratory data for material mapping. Hyperspectral imaging employs different acquisition methods: spatial scanners capture spectral data over time, while snapshot imagers use a focal plane array for instantaneous spectral fingerprints. Effective interpretation of HSI data requires an understanding of material properties and their spectral responses \cite{2022:HSI:TraditionalToDeepModels:Review:IEEIJ, 2018:HSI-ImageAnalysis:Review:IEEEAccess}.	
		\subsection{Electromagnetic Spectrum: Information in HSI Image }
		
\begin{figure*}[!htbp]
			\centering
			\includegraphics[width=0.75\linewidth, keepaspectratio]{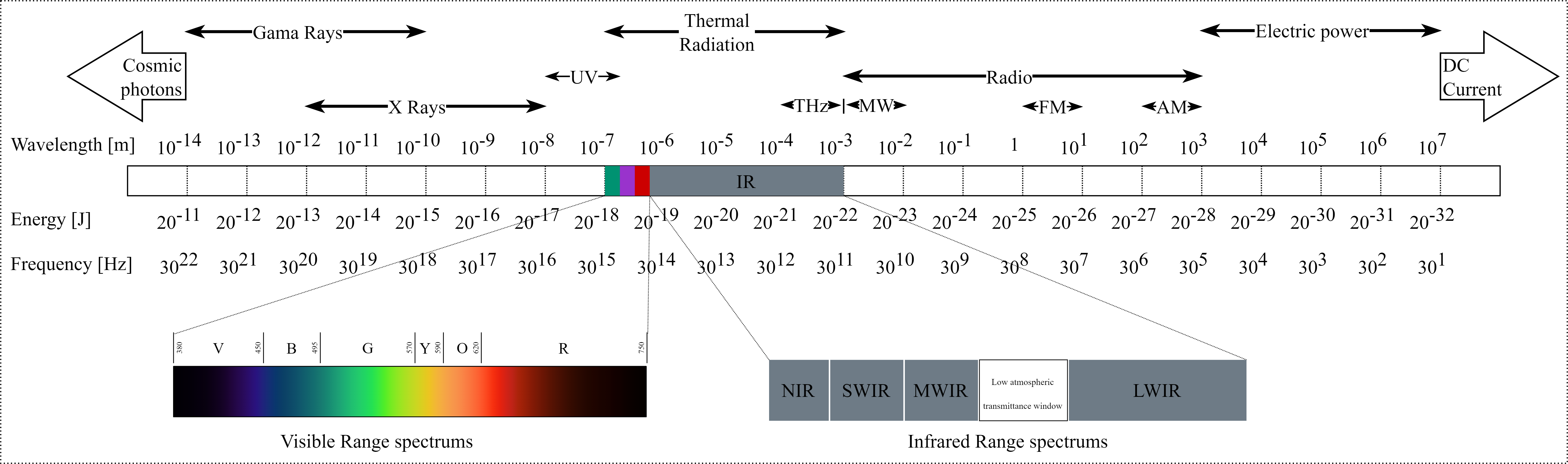}
			\caption{ElectromagneticSpectrumBandsandTheirWavelengths}
			\label{fig:ElectromagneticSpectrum}
		\end{figure*}

\begin{figure*}[!htbp]
			\centering
			\includegraphics[width=\linewidth, keepaspectratio]{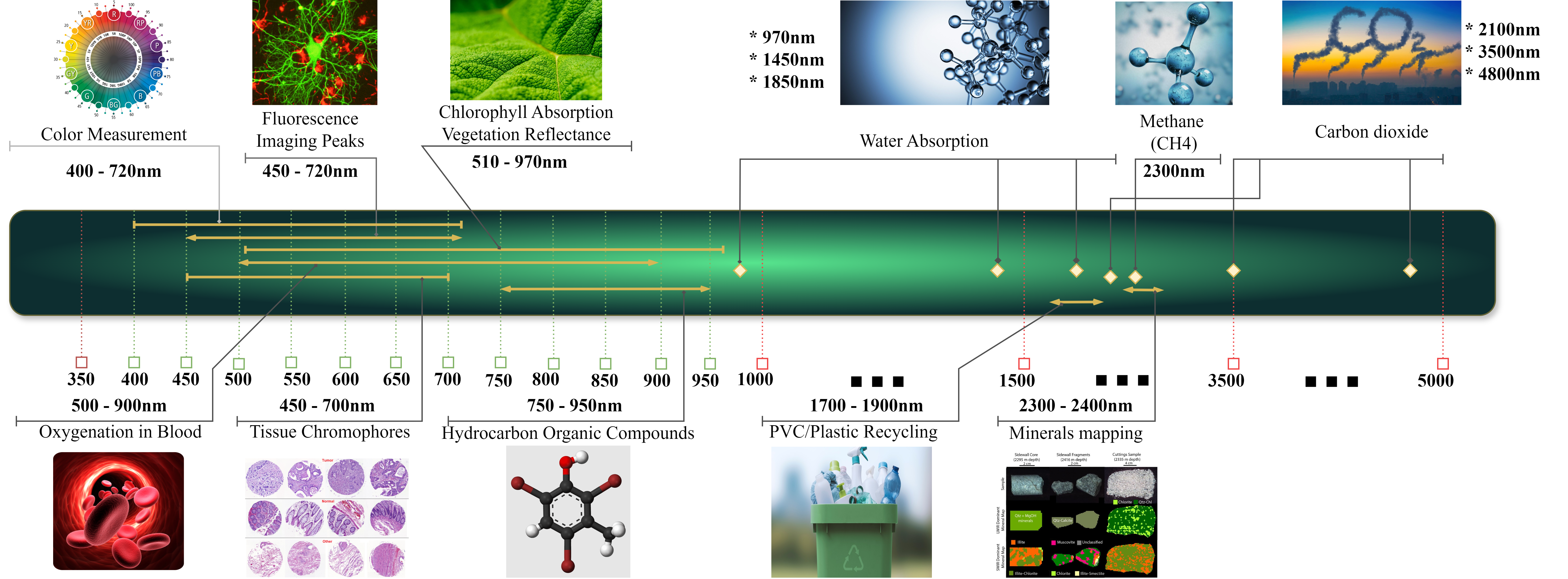}
			\caption{HSIInformationinBands}
			\label{fig:HSIBands.png}
		\end{figure*}

		Electromagnetic waves travel through space via oscillating electric and magnetic fields and can propagate through air, solids, and vacuum. The electromagnetic spectrum spans from radio waves to gamma rays, with visible light being a small segment. Each wavelength has varying energy levels, affecting how materials absorb or reflect light \cite{fath2018encyclopedia}. In hyperspectral imaging, this principle is crucial. By analyzing different wavelengths, hyperspectral imaging identifies material properties and environmental conditions, allowing precise analysis. Understanding electromagnetic wave interactions helps interpret hyperspectral data, making it valuable for remote sensing, agriculture, and medical diagnostics. The electromagnetic spectrum and HSI data across its bands are shown in Figure \ref{fig:ElectromagneticSpectrum} and Figure \ref{fig:HSIBands.png}.
		
		\subsection{HSI Acquisition Methods}\label{sec:HSIDataAcquisition}
\begin{figure}[!htpb]
			\centering
			\includegraphics[width=0.85\linewidth, keepaspectratio]{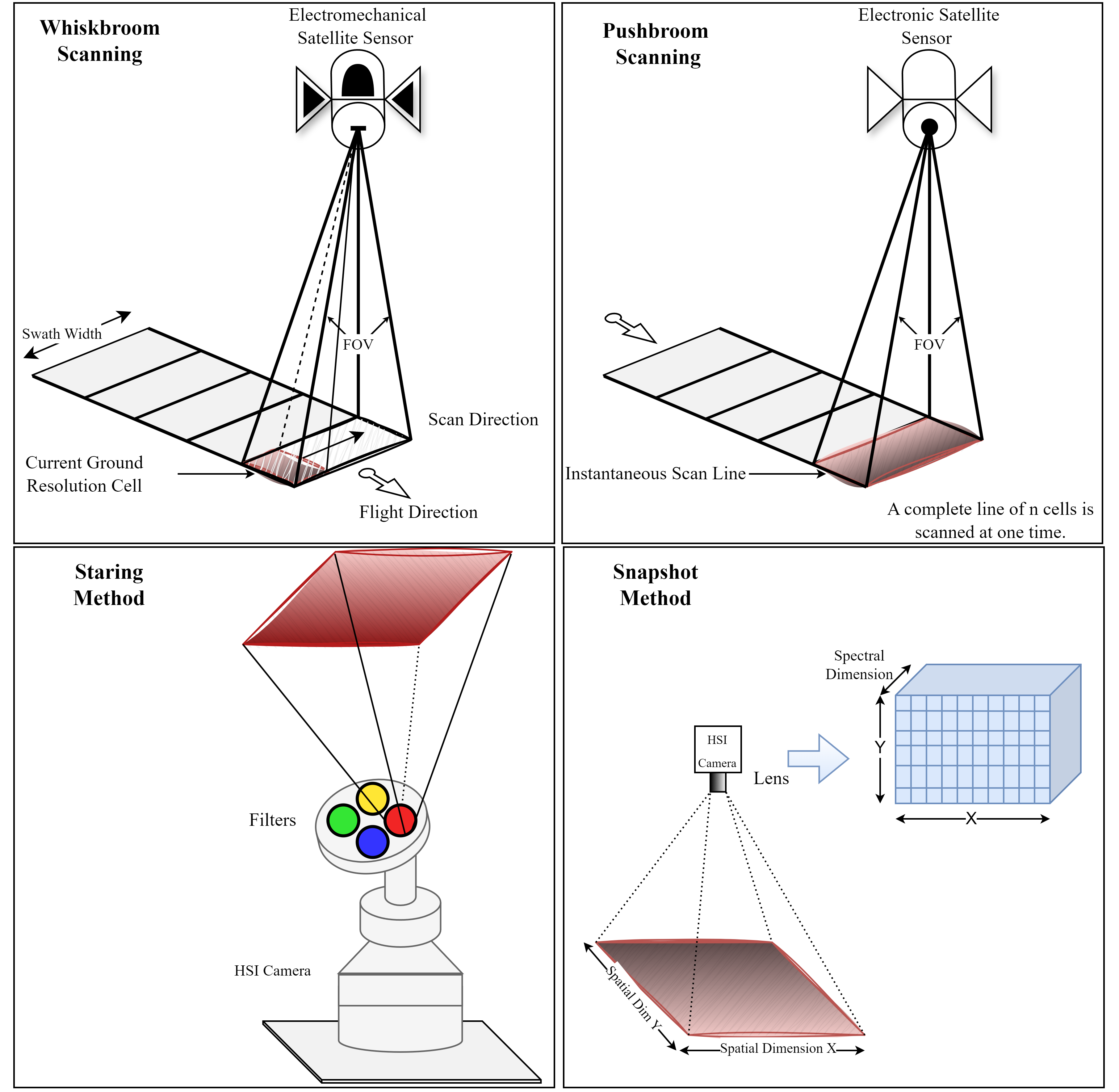}
			\caption{HSI Image Acquisition Methods}
			\label{fig:HSIImageAcquisitionMethodsFull}
		\end{figure}

		In HSI systems, various image acquisition methods offer distinct advantages and trade-offs. The \textbf{whiskbroom} method employs point scanning, providing high-resolution imaging over large areas. It is widely used in satellite imaging and environmental monitoring due to its precision, though its long acquisition time can be a drawback \cite{HSIDataAcquisition:Vishnu:2017}. Similarly, the \textbf{pushbroom} method uses line scanning, offering large coverage with high resolution. It is commonly applied in remote sensing and agriculture, where continuous data collection is essential. However, like whiskbroom scanning, its acquisition time limits its usability in dynamic environments. In contrast to pushbroom, spectral filter based HS imaging uses spectral filters, like absorption or interference filters, to capture specific spectral bands \cite{HSIDataAcquisition:Vishnu:2017}. The \textbf{staring} method relies on wavelength scanning, enabling high spatial resolution within a narrow range. This makes it particularly useful for applications like mineral mapping and surface analysis. Despite its efficiency, its lower spectral resolution and restricted working range may limit broader applications \cite{HSIDataAcquisition:Mary:2019,Tomasarson:Tomasarson:2020}. The \textbf{snapshot} method stands out for its rapid acquisition time and compact size. It is ideal for real-time imaging applications such as medical diagnostics and rapid environmental assessments. However, this method sacrifices both spectral and spatial resolution, making it less suitable for high-detail analysis \cite{HSIDataAcquisition:Vishnu:2017}. Each method balances resolution, speed, and coverage based on application needs. Figure \ref{fig:HSIImageAcquisitionMethodsFull} illustrates whiskbroom, pushbroom, starring and snapshot for better understanding.  
		\subsection{HSI Hypercube} 
		 HSI hypercube is a three-dimensional dataset where two dimensions represent spatial information (height, width), and the third dimension represents spectral information (wavelength), shown in Figure \ref{fig:SpectralInformation} where \(x,y,\lambda\) are representing height, width and wavelength. Spatial resolution defines the smallest ground area each pixel represents. Hyperspectral imaging has lower spatial resolution than multispectral or panchromatic imaging due to the need to capture many spectral bands. Satellite sensors typically range from 10 to 30 meters per pixel (e.g., A:National Aeronautics and Space
		 Administration (NASA) Hyperion, 30m) and airborne sensors range from 1–20m (e.g., CASI:Compact Airborne Spectrographic Imager, AISA:Airborne Imaging Spectrometer for Applications). UAV (Unmanned Aerial Vehicle) sensors offer the highest resolutions (0.01–0.5m), such as Headwall Hyperspec. Despite lower spatial resolution, HSI’s high spectral resolution enhances applications like vegetation monitoring, mineral analysis, and water quality assessment \cite{HSIAndApplications;BHARGAVA:2024:ElsvrJ}.  
		Spectral resolution measures a sensor’s ability to distinguish wavelengths. Higher spectral resolution enables precise material identification. HSI captures many narrow bands across visible, near-infrared, and mid-infrared regions. NASA’s Hyperion provides 10 nm resolution, ESA’s PROBA-CHRIS offers 34 nm (19 bands) and 17 nm (63 bands), while airborne sensors like AVIRIS (10 nm) and AISA (3.3 nm) offer finer detail. UAV sensors, like Headwall Hyperspec (2.5 nm), enhance spectral resolution \cite{HSIAndApplications;BHARGAVA:2024:ElsvrJ}.  
		Temporal resolution refers to how often a sensor captures data from the same location. High temporal resolution allows frequent revisits, while low resolution indicates longer intervals. This is crucial for weather patterns, environmental monitoring, and change detection \cite{HSIAndApplications;BHARGAVA:2024:ElsvrJ}.		
	\subsection{Hyperspectral Signature: HSI Pixel}  
	Hyperspectral imaging signatures represent the unique spectral characteristics of materials or objects. HSI captures and processes data across the electromagnetic spectrum, providing spectral information for each pixel. Materials reflect, absorb, or emit electromagnetic radiation differently across wavelengths, creating distinct spectral "fingerprints." These signatures enable the identification and analysis of materials based on their spectral properties. For example, vegetation, water, soil, and minerals each have distinct signatures, and subtle variations in plant species allow for differentiation \cite{ObjectDetect:Zubair:2022}.		
		\section{HSI Image Processing and Pre-Processing}\label{sec:HSIPreprocessing}
		Hyperspectral imaging, also referred to as imaging spectroscopy, emerged from the integration of traditional spectroscopic techniques with modern imaging system technologies. Spectrometry captures specific spectral data that facilitates the identification and analysis of materials, providing insights into their composition and related properties. Spectroscopy, in essence, examines the interaction between light and matter to reveal the intrinsic characteristics of a substance. Conversely, imaging systems are designed to translate the visual representation of an object’s internal structure into digital signals, which are subsequently processed by computational systems to generate a digital image. These systems typically encompass a camera, optical lens, and an illumination source. When combined, spectroscopy and advanced imaging technologies form the foundation of hyperspectral imaging, enabling the precise measurement of spectral content at each pixel of an image \cite{ImagingSpectroscopyZAHRA:2024}.	
		\subsection{Hyperspectral Band Selection}			
Hyperspectral imaging sensors offer exceptional spectral resolution by capturing responses across hundreds of narrow bands, essential for applications like environmental monitoring and target detection. However, the data generated by these sensors, such as NASA’s Hyperspectral Infrared Imager, which produces 65 Mb/s data rate leading to 372 Gb per orbit and 5.2 Tb per day creating computational challenges \cite{2019:HSBandSelection:IEEEJ}. This vast data volume increases processing demands and worsens the "curse of dimensionality," where classification accuracy drops with additional spectral bands. To address this, hyperspectral band selection techniques reduce redundancy and computational costs while preserving essential information. Methods include ranking-based, searching-based, clustering-based, sparsity-based, embedding-learning-based, and hybrid approaches. Ranking-based methods select top K bands but ignore inter-band correlations \cite{kim:2017:covariance}. Searching-based methods optimize a cost function but are computationally intensive \cite{wang:2018:band}. Clustering-based methods, like K-means, ensure noise resilience but may converge on suboptimal selections \cite{yuan:2015:dual}. Sparsity-based approaches reduce data complexity but add performance uncertainty \cite{sun:2015:band}. Embedding-learning methods integrate selection with learning models like SVMs, offering efficiency but requiring complex design \cite{kuo:2013:kernel}. Hybrid methods combine strategies for effective band selection but require careful implementation \cite{wang:2016:salient}. The goal of band selection is to maximize information content, minimize redundancy, and enhance class separability, improving classification accuracy, as shown in Table \ref{tab:band_selection_methods}.
		\begin{table}[!htpb]
			\centering
			\scriptsize
			\caption{Summary of Hyperspectral Band Selection Methods}
			\label{tab:band_selection_methods}
			\setlength{\tabcolsep}{0.0025\textwidth}
			\begin{tabular}{|p{1.15cm}|p{2.5cm}|p{2.5cm}|p{2.20cm}|}
				\hline
				\textbf{Category} & \textbf{Pros} & \textbf{Cons} & \textbf{Examples} \\ \hline
				{Ranking based} & low computational cost, good for large datasets & ignores inter-band correlation & Supervised\cite{kim:2017:covariance}, un-supervised  \cite{feng:2017:hyperspectral}\\ \hline
				{Searching based} & ignores the entire subset of bands and considers individual & requires extensive computation & Incremental searching based \cite{liu:2016:hyperspectral},Updated searching based \cite{wang:2018:band} \\ \hline
				{Clustering based} & simple, less affected by noise, entire subset of bands can be optimized & low in robustness, easily converging on local optimum solutions & K-means based \cite{yuan:2015:dual}, AP based \cite{yang:2017:discriminative}, Graph based \cite{yuan:2016:discovering} \\ \hline
				{Sparsity based} & reduces HSI data complexity in terms of space and time & uncertainty in model processing performance & SNMF based \cite{sun:2015:band}, Sparse representation based \cite{sun:2017:fast}, Sparse regression based \cite{damodaran:2017:sparse} \\ \hline
				{Embedding learning based} & avoids repetitive learning & difficult to construct objective function based upon performance parameters & Classifier learning \cite{kuo:2013:kernel} and Autoencoder  \cite{tschannerl:2018:segmented} based \\ \hline
				{Hybrid-scheme} & good at finding 
				least number of useful bands & complex & Manifold ranking \cite{wang:2016:salient} \\ \hline
			\end{tabular}
		\end{table}
		\subsection{Classification}
		HSI can be categorized by factors such as data acquisition method, wavelength band, and the number of spectral bands, which determines resolution. Hyperspectral imaging is broadly classified based on the acquisition method, either spatial or spectral scanning, each with advantages in how data is captured and processed into a hyperspectral cube, as explained in section \ref{sec:HSIDataAcquisition}. The measurement spectrum band is another key criterion for classifying hyperspectral cameras, which influences their application. Bands are categorized into UV (200–400 nm), VIS (400–600 nm), NIR (700–1,100 nm), SWIR (1.1–2.5 $\mu$m), and MWIR (2.5–7 $\mu$m) \cite{MAGNUSSON19991158,wilson2015review}. For example, UV–SWIR is used in biomedical fields, VIS–SWIR in atmospheric monitoring, and SWIR–MWIR in defense. Hyperspectral imaging can also be classified by the number of spectral bands: systems with 10 or fewer bands are multispectral, those with 10 or more bands are hyperspectral, and systems with 1,000+ bands are ultra-hyperspectral. Typical spectral resolution for hyperspectral imaging is $\Delta \lambda / \lambda \approx 0.01$. Lower resolution classifies as multispectral ($\Delta \lambda / \lambda \approx 0.1$), and higher as ultra-hyperspectral ($\Delta \lambda / \lambda > 0.001$). Applications vary by classification: multispectral is for macroscopic object classification, hyperspectral for chemical composition analysis in solids or liquids, and ultra-hyperspectral for gases \cite{Lee2019HyperspectralImaging}. A summarized view of this classification is given in Table \ref{table:HSIClassification}.
		
		\begin{table}[!htpb]
			\centering
			\scriptsize
			\caption{Spectroscopic Techniques and their Classification}
			\label{table:HSIClassification}
			\setlength{\tabcolsep}{0.0025\textwidth}
			
			\begin{tabular}{|p{1.80cm}|p{0.9cm}|p{0.65cm}|p{0.95cm}|p{1cm}|p{1cm}|p{1.5cm}|}
				\hline
				\textbf{Characteristics} & \textbf{Mono-chrome} & \textbf{RGB}&\textbf{Spectro- meter}&\textbf{MSI}&\textbf{HSI}&\textbf{USI} \\ \hline
				Spatial Info.   & $\checkmark$ & $\checkmark$ & $X$ & $\checkmark$ & $\checkmark$& $\checkmark$  \\ \hline
				Spectral Info. & $X$ & $X$ & $\checkmark$ & \textit{small} & $\checkmark$ & $\checkmark$ \\ \hline
				Number of Bands &1&2&Dozens to 100&3 to 10&10-100& more than 1000  \\ \hline
				Applications    &intensity&color&solid, fluid, gases&solid, fluid detection&solid, fluid analysis &solid, fluid, gases analysis\\ \hline
				Degree of Comm. & $\checkmark$ & $\checkmark$ & $\checkmark$ & $\checkmark$ & $\checkmark$& Emerging \\ \hline
				
			\end{tabular}
			
		\end{table}
		
		\subsection{Compression and Transmission}
HSI image compression and transmission differ from regular RGB or grayscale images due to the high dimensionality of data across spatial and spectral bands. Efficient compression is vital to preserve spectral and spatial information \cite{HSI:Compression:Zhang:2023}. Spatial redundancy can be reduced using methods like wavelet compression, discrete cosine transform, and block compression \cite{HSI:Compression:Dua:2020}. Techniques such as principal component analysis (PCA), independent component analysis (ICA), or singular value decomposition can reduce hyperspectral data dimensionality \cite{HSIDimRed:Jayaprakash:2018}. Band selection or fusion can also merge similar bands \cite{HSIDimRed:Radhesyam:2024}. After compression, entropy encoding (e.g., Huffman or arithmetic coding) reduces frequent values to fewer bits \cite{HSICompression:Coding:Mbewe:2017}.

		\subsection{Spectral Unmixing}
		
\begin{figure}[!htbp]
			\centering
			\includegraphics[width=0.75\linewidth, keepaspectratio]
			{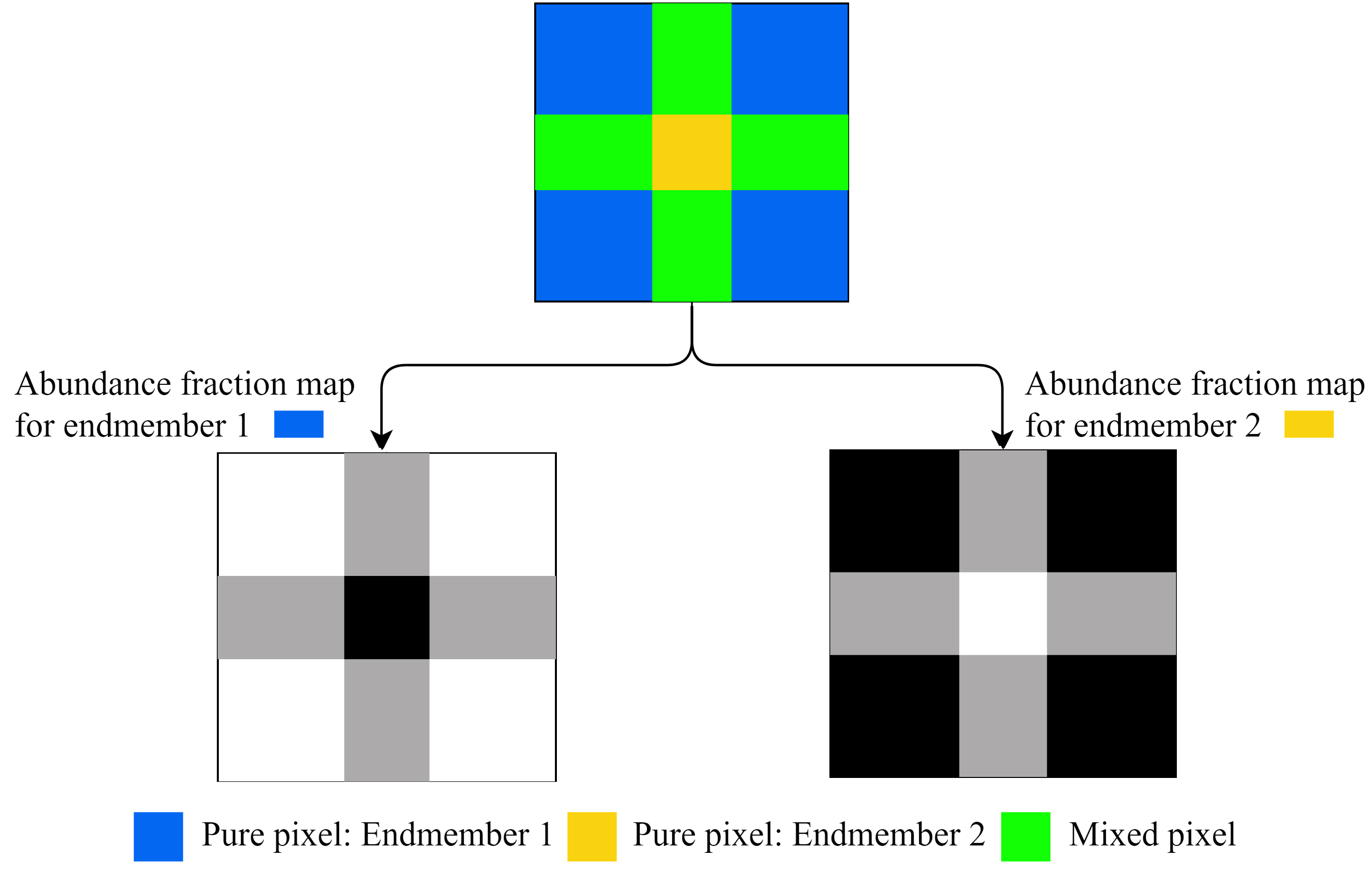}
			\caption{Spectral Un-mixing}
			\label{fig:spectralunmixing}
		\end{figure}

		HSI differentiates materials based on spectral fingerprints, but spectral mixing remains a challenge, where a single pixel captures multiple materials due to limited spatial resolution. This occurs in two forms: linear mixing, when materials are spatially separated within a pixel, and nonlinear mixing, arising from material interactions. Addressing these distinctions enhances HSI precision \cite{HSIUnmixing:Devi:2016}. Factors such as sensor resolution, material heterogeneity, and surface interactions contribute to spectral mixing. In agriculture, soil, crops, and water create mixed signatures, while urban and forest environments involve additional complexities. Spectral unmixing techniques decompose mixed pixels into endmembers through endmember extraction and abundance map estimation. Methods like Pixel Purity Index (PPI), Fast Iterative Pixel Purity Index (FIPPI), and N-Finder (N-FINDR) identify endmembers. Abundance map estimation determines each endmember’s proportion within a pixel, improving data interpretability \footnote{https://kr.mathworks.com/help/images/getting-started-with-hyperspectral-image-analysis.html}, as shown in Figure \ref{fig:spectralunmixing}.
	
		\subsection{Sub-Pixel Classification}
Sub-pixel classification in HSI identifies multiple materials within a single pixel, which is common due to limited spatial resolution. This occurs in remote sensing applications where features like vegetation, soil, or urban structures are mixed within one pixel. It addresses spectral mixing through unmixing techniques like the linear mixture model and least-squares linear unmixing, which decompose mixed pixels into endmembers and estimate abundance fractions. Unlike hard classification, sub-pixel classification uses soft classification to assign fractional class memberships, improving accuracy. This approach is applied in land cover analysis, precision agriculture, and mineral mapping, with deep learning methods also emerging \cite{DLSubPixelClassification:Zhu:2024,DLSubPixelClassification:ARUN:2018}.

		\subsubsection{Linear Mixture Model}
		In the linear mixture model, each pixel’s spectral value in a HSI image is represented as a combination of the spectra of pure components (endmembers) within that pixel \cite{HSIUnmixing:Devi:2016}. Mathematically, this model can be expressed as: \(P_i = \sum_{j=1}^n F_j \cdot R_{ij} + E_i\). 
		While expanding this equation across all spectral bands gives: 
		\begin{equation}\label{eq:linearunmix1}
			P = R \cdot F + E
		\end{equation}
		this represents an inversion problem, where we attempt to determine unknown causes based on observable data. In HSI unmixing, this means estimating the proportion of each material present in a pixel based on the combined light spectrum detected by the sensor. The observed spectrum is a mixture of the known spectra of pure materials (endmembers), and the challenge lies in determining the fractional abundance of each material within the pixel. In Equation \ref{eq:linearunmix1}, \({P}\) is the observed spectrum, \({R}\) contains the known spectra of the materials, \({F}\) represents the unknown fractions of each material that we seek to determine, and \({E}\) accounts for errors or noise. Solving this inversion problem involves estimating the values of \({F}\), which is modeled as a system of linear equations in Equation \ref{eq:linearunmix2}.
		\begin{equation} \label{eq:linearunmix2}
			\scriptsize
			\begin{bmatrix}
				P_1 \\
				P_2 \\
				\vdots \\
				P_m
			\end{bmatrix}
			=
			\begin{bmatrix}
				R_{11} & R_{12} & \dots & R_{1n} \\
				R_{21} & R_{22} & \dots & R_{2n} \\
				\vdots & \vdots & \ddots & \vdots \\
				R_{m1} & R_{m2} & \dots & R_{mn}
			\end{bmatrix}
			\begin{bmatrix}
				F_1 \\
				F_2 \\
				\vdots \\
				F_n
			\end{bmatrix}
			+
			\begin{bmatrix}
				E_1 \\
				E_2 \\
				\vdots \\
				E_m
			\end{bmatrix}
		\end{equation}
		To obtain a unique solution for the inversion problem, it is assumed that the number of spectral bands \( m \) is greater than the number of endmembers \( n \) in the pixel. Additionally, to ensure a physically realistic solution, two constraints are applied: (i) \(\sum_{j=1}^n F_j = 1\), which ensures that the sum of the fraction coefficients equals one, representing the entire pixel area; and (ii) \(F_j \geq 0 \text{ for } j = 1, \ldots, n\), which ensures that each fraction coefficient is non-negative to avoid negative areas. These constraints ensure that the model accurately reflects the contributions of each endmember to the observed pixel spectra. However, selecting an adequate number of endmembers remains a challenge, as insufficient endmembers can lead to inaccurate unmixing results \cite{tseng:spectralUnmixing:2000}.
		\subsubsection{Least-Squares (LS) Linear Unmixing}
		In the least-squares approach to the inversion problem, the unmixing coefficients are found by minimizing the sum of the squares of the errors. When the constraints (i) and (ii) discussed above are ignored, it is equivalent to solving a set of equations of the following form:
		\(\frac{\partial}{\partial F_j} \left( \sum_{i=1}^{m} \sum_{j=1}^{n} \left( P_i - \sum_{j} F_j R_{ij} \right)^2 \right) = 0\).
		Expanding this equation results in the classical multiple linear regression matrix equation, and the least-squares fit estimation for the unmixing coefficients \(F_j\) becomes \(\hat{{F}} = \left( {R}^T {R} \right)^{-1} {R}^T {P}\). Taking the difference between the observed and the calculated spectral reflectance results in the estimated error terms of the observations as \(\hat{{E}} = {P} - {R} \hat{{F}}\). The variance of the error terms is a quantitative measure of how well the mixture modeling fits the data, which indicates the feasibility of the solution. The estimation of the variance is \(\sigma^2 = \frac{\hat{{E}}^T \hat{{E}}}{m - n}\). The least-squares solution can be modified to fulfill the requirements of constrained fits. For the first requirement, the estimated coefficients \(F_j\) are modified by taking the linear constraints \({A}{F} = {b}\) into account:
		\begin{equation}
			\scriptsize
			\hat{{F}} = \left( {R}^T {R} \right)^{-1} {R}^T {P} + \left( {A} \left( {A}^T \left( {R}^T {R} \right)^{-1} {A} \right)^{-1} {A}^T \right) \left( {b} - {A} {F} \right)
		\end{equation}
		Where \({A}\) is a \(1 \times n\) matrix defined as: \({A} = \begin{bmatrix} 1 & 1 & \dots & 1 \end{bmatrix}, \quad {b} = \begin{bmatrix} 1 \end{bmatrix}\). No analytical solutions are known for the inversion problem when the coefficients are constrained by inequalities. The use of iterative approaches, such as the simplex method, is needed to implement those constraints \cite{tseng:spectralUnmixing:2000}. Other mathematical methods used for unmixing include PCA, Supervised Classification Analysis (SCA), and Multivariate Curve Resolution (MCR).\par
		
		Some of the state-of-the-art unmixing tools are listed in Table \ref{table:ToolsUsedForUnmixing}.
		
		\begin{table}[h!]
			\scriptsize
			\setlength{\tabcolsep}{0.0025\textwidth}
			\caption{Tools used for Unmixing}
			\label{table:ToolsUsedForUnmixing}
			\centering
			\begin{tabular}{|p{1.95cm}|p{3.5cm}|p{3cm}|}
				\hline
				\textbf{Tools} & \textbf{Features} & \textbf{Reference} \\ \hline
				NV5 GEOSPATIAL SOFTWARE & Matched Filtering (MF), Mixture Tuned Matched Filtering, Linear Spectral Unmixing & [\url{https://www.nv5geospatialsoftware.com}]\\ \hline
				PPI & Extract endmember signatures using PPI & [\url{https://kr.mathworks.com/help/images/ref/ppi.html}]\\ \hline
				Fast Iterative PPI & Extract endmember signatures using Fast Iterative PPI & \url{https://kr.mathworks.com/help/images/ref/fippi.html}\\ \hline
				N-FINDR & Extract endmember signatures using N-FINDR & [\url{https://kr.mathworks.com/help/images/ref/nfindr.html}]\\ \hline
				Estimate-AbundanceLS & Estimate abundance maps & [\url{https://kr.mathworks.com/help/images/ref/estimateabundancels.html}]\\ \hline
			\end{tabular}
		\end{table}
		
		\subsubsection{Spectral Unmixing Challenges}
		Spectral unmixing faces several key challenges that complicate its accuracy and practicality. One major issue is the selection of endmembers, meaning accurately identifying the pure materials in a scene is crucial, and errors in this step can lead to incorrect results. Computational complexity is another challenge, as nonlinear unmixing methods, while more accurate, are often too demanding for real-time applications or large datasets. The variability of material signatures due to environmental factors, such as moisture or temperature, further complicates the process. Additionally, sensor noise and atmospheric interference can distort spectral data, making accurate unmixing more difficult. However, ongoing advancements in algorithms and computational techniques continue to enhance the effectiveness of spectral unmixing.
		
		\section{Augmenting HSI with Deep Learning}\label{sec:AugmentingHSIWithDeepLearning}
		
	Instead of just referring to HSI, it is now more appropriate to call it "AI-driven HSI," where deep learning (DL) has revolutionized HSI and multimodal HSI by enhancing the processing and interpretation of high-dimensional data. HSI captures a wide spectrum of wavelengths, which presents challenges due to its complexity and large volume. Traditional machine learning methods struggle with such data, but DL models, particularly CNNs, excel by automatically learning representations from raw data, making them effective for tasks like material classification, anomaly detection, and environmental monitoring. In multimodal HSI, DL facilitates the fusion of data from sources like LiDAR (Light Detection and Ranging), thermal imaging, and SAR (synthetic aperture radar), enabling richer feature extraction and more accurate analyses \cite{2020:HSIMultiModelFusionNetwork:IEEEConf, 2023:HSISpatialEnhancementReview:IEEEJ, wang2022hyperspectral, zhuo2022deep, vivone2022panchromatic, li2020multi, wu2023hsr, wei2021crops}. For example, combining hyperspectral and thermal data improves precision agriculture by detecting early signs of crop health issues \cite{ang2021big, 2021:Multimoel:DL:HSI:RuitMaturityEstimation:Sensors,Huang2012, Archibald1999}. DL excels in modeling complex spatial-spectral correlations, with CNNs, transformers, and autoencoders effectively capturing patterns between spectral bands, improving classification and detection in HSI. This has advanced fields such as urban planning, environmental monitoring, and medical diagnostics \cite{GARG2021100370, 2007:HSI:MonitoringWoundHealing:IEEEC}. As DL evolves, its role in multimodal HSI will expand, enhancing data fusion for applications like remote sensing and industrial inspection. Moreover, advances in computational resources and open-source DL libraries like TensorFlow and PyTorch have lowered the barriers to deploying sophisticated models for HSI \cite{Zhang:DLFramworks:2021}.
		
		\subsection{State-of-the-art DL Models and Networks}
		
	Deep learning models such as Deep Neural Networks (DNNs), Multilayer Perceptrons (MLPs), and CNNs are essential in HSI and multimodal HSI tasks like classification, anomaly detection, and feature extraction. MLPs are effective in pixel-wise classification, leveraging high-dimensional spectral inputs to learn non-linear relationships, making them suitable for small datasets or hybrid approaches. CNNs, commonly used in environmental monitoring and object identification, capture spatial hierarchies through convolutional and pooling layers. For tasks involving temporal or sequential data, models like RNNs and their advanced variants, LSTM networks and GRUs, are key in multimodal HSI. Autoencoders assist in dimensionality reduction by compressing and reconstructing high-dimensional spectral data, while GANs generate new spectral data, augmenting datasets or reducing noise. Recent advancements like Vision Transformers (ViTs) and DBNs enhance feature representation and clustering in HSI. Hybrid models like CNN-LSTM combine spatial and spectral data for better classification and anomaly detection. Table \ref{Table:DLModelsApplications} provides an overview of these models in HSI and multimodal HSI tasks.
	
	In multimodal HSI, DL models such as ResNet50, VGG16, InceptionV3, and DenseNet are crucial for data fusion tasks involving multiple sensors. These models enhance applications like urban planning, land-use mapping, and environmental monitoring by fusing hyperspectral data with LiDAR or thermal imaging. For real-time object detection in HSI, Faster R-CNN and YOLO provide fast, accurate analysis, crucial for autonomous aerial monitoring using hyperspectral drones \cite{ELHARROUSS2024100645}. In land use mapping or infrastructure monitoring, these models improve detection accuracy by integrating HSI data with additional inputs like LiDAR or RGB imagery. Table \ref{tab:deep_learning_hsi} summarizes deep learning architectures, their features, and applications in hyperspectral and multimodal HSI, helping researchers select the right architecture for their problems.
		
		\begin{table*}[!htpb]
			\centering
			\caption{Deep Learning Models and their Applications in HSI and Multimodal HSI}
			\label{Table:DLModelsApplications}
			\setlength{\tabcolsep}{0.0025\textwidth} 
			\scriptsize
			\begin{tabular}{|p{1.3cm}|p{2.5cm}|p{8.9cm}|p{3.5cm}|p{1.15cm}|}
				\hline
				\textbf{Model} & \textbf{Category} & \textbf{Applications in HSI and Multimodal HSI} & \textbf{Examples}  & \textbf{References} \\ \hline
				{DNNs} & Foundational DL model & Non-linear mappings for tasks like material classification and anomaly detection in HSI. & Airborne Visible/Infrared Imaging Spectrometer (AVIRIS), Hyperion & \cite{rs10091454}  \\ \hline
				{MLP} & Fully Connected Neural Network & Used for basic classification tasks in HSI, especially when spectral data is low-dimensional. & Reflective Optics System Imaging Spectrometer, Sentinel-2 & \cite{rs12030355, ObjectDetection:Konstantinos:2015} \\ \hline
				{CNNs} & Convolutional Neural Network & Spatial feature extraction for environmental monitoring, object identification, and material classification in HSI. & AVIRIS, Hyperion, Reflective Optics System Imaging Spectrometer & \cite{rs15143532, HSIClassificationWithDL:2018:Yang} \\ \hline
				{RNNs} & Recurrent Neural Network & Temporal sequence processing in multimodal HSI for tasks involving spectral or temporal data sequences & CASI & \cite{8082108, Wu2017} \\ \hline
				{LSTMs/ GRUs} & Advanced RNNs & Sequential data processing for spectral or time-series data in multimodal HSI & Moderate Resolution Imaging Spectrometer (MODIS), CASI & \cite{Kong2018, ZhouFengLSTMs:2017} \\ \hline
				{Autoencoders} & Unsupervised Learning & Dimensionality by compressing and reconstructing high-dimensional hyperspectral data & MODIS, AVIRIS, Hyperion & \cite{StackedAutoEncoder:2016, StackedAutoEncoder:2017, 7875467:2017:Ronald} \\ \hline
				{GANs} & Generative Models & Data augmentation and noise reduction in HSI by generating plausible spectral data. & Hyperspectral Image Generator (HyGAN), AVIRIS & \cite{8307247:ZHU:2018, GANsBasedHSI:Zhan:2023, HSI:Hao:TransformerGANS} \\ \hline
				{Transformers} & Attention-based Models & Modeling relationships in hyperspectral or multimodal datasets (Vision Transformers) & Sentinel-2, AVIRIS, ROSIS & \cite{HSI:Hao:TransformerGANS} \\ \hline
				{DBNs / SOMs} & Clustering, Visualization & Clustering and visualizing high-dimensional HSI data for pattern analysis& AVIRIS, Hyperion & \cite{HUANG2019233, 8595297} \\ \hline
				{CNN-LSTM} & Hybrid Models & Combining spatial feature extraction (CNNs) with sequence-learning (LSTMs) for complex multimodal HSI tasks, including classification and anomaly detection. & MODIS, ROSIS, AVIRIS & \cite{ZhouFengLSTMs:2017} \\ \hline
				{DQN} & Reinforcement Learning & Decision-making in real-time HSI applications like autonomous drone navigation for remote sensing and environmental monitoring & CASI, AVIRIS, HYDICE & \cite{HSI:DoubleDeepQNetwork:Yang:2023} \\ \hline
			\end{tabular}
		\end{table*}
		
		\begin{table*}[!htpb]
			\centering
			\caption{Comparison of Deep Learning Models for Hyperspectral Imaging and Multimodal Hyperspectral Imaging}
			\label{tab:deep_learning_hsi}
			\scriptsize
			\setlength{\tabcolsep}{0.0025\textwidth} 
			\begin{tabular}{|p{1.85cm}|p{2cm}|p{4cm}|p{1.25cm}|p{2.25cm}|p{2.25cm}|p{2.5cm}|p{1.10cm}|}
				\hline
				\textbf{DL Model} & \textbf{Main Task} & \textbf{Application} & \textbf{Training Complexity} & \textbf{Training Requirement} & \textbf{Efficiency} & \textbf{Key Considerations} & \textbf{Studies} \\ \hline
				ResNet50 & Classification, Feature Extraction & Extracts spectral/spatial features; integrates HSI with RGB, LiDAR; material classification & Moderate to High & Large datasets, powerful GPUs & High accuracy & Needs modification for high-dimensional data & \cite{ResNetRGBHSI:Bu:2024, ResNetHSI:Banerjee:2023, ResNetHSI:Xiang:2024, VGG16ResNet:HSI:2024:Jannat, ResNet50VGG16InceptionV3:Rajendran:2022, ResNetShuffleNetDenseNetMobileNet:PourdarbaniSDRA:2023} \\ \hline
				VGG16 & Classification, Feature Representation & Manages high-dimensional spectral bands; fuses spectral features & High & Large datasets, high GPU memory & Moderate; overfitting risk & Computationally heavy & \cite{VGG16ResNet:HSI:2024:Jannat, VGGHSI:Grechkin:2023, VGG16HSI:Sun:2024, VGG16HSI:Ye:2021, ResNet50VGG16InceptionV3:Rajendran:2022} \\ \hline
				InceptionV3 & Multi-scale Feature Extraction & Captures fine and spatial details; effective for multi-scale HSI fusion & High & Requires large, annotated datasets & High efficiency & Captures both fine and large-scale features & \cite{ResNet50VGG16InceptionV3:Rajendran:2022} \\ \hline
				MobileNet & Efficient Classification & Lightweight for real-time HSI; ideal for low-power tasks & Low & Moderate hardware, suitable for mobile & Highly efficient & Limited depth for complex tasks & \cite{MobileNetDensNet:Hadi:2024, MobileNetDensNet:wang:2020, ResNetShuffleNetDenseNetMobileNet:PourdarbaniSDRA:2023} \\ \hline
				DenseNet & Classification, Feature Reuse & Efficient feature reuse for high-dimensional HS and multimodal fusion & Moderate & Reduced training time due to feature reuse & High efficiency & Avoids overfitting & \cite{MobileNetDensNet:Hadi:2024, MobileNetDensNet:wang:2020, DenseNet:Song:2020, ResNetShuffleNetDenseNetMobileNet:PourdarbaniSDRA:2023} \\ \hline
				RCNN & Object Detection, Region Proposal & Detects spectral anomalies; multimodal object detection & High & High computational resources & less efficient; slower than modern methods & Resource-intensive & \cite{FastFasterRCNN:YOLO:Alexnet:VGG16Resnet18Resnet50Resnet01InceptionV3:Azam:2022} \\ \hline
				Fast RCNN & Faster Object Detection & Speeds up spectral detection; fuses spectral/spatial data & High & Large datasets, optimized for speed & Faster than RCNN & Balances speed and accuracy & \cite{FastFasterRCNN:YOLO:Alexnet:VGG16Resnet18Resnet50Resnet01InceptionV3:Azam:2022} \\ \hline
				Faster RCNN & Real-time Object Detection & Detects materials/objects in real-time; fuses spectral/spatial data & Moderate to High & Large datasets & Efficient for real-time detection & Improved efficiency over RCNN & \cite{FasterRCNN:ALBAHLI2021} \\ \hline
				YOLO (You Only Look Once) & Real-time Object Detection & Fast detection in HSI; works in multimodal scenarios & Moderate & Moderate dataset size & Extremely efficient for real-time use & Low latency & \cite{YOLOHSI:2021:Yan, YOLOMSI:2024:James, YOLOHSI:2024:Ying} \\ \hline
				SSD (Single Shot Detector) & Multi-scale Object Detection & Handles varying object sizes; multimodal detection at different scales & Moderate & Balanced dataset required & High efficiency for size variation & Suited for large object detection & \cite{SSD:Shahin2021} \\ \hline
			\end{tabular}
		\end{table*}

		\subsection{Basic Understanding of HSI Image Reconstruction}
	HSI captures images across a wide spectrum, providing rich spectral information for each pixel. However, reconstructing HSI can be challenging due to noise and distortion. To address this, techniques like regularization methods (L1 and L2 regularization, Tikhonov regularization, and total variation regularization) \cite{HSISpetraReconstructionUsingRegularization:Ma:2023,HSISpetraReconstructionUsingRegularization:He:2022}, optimization algorithms (gradient descent, Adam optimizer, proximal gradient methods, alternating least squares, and coordinate descent) \cite{HSISpetraReconstructionUsingOptimization:Zhu:2021,96}, and advanced deep learning-based methods have been developed. Deep learning, especially CNNs, excels at identifying patterns by leveraging neighboring wavelengths and intensities, enabling accurate reconstruction of spectra from limited measurements. During acquisition, intensity measurements are collected across spectral bands and represented as a 3D data cube. The observed intensity at wavelength $\lambda$ for pixel $i$ can be expressed as:
		\begin{equation}
			I_i(\lambda) = S_i(\lambda) \cdot x(\lambda) + n_i(\lambda)
		\end{equation}
		where $S_i(\lambda)$ is the spectral sensitivity, $x(\lambda)$ is the target spectrum, and $n_i(\lambda)$ is measurement noise. The measured intensities are arranged into feature vectors for each pixel:
		\begin{equation}
			\mathbf{I}_i = [I_i(\lambda_1), I_i(\lambda_2), \ldots, I_i(\lambda_N)]^T
		\end{equation}
		These feature vectors serve as input to a deep learning model, commonly based on CNNs like U-Net, which effectively capture spatial and spectral features. The training process involves minimizing the difference between the reconstructed and ground truth spectra using a loss function, such as Root Mean Square Error (RMSE):
		\begin{equation}
			\text{RMSE} = \sqrt{\frac{1}{N} \sum_{i=1}^{N} \|x_{recovered,i} - x_{GT,i}\|_2^2}
		\end{equation}
		After training, the model is evaluated on a test set, and performance is assessed using metrics like RMSE or Peak Signal-to-Noise Ratio (PSNR). For new intensity measurements, the reconstructed spectrum is given by:
		\begin{equation}
			\mathbf{x}_{recovered,j} = DL(\mathbf{I}_j)
		\end{equation}
		Deep learning offers advantages such as improved accuracy, robustness to noise, and flexibility for different applications.
		
		\subsection{Basic Understanding of HSI Image Feature Extraction}

		Understanding HSI feature extraction using deep learning is crucial. The goal is to extract both spectral and spatial information, as HSI images contain high-dimensional data across many wavelengths, making them ideal for deep learning analysis. CNN models, for example, are excellent at identifying critical features by learning spectral signatures from contiguous bands \cite{chen2016deep}. These features help identify and classify materials, while spatial features from neighboring pixels provide contextual information like texture and structure. Given HSI's high dimensionality, dimensionality reduction is essential and can be achieved through methods like PCA \cite{ZhouFengLSTMs:2017,gonzalez2004fusion}, Independent ICA \cite{BARANWAL2022103118:ICA:2022, PCA:ICA:Clustering,Liu2017:PCA:ICA}, Spectral Angle Mapper (SAM) \cite{Lim2024:SAM}, Autoencoder \cite{AutoEncoder:FeatureExtraction:2022}, CNN, or a combination of PCA and CNN \cite{3DCNN:FeatureExtraction:2020}. These methods reduce dimensionality without losing key features. Deep learning enables end-to-end feature extraction, mapping raw data to labels directly. Multimodal-HSI combines additional data like LiDAR or RGB images, enhancing feature richness \cite{2020:HSIMultiModelFusionNetwork:IEEEConf}. Transfer learning, using pretrained models for adapting features from similar datasets, is valuable when labeled HSI data is scarce.\par		
		CNNs and autoencoders effectively extract patterns from high-dimensional hyperspectral data by capturing both spectral and spatial information. In feature extraction, CNNs apply convolution across spectral bands to capture unique signatures for each pixel, represented as a spectral vector \( \mathbf{x} = [x_1, x_2, \dots, x_B] \), where \( B \) is the number of spectral bands \cite{HSIFeatureExtract:Quan:2020}. For simplicity, the convolution is expressed as
		\[
		y_i = \sum_{j=-k}^k w_j \cdot x_{i+j}
		\]
		Here, \( y_i \) is the feature at the \( i \)-th spectral band, \( w_j \) are filter weights, and \( x_{i+j} \) represents neighboring spectral values, enabling the model to capture relationships across spectral bands, making it easier to distinguish different materials based on their spectral properties. For spatial feature extraction, CNNs apply 2D convolution across neighboring pixels \cite{HSIFeatureExtract:Quan:2020}, which can be expressed as:
		\[
		y_{i,j} = \sum_{m=-k}^k \sum_{n=-k}^k w_{m,n} \cdot x_{i+m, j+n}
		\]
		where \( y_{i,j} \) is the output feature at spatial position \( (i, j) \), \( w_{m,n} \) are the convolutional filter weights, and \( x_{i+m, j+n} \) are the pixel values in the local neighborhood around \( (i, j) \). By capturing spatial patterns, this operation enhances the model’s ability to identify textures and structural information. 
		
		In order to capture both spectral and spatial information simultaneously, models often employ 3D convolutional layers, which convolve across both the spatial dimensions and the spectral bands \cite{HSIFeatureExtract:Quan:2020}. The 3D convolution operation can be represented as:
		\[
		y_{i,j,b} = \sum_{m=-k}^k \sum_{n=-k}^k \sum_{p=-k}^k w_{m,n,p} \cdot x_{i+m, j+n, b+p}
		\]
		Here, \( y_{i,j,b} \) is the output feature at spatial position \( (i, j) \) and spectral band \( b \), with \( w_{m,n,p} \) as the 3D convolutional kernel weights, capturing spatial and spectral dependencies. To address the high dimensionality of hyperspectral data, autoencoders are often used for dimensionality reduction by learning compressed data representations. In an autoencoder, the encoder compresses the input spectral vector \( \mathbf{x} \) to a lower-dimensional representation \( \mathbf{z} \) as \( \mathbf{z} = f(\mathbf{x}) = \sigma(\mathbf{W} \mathbf{x} + \mathbf{b}) \), where \( \mathbf{W} \) and \( \mathbf{b} \) are the weight matrix and bias of the encoder, and \( \sigma \) is an activation function like ReLU or sigmoid. The decoder then reconstructs \( \mathbf{x} \) from \( \mathbf{z} \) using 
		\[
		\hat{\mathbf{x}} = g(\mathbf{z}) = \sigma'(\mathbf{W}' \mathbf{z} + \mathbf{b}')
		\]
		where \( \mathbf{W}' \) and \( \mathbf{b}' \) are the weights and bias for the decoder. The reconstructed \( \hat{\mathbf{x}} \) is compared to the original \( \mathbf{x} \) to minimize reconstruction loss, preserving essential information in a compact form.
		
		For optimizing these deep learning models, appropriate loss functions are employed. For classification tasks, cross-entropy loss 
		\[
		\mathcal{L}_{\text{cross-entropy}} = -\sum_{c=1}^C y_c \log(\hat{y}_c)
		\]
		can be used, where \( C \) is the number of classes, \( y_c \) is the true label, and \( \hat{y}_c \) is the predicted probability for each class. For reconstruction tasks, MSE (Mean Squared Error) and RMSE (Mean Squared Error) are commonly applied, which are:
		\[
		\mathcal{L}_{\text{MSE}} = \frac{1}{N} \sum_{i=1}^N (x_i - \hat{x}_i)^2
		\]
		\[
		\mathcal{L}_{\text{RMSE}} = \sqrt{\frac{1}{N} \sum_{i=1}^N (x_i - \hat{x}_i)^2}
		\]
		where \( N \) is the number of spectral bands, and \( x_i \) and \( \hat{x}_i \) are the original (ground truth) and reconstructed spectral values, respectively. To evaluate reconstruction performance, sometimes RMSE is preferred instead of MSE because RMSE is in the same units as the target variable, making it easier to interpret. In addition, RMSE penalizes larger errors more due to the squared differences, making it more sensitive to outliers. These features make it suitable as the loss function to quantify the difference between the ground truth spectra \( x \) and the reconstructed spectra \( \hat{x} \). Some important metrics other than are given in Table \ref{tab:hsi_metric} which are used in different situations.
			
		\begin{table*}[!htpb]
			\centering
			\setlength{\tabcolsep}{0.0025\textwidth}
			\caption{List of Commonly Used Evaluation Metrics for HSI Analysis}
			\label{tab:hsi_metric}
			\scriptsize
			
			\begin{tabular}{|p{3.5cm}|p{5cm}|p{7cm}|p{2cm}|}\hline 
				\textbf{Metric} & \textbf{Formula} & \textbf{Description} & \textbf{References} \\ \hline 
				
				OA \cite{nelson1983detecting} (Overall Accuracy) &
				\( \text{OA} = \frac{\sum_{i=1}^{C} TP_i}{N} \) &
				Calculated by dividing the number of true positive pixels by the total pixel count, reflecting accuracy across all classes. &
				\cite{chen2016deep,GANsBasedHSI:Zhan:2023,ma2024multilevel,yu2024hyperspectral,yang2022hyperspectral} \\ \hline
				
				AA \cite{nelson1983detecting} (Average Accuracy) &
				\( \text{AA} = \frac{1}{C} \sum_{i=1}^{C} \frac{TP_i}{T_i} \) &
				Computed as the mean accuracy across all classes, reflecting model performance for each class independently. &
				\cite{chen2016deep,GANsBasedHSI:Zhan:2023,yang2022hyperspectral} \\ \hline
				
				Kappa coefficient \cite{cohen1960coefficient} &
				\( \kappa = \frac{p_o - p_e}{1 - p_e} \) &
				Measures the agreement between predicted and actual classifications while accounting for chance agreement. &
				\cite{chen2016deep,GANsBasedHSI:Zhan:2023,yu2024hyperspectral,yang2022hyperspectral} \\ \hline
				
				AUC \cite{fawcett2006introduction} (Area Under the Curve) &
				\( \text{AUC} = \int_{0}^{1} \frac{\text{TP}}{\text{TP} + \text{FN}} \, d\left(\frac{\text{FP}}{\text{FP} + \text{TN}}\right) \) &
				Evaluates model performance by calculating the area under the ROC curve, indicating discriminatory power. &
				\cite{imamoglu2018hyperspectral,wang2022meta} \\ \hline
				
				MSE \cite{RMSE:MSE:PriyaVarshini2021} (Mean Squared Error) &
				\( \text{MSE} = \frac{1}{N} \sum_{i=1}^{N} (y_i - \hat{y}_i)^2 \) &
				Measures average squared differences between predicted and actual values, emphasizing larger errors. &
				\cite{zhao2020deep} \\ \hline
				
				RMSE \cite{RMSE:MAE:Hodson:2022} (Root Mean Squared Error) &
				\( \text{RMSE} = \sqrt{\frac{1}{N} \sum_{i=1}^{N} (y_i - \hat{y}_i)^2} \) &
				Provides error metric in the same unit as the data, reflecting the typical deviation of predictions. &
				\cite{wang2022hsisuper,yuan2017hyperspectral} \\ \hline
				
				PSNR \cite{EvaluationParams:Sethi:2022} (Peak Signal-to-Noise Ratio) &
				\( \text{PSNR} = 10 \cdot \log_{10} \left(\frac{\text{MAX}^2}{\text{MSE}}\right) \) &
				Measures image quality by comparing peak signal to noise, where higher values indicate better fidelity. &
				\cite{2023:HSISpatialEnhancementReview:IEEEJ,wang2022hsisuper,avagian2019fpga,yang2019hyperspectral} \\ \hline
				
				ERGAS \cite{wald2002data} (Relative Dimensionless Global Error in Synthesis) &
				\( \text{ERGAS} = 100 \cdot \frac{1}{B} \sqrt{\sum_{b=1}^{B} \left(\frac{\text{MSE}_b}{\mu_b^2}\right)} \) &
				Normalizes error across multiple image bands, where lower values reflect greater accuracy. &
				\cite{2023:HSISpatialEnhancementReview:IEEEJ,wang2022hsisuper,yuan2017hyperspectral,yang2019hyperspectral} \\ \hline
				
				SAM \cite{kruse1993spectral} (Spectral Angle Mapper) &
				\( \text{SAM} = \arccos \left( \frac{\sum_{i=1}^{N} y_i \hat{y}_i}{\sqrt{\sum_{i=1}^{N} y_i^2} \cdot \sqrt{\sum_{i=1}^{N} \hat{y}_i^2}} \right) \) &
				Measures spectral similarity, expressed as the angle between two spectra, where smaller angles indicate higher similarity. &
				\cite{2023:HSISpatialEnhancementReview:IEEEJ}, \cite{wang2022hsisuper}, \cite{yuan2017hyperspectral}, \cite{yang2019hyperspectral}, \cite{shah2020nonlinear} \\ \hline
				
				SSIM \cite{wang2004image} (Structural Similarity Index Measure) &
				\( \text{SSIM}(x, y) = \frac{(2 \mu_x \mu_y + C_1)(2 \sigma_{xy} + C_2)}{(\mu_x^2 + \mu_y^2 + C_1)(\sigma_x^2 + \sigma_y^2 + C_2)} \) &
				Evaluates image quality by measuring structural information, luminance, and contrast differences. &
				\cite{2023:HSISpatialEnhancementReview:IEEEJ,xu2021similarity,wang2022hsisuper,avagian2019fpga,yang2019hyperspectral} \\ \hline
			\end{tabular}
		\end{table*}	
				
		\subsection{Basic Understanding of HSI Image Classification}
		In HSI image classification, the primary objective is to assign a unique label to each pixel vector within the HSI image cube based on its spectral or combined spectral-spatial properties. Mathematically, an HSI cube can be represented as:
		
		\[
		\mathbf{X} = [\mathbf{x}_1, \mathbf{x}_2, \dots, \mathbf{x}_B]^T = 
		\begin{bmatrix}
			x_{1,1} & x_{1,2} & \dots & x_{1,N \times M} \\
			x_{2,1} & x_{2,2} & \dots & x_{2,N \times M} \\
			\vdots & \vdots & \ddots & \vdots \\
			x_{B,1} & x_{B,2} & \dots & x_{B,N \times M}
		\end{bmatrix}
		\]
		
		\begin{figure}[!htbp]
			\centering
			\includegraphics[width=0.45\linewidth, keepaspectratio]{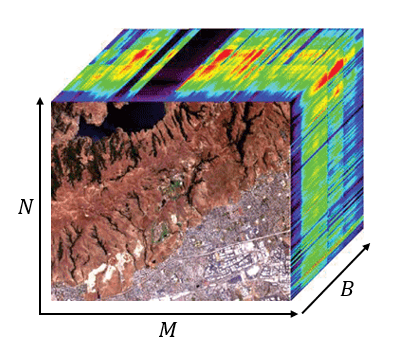}
			\caption{HSI image representation}
			\label{fig:HSIBMNHyperCube}
		\end{figure}  
		
		The dimensions of the matrix \( \mathbf{X} \) are \( B \times (N \times M) \), where each row corresponds to a spectral band, and each column corresponds to a pixel. \( B \) denotes the total number of spectral bands. Each band consists of \( N \times M \) spatial samples, corresponding to the pixels in the image, as shown in Figure \ref{fig:HSIBMNHyperCube}. These pixels belong to one of \( Y \) classes. For each pixel, the corresponding spectral vector \( \mathbf{x}_i = [x_{1,i}, x_{2,i}, x_{3,i}, \dots, x_{B,i}]^T \) captures the reflectance values across all \( B \) spectral bands. Each spectral vector \( \mathbf{x}_i \) is associated with a class label \( y_i \in \{1, 2, \dots, Y\} \), indicating the material or object type that the pixel represents. 
		
		The classification task in HSIC can be viewed as an optimization problem, where the goal is to learn a mapping function \( f_c(\cdot) \) that takes the input data \( \mathbf{X} \) and produces the corresponding predicted labels \( \hat{\mathbf{Y}} \). The objective is to minimize the difference between the predicted labels and the true labels, which can be mathematically expressed as:
		
		\[
		\hat{\mathbf{Y}} = f_c(\mathbf{X}, \phi)
		\]
		
		Here, \( \phi \) represents the parameters of the model that are learned during the training process. These parameters govern the transformations applied to the input data \( \mathbf{X} \) to generate the predicted labels \( \hat{\mathbf{Y}} \). In traditional machine learning approaches, \( f_c(\cdot) \) might be a simpler, linear function or a basic classifier that operates on hand-crafted features extracted from the HSI data. However, in the context of deep learning, \( f_c(\cdot) \) is typically a deep neural network, such as a CNNs, which can automatically learn complex, hierarchical features directly from the raw hyperspectral data. This enables deep learning models to capture intricate patterns and spatial dependencies within the HSI cube, leading to significantly improved classification accuracy. Furthermore, deep learning models, such as 3D CNNs or hybrid CNN-RNN models, are particularly effective in HSIC because they can simultaneously exploit both the spectral and spatial information present in the HSI cube. By learning both local spatial features and global spectral features, these models are able to generate more robust and accurate predictions of the class labels for each pixel in the HSI cube \cite{2022:HSI:TraditionalToDeepModels:Review:IEEIJ}.
		\subsection{Convolutional Neural Networks} \cite{2023:HSISpatialEnhancementReview:IEEEJ}
		
\begin{figure}[!htbp]
			\centering
			\includegraphics[width=0.75\linewidth, keepaspectratio]{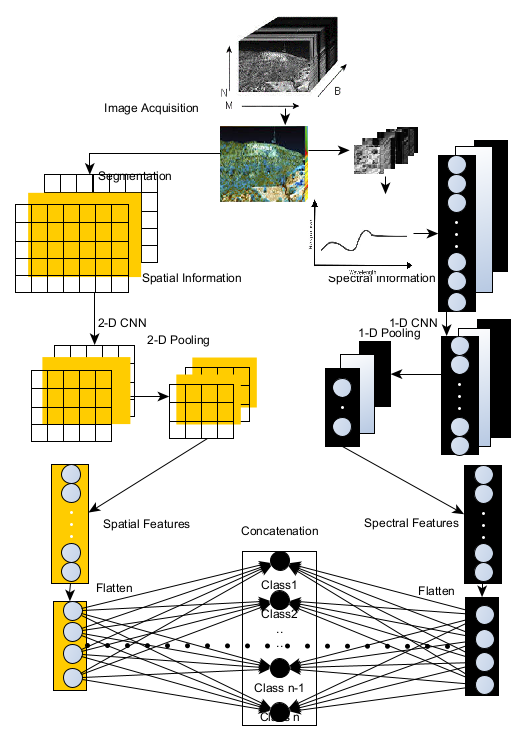}
			\caption{CNN}
			\label{fig:HSICNN}
		\end{figure}

		Deep learning techniques have shown remarkable success across many domains, with CNNs central to visual data processing. Inspired by biological neural structures, CNNs are multi-layered models that learn directly from raw image data to generate classification outputs. Over time, CNN architectures have been refined, becoming simpler and more efficient. With large datasets and GPU advancements, CNNs have surpassed traditional methods and sometimes human capabilities. They have proven effective in computer vision tasks like image classification, object detection, and facial recognition \cite{DCNN:HSI:Hu:2015}, as well as in fields like speech recognition. CNNs are crucial for interpreting visual data and delivering state-of-the-art results in image classification. In HSI image classification, CNNs have been integrated into frameworks, using architectures like stacked autoencoders (SAEs) to extract features and improve accuracy \cite{SAE:liu:2021:swinTransformer,yang:2023a:PixelDecomposition}. HSI's rich spectral data allows for more precise classification, but its high dimensionality and limited labeled data present challenges. Deep learning, particularly CNNs, provides an effective solution. Combining spectral and spatial information enhances classification, with 1D-CNN, 2D-CNN, 3D-CNN, and dual-channel CNNs proving effective. Table \ref{table:CNN_diff} outlines their differences and applications. While 3D-CNNs extract spectral-spatial features without dimensionality reduction, they may face overfitting and oversmoothing. Recent techniques like group feature extraction and channel shuffling address these issues, improving feature representation and reducing overfitting \cite{8803839}.	Liu et al. \cite{LIU:2BCNN:HSIClassification} proposed a CNN model for HSI classification, shown in 
		Figure \ref{fig:HSICNN}. It combines spatial and spectral feature extraction through two branches. The 2D convolutional branch treats the data cube as an image with 181 feature maps, extracting spatial features through convolution and pooling layers. The 1D convolutional branch processes the pixel spectrum, capturing spectral features with 1D convolution and pooling layers. After extracting features, they are flattened and concatenated into a unified representation, which is passed through a softmax output layer for classification. By omitting fully connected layers, the model reduces parameters, improving robustness. Batch normalization and dropout accelerate training and prevent overfitting, making the framework efficient and effective for HSI classification.
		
		Despite the fact that CNNs are the most popular deep learning models for HSI, they still face some \textbf{limitations}, as outlined below.
		\begin{enumerate}
			\item CNNs require vast amounts of labeled data for effective training, which can be costly and time-consuming to obtain for hyperspectral datasets, as domain-specific expertise is often needed for accurate labeling.
			\item Hyperspectral data typically consist of hundreds of spectral bands, leading to computational and memory challenges when processing such high-dimensional datasets.
			\item CNNs have complex architectures that require extensive parameter tuning, making them difficult to optimize without significant computational resources.
			\item The high computational and memory demands of CNNs make them less suitable for real-time applications or deployment in resource-constrained environments.
		\end{enumerate}
			
		\begin{table}[!htbp]
			\scriptsize
			\centering
			\caption{Differences between 1D, 2D, and 3D CNNs \cite{CNN:HSI:2021:Saeed}}
			\label{table:CNN_diff}
			\setlength{\tabcolsep}{0.001\textwidth} 
			\begin{tabular}{|p{0.5cm}|p{2.5cm}|p{2.5cm}|p{3cm}|} \hline
				\textbf{Type} & \textbf{Input Data} & \textbf{Convolution} & \textbf{Applications} \\ \hline
				1D CNN & 1D data (e.g., time series, spectral data) & Along one dimension & Speech recognition, spectral data analysis \\ \hline
				2D CNN & 2D data (e.g., images) & Across two dimensions (height, width) & Image processing, spatial feature extraction \\ \hline
				3D CNN & 3D data (e.g., videos, volumetric data) & Across three dimensions (height, width, depth) & Video analysis, medical imaging, spectral-spatial analysis \\ \hline
			\end{tabular}
		\end{table}
		
		\begin{table}[h]
			\caption{Applications of CNNs in HSI}
			\label{table:CNNs}
			\scriptsize
			\centering
			\setlength{\tabcolsep}{0.0025\textwidth}
			\begin{tabular}{|p{1.5cm}|p{6.25cm}|p{0.70cm}|}\hline
				\textbf{Application} & \textbf{Description} & \textbf{Ref.} \\ \hline
				Feature Extraction/Reduction & CNNs extract meaningful spatial and spectral features from high-dimensional hyperspectral data while reducing redundant information. This enhances efficiency and facilitates downstream tasks like classification. & \cite{3DCNN:FeatureExtraction:2020,HSIFeatureExtract:Quan:2020,CNNHSI:Giri:2023,CNNHSI:Yushi:2016} \\ \hline
				Classification & CNNs are widely used for pixel-wise and object-based classification by leveraging their ability to model spatial-spectral correlations in HSI data. & \cite{CNNHSI:Giri:2023,CNNHSI:Vaddi:2020,CNNHSI:Yushi:2016} \\ \hline
				Target Detection & CNNs detect specific targets in HSI by analyzing spatial and spectral patterns, enabling applications in remote sensing, surveillance, and mineral exploration. & \cite{CNNTargetDetection:Du:2018,CNNTargetDetection:Liu:2019,CNNTargetDetection:Freitas:2019,CNNTargetDetection:Zhao:2024} \\ \hline
				Semi-Supervised Learning & CNNs integrate labeled and unlabeled data effectively to improve performance in HSI tasks where labeled data is scarce. Techniques often incorporate GANs or pre-trained models. & \cite{HSICNNSemi:2018:Ling,HSICNNSemi:2022:Yao,GANsBasedHSI:Zhan:2023,HSICNNSemi:Kang:2019} \\ \hline
				Unmixing Task & CNNs decompose mixed HSI pixels into pure spectral signatures (endmembers) and their respective abundances, improving unmixing accuracy with spatial context awareness. & \cite{HSICNN:Unmixing:Zhang:2018,HSICNN:Elkholy:2020,HSI3DCNN:Unmixing:Zhao:2022} \\ \hline
				Denoising Strategies & CNN-based denoising techniques effectively remove noise from HSI data, preserving critical spectral and spatial features for analysis. & \cite{HSICNN:Denoising:Orhan:2024,HSICNN:Denoising:Nguyen:2020,HSICNN:Denoising:Maffei:2020} \\ \hline
				Feature Reduction & CNNs reduce hyperspectral data dimensions while retaining relevant information, easing computational burdens for further processing. & \cite{HSIDIMred:CNNLSTM:Tulapurkar:2020,CNNPCA:HSIDIMred:Xuefeng:2017} \\ \hline
				Spatial-Spectral Attention & CNNs equipped with attention mechanisms highlight critical spatial and spectral regions, improving accuracy in classification and target detection tasks. & \cite{HSIDIMAttention;CNN:Hang:2021} \\ \hline
				Fusion & CNNs fused with other models, such as attention networks or RNNs, result in better image classification. & \cite{CNNAttentionNewrk:HSIFusuion:Pan,CNNRNNFusion2018m3fusion} \\ \hline
			\end{tabular}
		\end{table}
		
		\subsection{Stacked Auto-Encoder}
\begin{figure}[!htbp]
			\centering
			\includegraphics[width=0.75\linewidth, keepaspectratio]{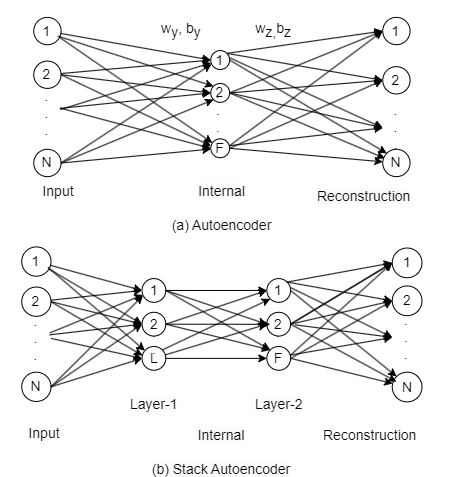}
			\caption{Autoencoder and Stacked Autoencoder}
			\label{fig:SAEncoders}
		\end{figure}

		Autoencoders, in the context of HSI and multimodal HSI, are feed-forward neural networks designed to replicate input data at the output layer. Unlike typical neural networks that focus on classification, autoencoders aim to encode and then reconstruct input data as accurately as possible. An autoencoder has input and output layers of the same size. It compresses the input data into a latent (hidden) representation, which it then uses to reconstruct the output. The encoder transforms the input into the latent representation using \(
		l = f(w_l x + b_l), \quad m = f(w_m y + b_m)\), where \( w_l \) and \( w_m \) are weights, and \( b_l \) and \( b_m \) are biases. The goal is to minimize the error between the input \( x \) and the reconstructed output \( z \) using \(\text{arg min}_{w_l, w_m, b_l, b_m} \left[ \text{error}(l, m) \right]\). The decoder reconstructs the output based on the latent representation, using weights \( W_a \) and \( W_b \). If the hidden layer is smaller than the input layer, the autoencoder compresses the data, focusing on the most important features for reconstruction. The loss function compares the input \( x_k \) and reconstructed data \( z_k \) over all dimensions \( d \), with regularization: \(C(x, z) = - \frac{1}{m} \sum_{k=1}^{d} \left( (x_k - z_k)^2 + (1 - x_{ik}) \right) + \frac{\beta}{2} W^2\). Here, \( m \) is the number of samples, \( x \) and \( z \) are the input and reconstructed data, and \( W \) is the weight matrix with regularization. The basic graphical representation of encoders and stacked encoders is shown in Figure \ref{fig:SAEncoders}. In HSI and multimodal HSI, autoencoders help extract meaningful features from large, complex datasets for tasks like classification or dimensionality reduction. In addition to the uses given in Table \ref{table:Autoencoders}, they are good at compression and shipment of compressed images from source to destination. SAEs have some data processing \textbf{challenges}, which are listed below.
		
		\begin{enumerate}
			\item Choosing the right architecture and hyperparameters is critical, as these directly influence the quality of the learned features and the overall performance of the model.
			\item SAEs require significant computational resources for both training and inference, making them difficult to deploy in real-time applications or environments with limited resources.
			\item The complex hierarchical representations learned by SAEs are often difficult to interpret, which poses challenges in hyperspectral data analysis, especially in scientific and regulatory applications where transparency is essential.
		\end{enumerate}
		
		\begin{table}[h]
			\caption{Autoencoders}
			\label{table:Autoencoders}
			\scriptsize
			\centering
			\setlength{\tabcolsep}{0.0025\textwidth}
			\begin{tabular}{|p{2cm}|p{5.75cm}|p{0.70cm}|}\hline
				\textbf{Application} & \textbf{Description} & \textbf{Ref.} \\ \hline
				Feature Extraction & SAEs encode input data into a lower-dimensional latent space and decode it back, preserving spectral and spatial information. & \cite{SAE:lin:2013:spectral} \\ \hline
				Classification & SAEs facilitate data compression for easier analysis, especially for classification tasks, through unsupervised feature learning. & \cite{SAE:xie:2019:srun,SAE:han:2014:objectdetection} \\ \hline
				Target Detection & Utilizing learned features for improved target detection in hyperspectral data. & \cite{SAE:xie:2019:srun, SAE:han:2014:objectdetection} \\ \hline
				Anomaly Detection & Detects anomalies by modeling dependencies in HSI data. & \cite{hu2021hyperspectral} \\ \hline
				Semi-Supervised Learning & Pre-training layers unsupervised, then fine-tuning with limited labeled data to enhance accuracy. & \cite{SAE:xing:2016:LowLables} \\ \hline
				Unmixing Task & Decomposing pixel spectra into spectral signatures (endmembers) and corresponding abundances. & \cite{yang:2023a:PixelDecomposition} \\ \hline
				Denoising Strategies & Using denoising autoencoders (DAE) and stacked denoising autoencoders (SDAE) for robust feature extraction. & \cite{SAE:vincent:2008:Denoising,SAE:deng:2023:Denoising} \\ \hline
				Compact Features & Framework (CDSAE) for extracting low-dimensional features and enhancing classification performance. & \cite{SAE:zhou:2019:LowDimensionalFeatureExt} \\ \hline
				Two-Stage SAEs & A two-stage approach for better spectral-only and spectral-spatial representations using hybrid autoencoders. & \cite{SAE:bai:2024:twoStageDimensionRed} \\ \hline
				Spatial-Spectral Attention & Attention mechanisms to effectively extract abundance maps and endmembers from hyperspectral data. & \cite{SAE:liu:2021:swinTransformer,yang:2023a:PixelDecomposition} \\ \hline
			\end{tabular}
		\end{table}
		\subsection{Deep Belief Networks (DBNs)}
\begin{figure}[!htbp]
			\centering
			\includegraphics[width=0.75\linewidth, keepaspectratio]{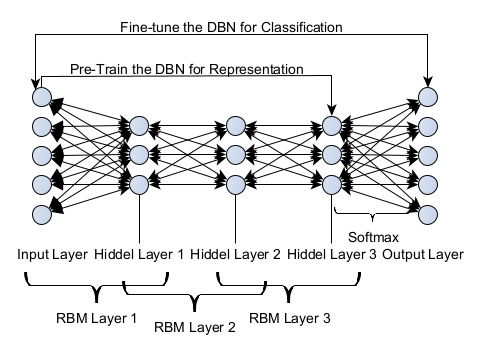}
			\caption{DBN}
			\label{fig:HSIDBN}
		\end{figure}

		DBNs consist of multiple layers of neurons, where each layer is only connected to the next, forming a structure similar to multilayer perceptrons (MLPs) but without intra-layer connections. Each layer trains independently, learning features from the output of the previous layer, allowing DBNs to capture hierarchical and abstract patterns from data. The key building block of DBNs is the Restricted Boltzmann Machine (RBM), which has a visible layer interacting with input data and a hidden layer that learns features. As can be seen from Figure \ref{fig:HSIDBN}, DBNs are formed by stacking RBMs, where the hidden layer of one serves as the visible layer of the next, enabling deeper feature extraction \cite{DLTechniquesforHSIAgri:2024:Guerri}. Training DBNs involves a two-step process. First, an unsupervised pre-training phase initializes the weights by training each RBM layer using the Contrastive Divergence algorithm. This allows the model to learn underlying data representations efficiently. The second phase involves supervised fine-tuning using labeled data, optimizing the DBN for specific tasks such as classification or regression. DBNs excel in tasks like HSI by automatically extracting complex spatial and spectral features from high-dimensional data, enhancing the accuracy of classification, target detection, and anomaly identification, as shown in Table \ref{table:DBN_HSI}.
		
		\begin{table}[!htpb]
			\caption{Applications of DBN in HSI}
			\label{table:DBN_HSI}
			\centering
			\setlength{\tabcolsep}{0.0025\textwidth}	
			\scriptsize
			\begin{tabular}{|p{1.75cm}|p{5.75cm}|p{0.75cm}|}\hline
				\textbf{Application} & \textbf{Description} & \textbf{Ref.} \\ \hline
				Feature Extraction & Learns hierarchical feature representations for HSI classification and analysis. & \cite{HSI:DBN:FExtractAndClass:2019:Mughees,DLTechniquesforHSIAgri:2024:Guerri} \\ \hline
				Classification & Dominates in spectral-spatial HSI classification by leveraging unsupervised pretraining for better accuracy. & \cite{HSI:DBN:FExtractAndClass:2019:Mughees,HSI:DBN:Class:Kumar:2024,HSI:DBN:Class:Shenming:2022} \\ \hline
				Dimensionality Reduction & Reduces spectral dimensions by learning compact feature representations without losing critical information. & \cite{HSI:DIMRed:DBN:Arsa:2016} \\ \hline
				Spectral Unmixing & Estimates material abundances and unmixes spectral components using its ability to model complex distributions. & \cite{HSI:PixelUnmix:DBN:Deng:2019} \\ \hline
				Data Fusion & Integrates spectral and spatial information for improved HSI analysis and target detection. & \cite{HSI:featureFusion:DBN:Ghassemi:2018} \\ \hline
				Anomaly Detection & Identifies anomalies by learning probabilistic spectral patterns in unsupervised settings. & \cite{HSI:TD:DBN:Ma2017ADB,HSI:TD:DBN:Liu:2018} \\ \hline
				Noise Reduction & Removes noise or artifacts in HSI images by reconstructing clean spectral bands. & \cite{HSI:Denoising:MOHAN:2024} \\ \hline
				Land Cover Mapping & Performs segmentation and classification of land cover types using deep spectral-spatial modeling. & \cite{HSI:LandCover:DBN:Ayhan2017} \\ \hline
			\end{tabular}
		\end{table}
		
		But the use of DBNs in HSI is not free from some \textbf{challenges}, which are listed below.
		
	\begin{enumerate}
		\item DBNs for large hyperspectral datasets requires significant computational resources, making them unsuitable for real-time or resource-constrained applications.
		\item DBNs often lack transparency, complicating the interpretability of decision-making, especially in critical applications where understanding model behavior is crucial.
		\item The success of DBNs heavily depends on selecting appropriate hyperparameters and architecture, a time-consuming task. Additionally, random weight initialization can hinder training efficiency and performance, though advanced techniques like layered approaches are being explored to mitigate this.
		\item While DBNs are proficient in feature extraction, extracting discriminative features from hyperspectral data requires advanced strategies, such as graph-based penalties to capture manifold structures. Moreover, reducing computational time remains a challenge, although recent improvements show progress.
	\end{enumerate}
	
		\subsection{Generative Adversarial Networks (GANs)}
\begin{figure}[!htbp]
			\centering
			\includegraphics[width=0.75\linewidth, keepaspectratio]{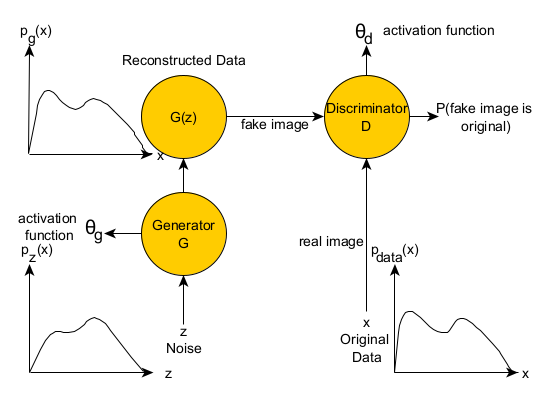}
			\caption{GANs}
			\label{fig:HSIGANs}
		\end{figure}

		A GAN consists of two components: the generator \( G \) and the discriminator \( D \). The generator \( G \) captures the potential distribution of real data and generates new data, while the discriminator \( D \) is a binary classifier that evaluates whether the input samples are real or generated (fake). The information flow in a GAN involves a feed-forward process: the generator \( G \) produces fake data, and the discriminator \( D \) evaluates the authenticity of this data \cite{HSIGAN:2018:Zhu}. The architecture of a GAN is shown in Figure \ref{fig:HSIGANs}. To learn the generator’s distribution \( p_g(x) \) over data \( x \), we assume that real samples are drawn from a true data distribution \( p(x) \), and the input noise variable \( z \) follows a prior distribution \( p(z) \). The generator takes random noise \( z \) as input and maps it to the data space as \( G(z) \). The discriminator \( D(x) \) estimates the probability that \( x \) is a real sample from the training data. During optimization, the discriminator \( D \) is trained to maximize \( \log(D(x)) \), which is the probability of assigning the correct label to real and fake samples. Meanwhile, the generator \( G \) is trained to minimize \( \log(1 - D(G(z))) \), where the goal is to generate samples that the discriminator classifies as real. This results in a minimax optimization problem as shown in Equation \ref{Eq:GANMinMaxProb}, where \( \mathbb{E} \) denotes the expectation operator.
		\begin{equation}\label{Eq:GANMinMaxProb}
			\scriptsize
			\min_G \max_D V(D, G) = \mathbb{E}_{x \sim p(x)}[\log D(x)] + \mathbb{E}_{z \sim p(z)}[\log(1 - D(G(z)))]
		\end{equation}
		
		One of the challenges in training GANs is that when the discriminator becomes too good at distinguishing real from fake data, the gradient updates for the generator can vanish, causing the training process to stall. To address this, the generator’s loss function is typically modified to maximize the probability of classifying a sample as real, rather than minimizing the probability of classifying it as fake. The modified objective is:
		
		\[
		\min_G V_F(D, G) = \mathbb{E}_{z \sim p(z)}[\log D(G(z))]
		\]
		
		The generator \( G \) updates its parameters based on the backpropagation of the discriminator's gradient, rather than directly using real data samples. There is also no constraint on the dimensionality of the latent code or on the generator network being invertible. GANs can learn models that generate points on a thin manifold close to the real data distribution, enabling the training of various types of generator networks \cite{gu2021multimodal}. The key applications of GANs are listed in Table \ref{table:GAN_HSI}.
		
		\begin{table}[!htpb]
			\caption{Applications of GANs in HSI}
			\label{table:GAN_HSI}
			\centering
			\scriptsize
			\setlength{\tabcolsep}{0.0025\textwidth}
			\begin{tabular}{|p{2cm}|p{5.5cm}|p{0.75cm}|}\hline
				\textbf{Application} & \textbf{Description} & \textbf{Ref.} \\ \hline
				Data Augmentation & Generates synthetic HSI data to address limited labeled data and improve model training.& \cite{HSI:DataAug:GANs:zhang:2022,HSI:DataAugm:GANs:Zhang:2023} \\ \hline
				Classification & Enhances HSI classification by generating realistic spectral-spatial features for imbalanced datasets.& \cite{HSI:Class:GANS:Sireesha:2022,HSI:Class:GANS:Tayeb:2020} \\ \hline
				Spectral Unmixing & Learns the nonlinear mixing process and generates endmember distributions for accurate unmixing.& \cite{HSI:Unmix:Suresh:2023,HSI:Unmix:Tang:2020} \\ \hline
				Anomaly Detection & Detects anomalies by modeling the spectral data distribution and identifying outliers. &\cite{HSI:AnomalyDetect:GANs:Wang:2023,HSI:AnomalyDetect:GANs:Arisoy:2021,HSI:AnomalyDetect:GANs:Xia:2022,HSI:DomainAdaptation:Bejiga:2018} \\ \hline
				Super-Resolution & Reconstructs high-resolution HSI from low-resolution inputs using adversarial training.&\cite{HSI:SuprRes:GANs:shi:2022} \\ \hline
				Data Fusion & Combines data from multiple hyperspectral sensors or integrates HSI with multispectral data for enhanced analysis. & \cite{HSI:DataFusion:GANs:Xiao:2021,HSI:DataFusion:GANs:Liu:2022} \\ \hline
				Noise Reduction & Removes noise and artifacts in HSI by learning the mapping between noisy and clean images.&\cite{HSI:Denoise:GANs:zhang:2022} \\ \hline
				Domain Adaptation & Aligns spectral data from different domains to improve cross-domain HSI classification.&\cite{HSI:DomainAdaptation:Zhen:2022,HSI:DomainAdaptation:Feng:2024,HSI:DomainAdaptation:Sun:2020,HSI:DomainAdaptation:Chen:2020} \\ \hline
				Change Detection & Learns temporal spectral-spatial changes to identify significant environmental or land-use transitions.&\cite{HSI:CD:GANs:deecke:2019} \\ \hline
				Spectral Reconstruction&Reconstructs missing or corrupted spectral bands in HSI data using adversarial training.&\cite{HSI:Reconstruct:GANs:Eek:2019} \\ \hline
			\end{tabular}
		\end{table}
		
		No doubt, GANs have gained significant attention from the research community to generate realistic images, making them valuable for various modern applications, including image synthesis and domain adaptation. In terms HSI, GANs have shown promise in tasks such as data augmentation, super-resolution, and domain translation, but pose some challenges such as:
		\begin{enumerate}
			\item Training GANs remains challenging due to issues like mode collapse, non-convergence, and stability concerns. Addressing these challenges requires designing efficient models with appropriate network architectures, selecting suitable objective functions, or employing effective optimization techniques. Despite the introduction of different GAN variants with diverse characteristics, several issues remain unsolved yet, which require further research and innovation in field \cite{GANSSurvey:Saxena:2022}.
		\end{enumerate}

		\subsection {Recurrent Neural Networks (RNNs)}
\begin{figure}[!htbp]
			\centering
			\includegraphics[width=0.5\linewidth, keepaspectratio]{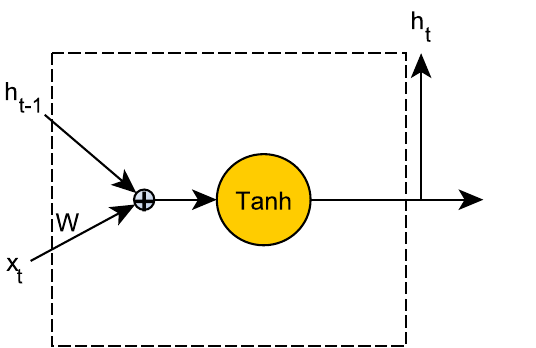}
			\caption{RNN}
			\label{fig:RNN}
		\end{figure}

RNNs have become an effective tool for HSI, offering a unique approach to analyzing spectral data cubes. Unlike standard neural networks, RNNs process sequential data, making them ideal for hyperspectral time-series analysis. By capturing temporal dependencies, RNNs excel in tasks like change detection and object tracking, tracking subtle variations across spectral bands over time. They model the sequential nature of hyperspectral data acquisition, enabling dynamic scene analysis. RNNs also denoise and restore hyperspectral data by leveraging spectral correlations, reducing noise and compensating for sensor or atmospheric interference. A typical RNN architecture, shown in Figure \ref{fig:RNN}, processes the input data cube by treating each pixel spectrum as a sequence, where recurrent connections maintain memory across time steps. This allows RNNs to handle the complex, high-dimensional nature of hyperspectral data, making them vital for remote sensing applications.
	
		RNNs can effectively handle sequence learning by adding recurrent connections, allowing neurons to link with past outputs over time. Given an input sequence $\{x_1, x_2, \dots, x_T\}$ and hidden states $\{h_1, h_2, \dots, h_T\}$, at time $t$, the node processes input $x_t$ and the previous state $h_{t-1}$, producing:
		
		\[
		h_t = \sigma( W_{hx}x_t + W_{hh}h_{t-1} + b )
		\]
		
		where $W_{hx}$ is the weight between the input and hidden state, $W_{hh}$ is the recurrent weight, $b$ is the bias, and $\sigma$ is the activation function.
		
		During training, RNNs encounter two significant challenges known as the vanishing gradient and exploding gradient problems. The vanishing gradient problem occurs when the weights \(W_{hh}\) are small (\(|W_{hh}| < 1\)), causing the gradients of earlier hidden states to diminish as they are backpropagated through the network. This results in very small updates to the weights associated with those states, making it hard for the model to learn from them. Consequently, RNNs struggle to capture long-term dependencies, as earlier inputs have little influence on the output. The exploding gradient problem occurs if the weights \(W_{hh}\) are large (\(|W_{hh}| > 1\)), causing the gradients to grow exponentially as they are backpropagated through time. This can cause the weights to become excessively large, leading to numerical instability and making training difficult. The model may produce erratic outputs or fail to converge. Together, these problems hinder the ability of RNNs to learn effectively over long sequences, limiting their performance in tasks requiring long-term context \cite{HSI:LSTM:Zhou:2019}.
		
		\subsection{Long Short-Term Memory (LSTM), Gated Recurrent Units (GRUs)}
\begin{figure}[!htbp]
			\centering
			\includegraphics[width=0.75\linewidth, keepaspectratio]{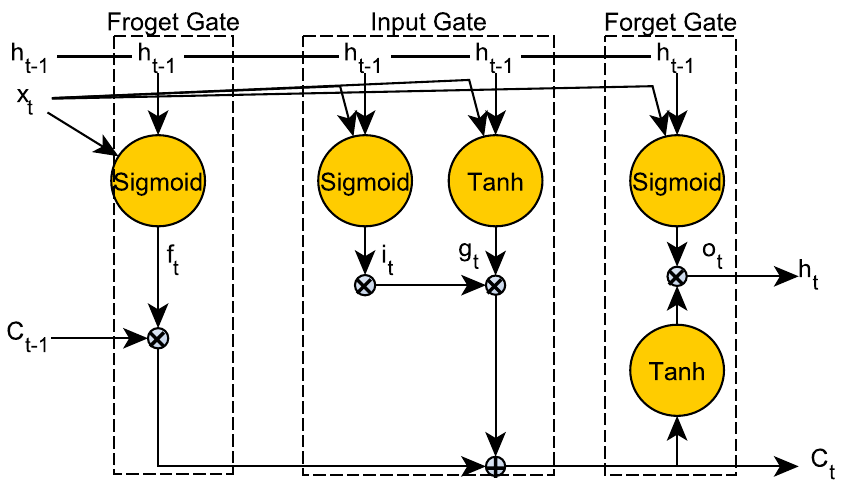}
			\caption{LSTM}
			\label{fig:HSILSTM}
		\end{figure}

\begin{figure}[!htbp]
			\centering
			\includegraphics[width=0.75\linewidth, keepaspectratio]{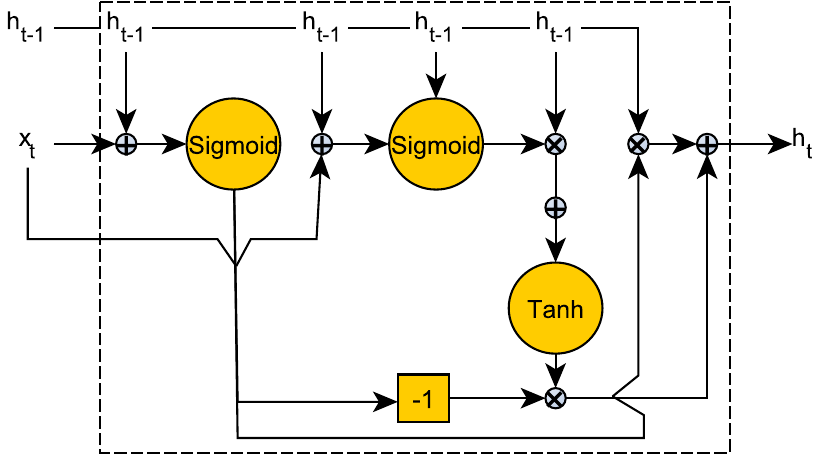}
			\caption{GRU}
			\label{fig:HSIGRU}
		\end{figure}

Hyperspectral data exhibit strong inter-band and long-term dependencies, making them suitable for sequential modeling with RNNs. However, RNNs struggle with vanishing gradients, limiting their ability to capture these dependencies. GRUs address this with a simplified structure that improves computational efficiency while maintaining sequential modeling capabilities. GRUs use an update gate to control past information retention and a reset gate to decide what to forget. Unlike LSTMs, which separate memory cells and hidden states, GRUs combine them \cite{GRUvsLSTM:Pudikov:2020}. Basic architectures of LSTM and GRU are shown in Figures \ref{fig:HSILSTM} and \ref{fig:HSIGRU}. In hyperspectral imaging, GRUs are used for spectral-temporal analysis, noise filtering, dimensionality reduction, and real-time tracking. They offer faster training and fewer parameters than LSTMs but may not perform as well on complex sequences and require empirical validation \cite{HSI:LSTM:Zhou:2019, RNNLSTMTransforer:Viel:2023}.
		
		\begin{table*}[!htpb]
			\caption{Applications of RNN, GRU, and LSTM in HSI}
			\label{table:RNN_GRU_LSTM_HSI}
			\scriptsize
			\centering
			\setlength{\tabcolsep}{0.0025\textwidth}
			\begin{tabular}{|p{4.75cm}|p{10cm}|p{2.75cm}|}\hline
				\textbf{Application} & \textbf{Description} & \textbf{Ref.} \\ \hline
				\multicolumn{3}{|c|}{\textbf{RNN Applications in HSI}} \\ \hline
				Classification (Spectral, Image) & Captures sequential spectral information to classify hyperspectral pixels. & \cite{DRNN:HSIImgClass:Hao2021,8082108,cascadRNN:HSIImgClass:Hang2019,RNNLSTMTransformer:HSI:2023:Viel} \\ \hline
				Change Detection & Analyzes temporal HSI sequences to detect land use or environmental changes over time. & \cite{ChangeDetect:RNNFCGRULSTN:Mou2019} \\ \hline
				Spectral Unmixing & Estimates material abundances in mixed pixels by learning nonlinear spectral relationships. & \cite{RNN:HSIUnMix:Lei:2020} \\ \hline
				Anomaly Detection & Identifies deviations in spectral sequences to detect anomalous patterns in time series data. & \cite{Nanduri2016} \\ \hline
				Fusion & RNNs fused with other models such as CNN result in better image classification. & \cite{CNNRNNFusion2018m3fusion} \\ \hline
				Land Cover Mapping & Segments and classifies land cover types using spatial-spectral relationships. & \cite{LSTMRNN:HSILandCover:Dino:2017,RNN:HSILandCover:Ananta:2024,Handson:HSI:LandCover:kakarla:2020,RNN:HSILandCover:Tang:2022} \\ \hline
				\multicolumn{3}{|c|}{\textbf{GRU Applications in HSI}} \\ \hline
				Spectral-Spatial Feature Extraction & Efficiently captures spatial and spectral features with reduced computational complexity. & \cite{HSIFeatureExtract:GRU:Wu:2023} \\ \hline
				HSI Image Classification & Classifies hyperspectral pixels based on spatial and spectral sequences. & \cite{HSIClassification:RNNFCGRULSTN:Haowen:2018,GRU:HSIClass:Pan2020} \\ \hline
				Change/Target Detection & Learns dynamic spectral patterns to detect changes, specific targets, or materials. & \cite{ChangeDetect:RNNFCGRULSTN:Mou2019} \\ \hline
				Data Fusion & Integrates data from multiple hyperspectral sensors or sources. & \cite{GRU:HSI:Fusion:2021:Xiao} \\ \hline
				Dimensionality Reduction & Reduces HSI data dimensions while preserving essential spectral information for clustering tasks. & \cite{GRU:HSIDimRed:Pande:2021} \\ \hline
				\multicolumn{3}{|c|}{\textbf{LSTM Applications in HSI}} \\ \hline
				Time-Series Analysis & Models long-term dependencies in temporal HSI data for crop monitoring or environmental changes. & \cite{HSI:TimeSeriesAnal:LSTM:Jiang:2018} \\ \hline
				HSI Image Classification & Classifies hyperspectral pixels based on spatial and spectral sequences. & \cite{8082108,RNNLSTMTransformer:HSI:2023:Viel,HSIClassification:RNNFCGRULSTN:Haowen:2018} \\ \hline
				Change/Target Detection & Learns dynamic spectral patterns to detect changes, specific targets, or materials. & \cite{ChangeDetect:RNNFCGRULSTN:Mou2019} \\ \hline
				Spectral-Spatial Feature Extraction & Efficiently captures spatial and spectral features with reduced computational complexity. & \cite{SpatialSpectralFeatureExtraction:Hu:202:Hu, LSTM:HSIFeaftureExtract:Ma2021,LSTM:HSIAnomalyDetect:2021:Tongbin} \\ \hline
				Spectral Anomaly Detection & Detects spectral anomalies by modeling long-range dependencies in HSI data. & \cite{LSTM:HSIAnomalyDetect:2022:Zhi,LSTM:HSIAnomalyDetect:2021:Tongbin} \\ \hline
				Land Cover Mapping & Segments and classifies land cover types using spatial-spectral relationships. & \cite{LSTM:HSILandCover:Tejasree:2024,LSTMRNN:HSILandCover:Dino:2017,Handson:HSI:LandCover:kakarla:2020,LSTM:HSILandCover:Zhu:2021} \\ \hline
				Noise Reduction, Compression, Reconstruction & Removes noise or reconstructs missing spectral bands in HSI data. & \cite{DLTechniquesforHSIAgri:2024:Guerri,HSICompress:LSTM:Webster:2022,LSTM:HSINoiseRedRecons:Wang:2023} \\ \hline
				Endmember Extraction and Spectral Unmixing & Extracts endmember spectra and estimates fractional abundances in hyperspectral pixels. & \cite{CNN:LSTM:Unmixing:HSI:Kumar:2024} \\ \hline
			\end{tabular}
		\end{table*}
		RNNs are well-suited for modeling simpler sequential dependencies but often face challenges, such as vanishing gradient issues, when dealing with long sequences. GRUs address some of these limitations by reducing computational complexity, making them more efficient than LSTMs for smaller datasets or less complex sequences. LSTM networks, on the other hand, are highly effective at handling long-term dependencies and are widely preferred for complex HSI tasks involving lengthy sequences. The key applications of RNNs, GRUs, and LSTMs are summarized in Table \ref{table:RNN_GRU_LSTM_HSI}. These architectures are frequently combined with CNNs to integrate spatial features with sequential dependencies, thereby enhancing performance in HSI analysis.\par
		RNN, LSTM, and GRU architectures in hyperspectral imaging (HSI) face some challenges given below, which necessitate futher research for developing develop scalable solutions for HSI applications.
		\begin{enumerate}
			\item Traditional RNNs struggle with long-term dependencies, whereas LSTMs and GRUs mitigate this through memory mechanisms, improving spectral-temporal feature extraction.
			\item Hybrid models, such as LSTMs combined with CNNs, enhance HSI analysis, while wavelet transforms aid in multi-scale feature learning. Encoder-decoder frameworks and ensemble learning further refine performance. Despite these advancements in tasks like classification and anomaly detection, challenges remain in training data requirements, computational efficiency, and model interpretability \cite{RNNLSTGRU:WAQAS:2024}.
		\end{enumerate}
			
		\subsection{Transformers in HSI} 
		Transformers, a highly versatile and powerful architecture, have become increasingly prominent in HSI due to their ability to handle sequential and grid-like data. At the heart of the transformer architecture lies the {self-attention mechanism}, which dynamically weighs the importance of different input components. This is particularly advantageous in HSI, as it enables the model to identify and emphasize meaningful spectral-spatial correlations across bands. Mathematically, the attention mechanism is expressed using Equation \ref{eq:TransformerAttentionMech} \cite{HSI:Transformer:Zhang:2023}.
		\begin{equation}\label{eq:TransformerAttentionMech}
			\text{Attention}(Q, K, V) = \text{softmax}\left(\frac{QK^T}{\sqrt{d_k}}\right)V
		\end{equation}
		
		where \( Q \), \( K \), and \( V \) represent the query, key, and value projections of the input data, and \( d_k \) is the dimensionality of the key vector. The scaling factor \( \frac{1}{\sqrt{d_k}} \) ensures numerical stability during gradient computation. To further enhance its capability, the transformer architecture employs {multi-head attention}, which allows the model to simultaneously focus on multiple aspects of the input. This mechanism is defined as:
		\(
		\text{MH}(Q, K, V) = \text{Concat}(\text{head}_1, \ldots, \text{head}_h)W^O
		\)
		where each attention head is computed as:
		\(
		\text{head}_i = \text{Attention}(QW_i^Q, KW_i^K, VW_i^V)
		\)
		Here, \( W_i^Q \), \( W_i^K \), \( W_i^V \), and \( W^O \) are learnable projection matrices, enabling the model to capture diverse relationships within the hyperspectral data. Transformers also integrate feedforward networks with residual connections and layer normalization, contributing to model stability and learning efficiency. The feedforward network is defined as:
		\(
		\text{FFN}(x) = \max(0, xW_1 + b_1)W_2 + b_2
		\)
		where \( W_1, W_2 \) are weight matrices, and \( b_1, b_2 \) are biases. Layer normalization is applied as:
		\[
		\text{LayerNorm}(x) = \gamma \frac{x - \mu}{\sqrt{\sigma^2 + \epsilon}} + \beta
		\]
		where \( \mu \) and \( \sigma \) denote the mean and standard deviation of \( x \), while \( \gamma \) and \( \beta \) are trainable scale and shift parameters.
		
These mechanisms make transformers highly effective for spectral-spatial classification in HSI, where both spectral signatures and spatial context are crucial. For example, Vision Transformers \cite{VisionTranformer:Scheibenreif:2023} adapt by dividing the input into patches, using self-attention, and incorporating positional embeddings to preserve spatial relationships. Hybrid architectures combining transformers with CNNs or RNNs are also common, leveraging CNNs’ feature extraction or RNNs’ sequential modeling alongside the transformer’s dynamic attention. Transformers have advantages in HSI, with the self-attention mechanism allowing the model to focus on the most relevant spectral-spatial regions, improving interpretability and accuracy. Their versatility makes them suitable for various HSI tasks, such as classification, segmentation, and anomaly detection, as shown in Table \ref{Table:Transformers:HSI:Aps}.\par

		Designing hyperspectral-specific transformer models presents a few challenges as listed below:
		
		\begin{enumerate}
			\item Methods like Transformer Architecture Search (TAS), such as AutoFormer and ViTAS (Vision Transformer Architecture Search), streamline model design but require substantial computational power, with AutoFormer it takes up to 24 GPU days. 
			\item Zero-cost proxies, which evaluate transformer architectures without training, offer a faster alternative providing: i) efficiency that enables the identification of high-performing models in minutes, and ii) data independence as many proxies work without real datasets. They reduce the cost and complexity of data collection and calibration, but their accuracy in evaluating hyperspectral-specific architectures remains uncertain and requires further research to refine transformers for hyperspectral imaging applications \cite{HyTAS:HSITransformer:2024}.
		\end{enumerate}
		
		\begin{table}[h]
			\caption{Applications of Transformers in HSI}
			\label{Table:Transformers:HSI:Aps}
			\scriptsize
			\centering
			\setlength{\tabcolsep}{0.0025\textwidth}
			\begin{tabular}{|p{1.5cm}|p{6cm}|p{0.75cm}|}\hline
				\textbf{Application} & \textbf{Description} & \textbf{Ref.} \\ \hline
				Feature Extraction, Dimensionality Reduction & Transformers use self-attention to extract long-range spatial and spectral features, providing more accurate representations than traditional methods. & \cite{Transformers:HSIFusion:Dang:2023,Transformers:HSI:DimRed:Shi:2024} \\ \hline
				Super-resolution & Transformers like ESSAformer use spectral and spatial attention mechanisms to enhance HSI image resolution, improving detail and accuracy for downstream tasks. & \cite{Transformers:HSISuperResolution:Mingjin:2023,Transformers:FusionSuperRes:Qiao:2023} \\ \hline
				Classification & Vision Transformers (ViTs) and spectral-spatial transformers are effective for pixel-level and global classification, capturing multi-scale spatial and spectral correlations. & \cite{Transformers:HSIClass:Zhang:2023,Transformers:HSIClass:Yang:2022,Transformers:HSIClass:Arshad:2024} \\ \hline
				Spatial-Spectral Attention & Transformers dynamically weigh spatial and spectral regions, improving performance in classification, segmentation, and detection tasks. & \cite{Transformers:HSIDenoising:Chen:2024,Transformers:HSIDenoising:Li:2023,Transformers:HSIDenoising:Zhang:2024} \\ \hline
				Unmixing & Transformers model spatial-spectral dependencies, enhancing the accuracy of hyperspectral unmixing by estimating pure spectral signatures (endmembers) and their abundances. & \cite{Transformers:HSI:UnMix:Ghosh:2022} \\ \hline
				Denoising & Self-attention in transformers helps denoise HSI images while preserving key spectral-spatial features for downstream tasks. & \cite{Transformers:HSIDenoising:Chen:2024,Transformers:HSIDenoising:Li:2023,Transformers:HSIDenoising:Zhang:2024} \\ \hline
				Image reconstruction & Sparse transformers extract crucial spatial-spectral features during HSI image reconstruction, aiding in high-quality image restoration with reduced redundancy. & \cite{Transformers:HSIRecons:Cai:2022,Transformers:HSIRecons:Miaoyu:2024} \\ \hline
				Change Detection & Transformers detect changes in HSI image sequences by leveraging spectral-spatial context, aiding environmental monitoring and disaster assessment. & \cite{Transformers:HSICD:Wang:2024,Transformers:HSICD:Wang:2022,Transformers:HSIDenoising:Fu:2024} \\ \hline
				Fusion & Combining transformers with other architectures (e.g., CNNs, RNNs) enhances spatial-spectral representation, improving performance in classification and target detection. & \cite{Transformers:HSIFusion:Dang:2023,Transformers:FusionSuperRes:Qiao:2023} \\ \hline
				Target Detection & Transformers excel at identifying targets in hyperspectral imagery by analyzing subtle spectral and spatial patterns, benefiting remote sensing and surveillance. & \cite{Transformers:HSI:TD:Rao:2022,Transformers:HSI:TD:Zhao:2024} \\ \hline
			\end{tabular}
		\end{table}
		
			
		\section{Multimodal HSI, DL-based Fusion, Challenges, and Integration of LLMs}\label{sec:MultimodalHSIFusionChallengesLLMs}
		Multimodal hyperspectral imaging (MHSI) overcomes the limitations of traditional hyperspectral systems by integrating other data modalities such as LiDAR, RGB, RADAR (Radio Detection and Ranging), and Thermal. Traditional HSI lacks temporal and spatial flexibility, capturing data at a single point in time and space. MHSI enhances hyperspectral technology by incorporating temporal and spatial dimensions into spectral data. It focuses on dynamic hyperspectral imaging and hyperspectral stereo imaging, which extend time and spatial information. Dynamic hyperspectral imaging includes hyperspectral video and multitemporal imaging, capturing time-varying scene data. Hyperspectral stereo imaging collects 3D spatial and spectral data, preserving elevation information that standard HSI might lose. Both techniques are designed for diverse remote sensing applications \cite{gu2021multimodal}. HSI's limited altitude flexibility makes it hard to detect occluded objects or those at varying altitudes, like grass on the ground or rooftops. To capture altitude information or identify hidden objects, LiDAR \cite{huo2018supervised} or SAR are integrated with HSI. This integration has garnered increasing attention in remote sensing research \cite{huo2023multimodal}.

		Remote sensing (RS) technologies are responsible for observing various aspects of the Earth's surface. These include the identification of constituent materials, spatial arrangement of objects in a specified region, ground layer composition, and surface characteristics. The information acquired from remote sensing is usually based on one or a few of the features or modalities mentioned earlier. To achieve an in-depth understanding of the scene, information from multiple sources can be merged using data fusion (DF) techniques for combined analyses. The rapid advancement of sensor technology has made it easier to obtain multimodal data for a variety of uses, especially in RS. Many types of data, such as LiDAR, multispectral imagery, and hyperspectral imagery, are now available. The idea of merging these data sources has become increasingly significant due to its enhanced ability to record detailed information \cite{zhang2010multi, dalla2015challenges}. 
		\subsection{Fusion}
		Fusion is a term used to describe the collaborative use of several remote sensing systems to provide an accurate and comprehensive description of Earth's surface. Fusion is now considered to be a common scenario in remote sensing applications. Large-scale horizontal land-cover types have been widely mapped using passive optical sensors, but these sensors have the drawback of only working properly in daylight and clear weather conditions. To overcome these limitations, SAR systems are used. SAR can work at any time in any weather condition and captures unique responses from man-made targets and terrain. Recently, LiDAR has been validated for its ability to measure the vertical height of structures by estimating the delays of transmitted pulses. Further, it provides information about the reflective properties of materials by detecting the intensity of returned signals. However, it has high running costs, making it expensive to operate. 
		
\begin{figure}[!htbp]
				\centering
				\includegraphics[width=\linewidth, trim=1in 0in 1in 0in, clip, keepaspectratio]{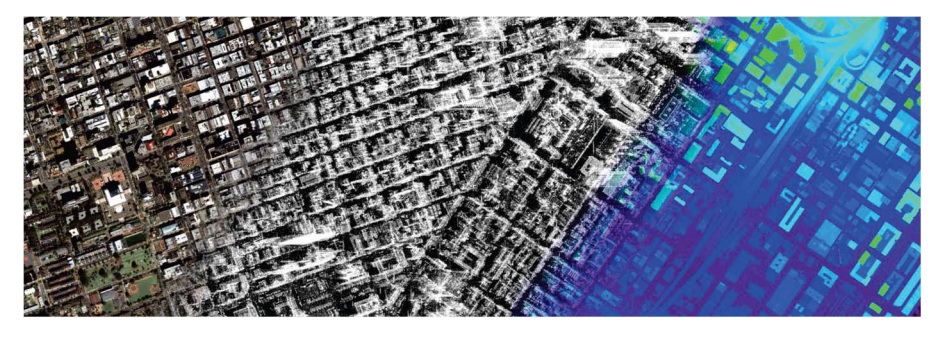}
				\caption{City: San Francisco, California, USA, from left to right the composition of optical, SAR and LiDAR elevation data}
				\label{fig:HSI.png}
			\end{figure}

		Hence, by applying the fusion paradigm, the joint information from optical, SAR, and LiDAR sources can provide a more detailed view of the surveyed region. Figure~\ref{fig:HSI.png} shows the data from different modalities and the composition of three acquisitions \cite{dalla2015challenges}.
		
		\begin{figure*}[!htbp]
			\centering
			\includegraphics[width=0.85\linewidth,trim=0.08in 0in 0.0in 0in, clip, keepaspectratio]{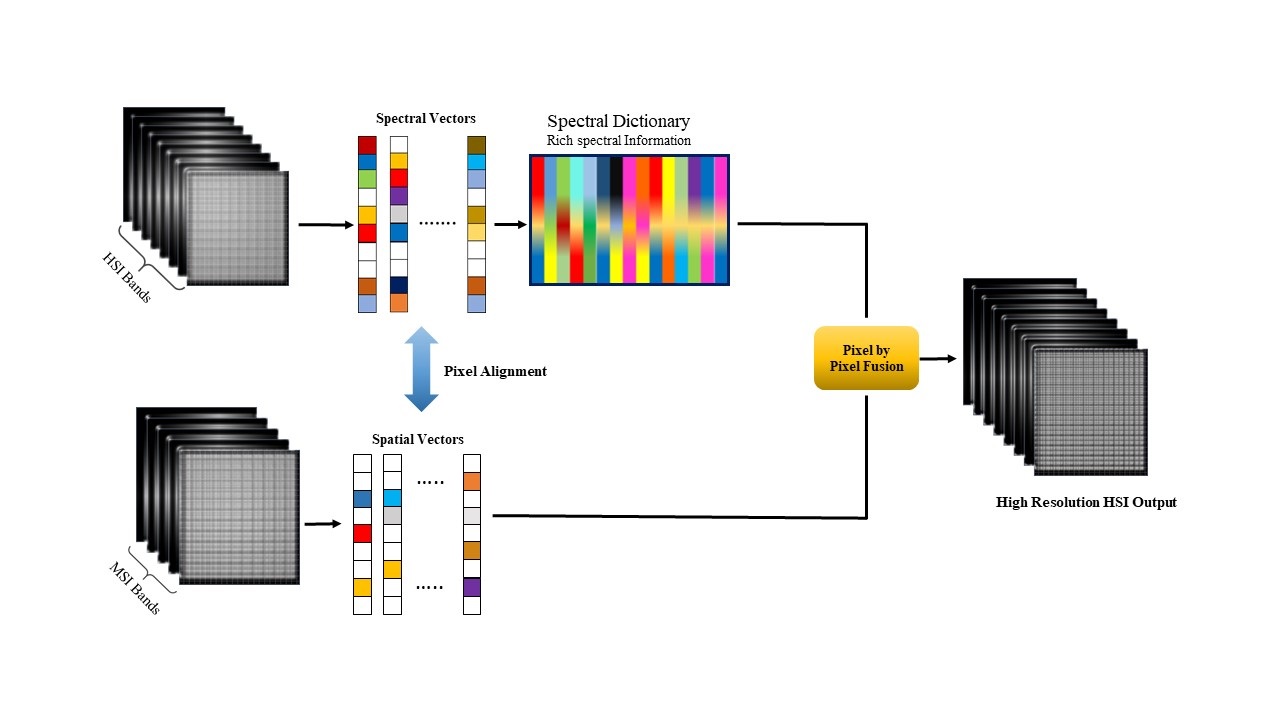}
			\caption{Pixel Level Fusion of HSI and MSI Images}
			\label{fig:pixel fusion.jpg}
		\end{figure*}
		\subsection{Data Fusion Processing}
		Data fusion is a widely used paradigm for processing data from many sensors and is applicable to a wide range of applications \cite{khaleghi2013multisensor}. In general, there are three processing stages at which data fusion can be carried out when focusing on remote sensing applications: raw data, feature, and decision levels \cite{javan2021review}. A complete description of the three fusion processing levels, along with relevant applications, is provided in the table below:
		\begin{table}[!htbp]
			\setlength{\tabcolsep}{0.0025\textwidth} 
			\caption{Fusion Processing Levels and Applications}
			\centering
			\scriptsize
			\begin{tabular}{|p{1.5cm}|p{3.5cm}|p{3.5cm}|}\hline
				
					\textbf{Processing}& \textbf{Description} & \textbf{Applications} \\
					\hline
					Pixel level & The process in which the resultant image is obtained by combining the values of input images pixel by pixel shown in figure ~\ref{fig:pixel fusion.jpg} . & 
					- Image sharpening \cite{li2017pixel}
					\newline
					- Super-resolution \cite{yin2013simultaneous}
					\newline
					- 3D model reconstruction from 2D views \cite{aharchi2020review} \\
					\hline
					Feature level & Geometric, spectral, and structural features such as textures, edges, angles, spectrums, and forms are extracted from the input images and merged. &
					- Land-cover classification. \newline
					- Combining different feature sets (e.g., linear or spatial features) for improved accuracy \cite{hedman2009road, pedergnana2012classification} \\
					\hline
					Decision level & Input images are processed separately, and instead of original images, the decisions taken from the original images are fused. 
					& 
					- Pattern recognition via ensemble learning \cite{bishop2006pattern} \\
					\hline
				\end{tabular}
				\label{tab:processing_levels}
			\end{table}
			
			\subsection{Fusion Methods in Remote Sensing}
			\subsubsection{Pansharpening}
			The term pansharpening refers to a fusion method in which multispectral (MS) and panchromatic (PAN) images of the same scene are fused together. MS images typically have low spatial resolution but high spectral content, while PAN images provide high spatial resolution but lower spectral content. By combining these, a complete set of information is obtained without discarding any details \cite{hasanlou2016quality, rodriguez2017object}. Several remote sensing platforms acquire both MS and PAN images using the same satellite to produce enhanced images \cite{alidoost2015region, jagalingam2015review}.
			
			Similar to fusion processing methods, pansharpening techniques are also divided into three categories: pixel level, feature level, and decision level \cite{belgiu2019spatiotemporal}. Among these, pixel-level pansharpening methods are most commonly used in remote sensing and computer vision applications for generating high spatial and spectral fused images. For pixel-level pansharpening, several strategies have been introduced. These techniques are further divided into four groups: Hybrid, Variational Optimization (VO), Multiresolution Analysis (MRA), and Component Substitution (CS). Table \ref{tab:pansharpening_techniques} lists 41 pixel-level pansharpening techniques categorized by each group \cite{javan2021review}.
			
			\begin{table*}[ht]
				\centering
				\caption{Pixel Level Pansharpening Methods} \label{tab:pansharpening_techniques}
				\renewcommand{\arraystretch}{1.5} 
				\setlength{\tabcolsep}{0.0025\textwidth} 
				\scriptsize
				
				\begin{tabular}{|p{1.5cm}|p{4cm}|p{4cm}|p{4cm}|p{4cm}|}
					\hline
					\bfseries Groups & \bfseries Component Substitution (CS) & \bfseries Multi-Resolution Analysis (MRA) & \bfseries Hybrid & \bfseries Variational Optimization (VO) \\ 
					\hline
					\textbf{Description} & 
					CS-based methods use specific transformation on Multispectral bands to split the spatial and spectral content then histogram matching is applied to adjust the PAN image with the extracted spatial information from MS image. After that, instead of the original PAN band, the adjusted spatial component is used. In the end, the modified component is converted back to MS image space by inverse transformation \cite{ghassemian2016review}. & 
					MRA-based methods focus on addressing the pansharpening problems by decomposing PAN and MS images into multiple resolutions or scale levels using wavelet or pyramid transformations. Spatial content from the PAN image at the chosen resolution is integrated into the corresponding MS level, then the fused image is generated by inverse decomposition \cite{aiazzi2002context,vivone2014critical }. & 
					Hybrid methods are used to overcome the limitations of MRA and CS-based methods by aiming to preserve both the spatial and spectral content of the input image. It usually combines wavelets and PCA-based or wavelets and IHS-based fusion methods \cite{gonzalez2005comparison}.& 
					VO-based methods often require extensive computations as compared to other methods. Most VO-based techniques rely on a model of the input image and the pansharpened outcome. They suggest an energy functional and use constraints derived from earlier assumptions to turn the pansharpening into an optimization problem \cite{wang2024vogtnet}.  \\ 
					\hline
					\bfseries Methods & \bfseries CS-Based Methods & \bfseries MRA-Based Methods & \bfseries Hybrid Methods & \bfseries VO-Based Methods \\ & 
					\begin{itemize}
						\item Principal Component Analysis (PCA) \cite{jiang2011survey}
						\item Brovey Transformation (BT) \cite{jiang2011survey, mandhare2013pixel}
						\item Generalized Intensity-Hue-Saturation (GIHS)\cite{tu2001new, tu2004fast, tu2007best, zhang2005ihs}
						\item GIHS with Tradeoff Parameter (GIHS-TP) \cite{choi2006new}
						\item GIHS with Best Tradeoff Parameter (GIHS-BTP) \cite{tu2007best}
						\item GIHS with Additive Weights (GIHS-AW) \cite{aiazzi2007improving}
						\item Improved GIHS-AW (IGIHS-AW) \cite{xu2008improved}
						\item Nonlinear IHS (NIHS) \cite{ghahremani2016nonlinear}
						\item Ehlers \cite{klonus2009performance, snehmani2017comparative}
						\item Gram-Schmidt (GS) \cite{klonus2009performance}
						\item Smoothing Filter-based Intensity Modulation (SFIM) \cite{liu2000smoothing, wald2002liu}
						\item Matting Model Pan-sharpening (MMP) \cite{kang2013pansharpening}
						\item Band Dependent Spatial Detail (BDSD) \cite{garzelli2007optimal, vivone2014critical}
						\item BDSD with Physical Constraints (BDSD-PC) \cite{vivone2019robust}
						\item Partial Replacement Adaptive Component Substitution (PRACS) \cite{choi2010new, vivone2014critical}
						\item High Pass Filter (HPF) \cite{chavez1991comparison}
						\item Indusion \cite{khan2008indusion}
						\item Hyperspherical Color Space (HCS) \cite{padwick2010worldview}
					\end{itemize} & 
					\begin{itemize}
						\item Substitutive Wavelet (SW) \cite{amolins2007wavelet, kim2010improved}
						\item Additive Wavelet (AW) \cite{amolins2007wavelet}
						\item Contrast pyramid \cite{zhang2004multiscale}
						\item Morphological Half Gradient (MF-HG) \cite{restaino2016fusion}
						\item Modulation Transfer Function-Generalized Laplacian Pyramid (MTF-GLP) \cite{aiazzi2002context, aiazzi2006mtf}
						\item MTF-GLP-Context-Based Decision (MTF-GLP-CBD) (MTF-GLP-HPM)\cite{aiazzi2002context, aiazzi2006mtf, aiazzi2007improving }
						\item MTF-GLP with High-Pass Modulation model (MTF-GLP-HPM) \cite{aiazzi2003mtf, aiazzi2006mtf}
						\item MTF-GLP-HPM with Post-Processing (MTF-GLP-HPM-PP) \cite{aiazzi2006mtf, lee2009fast}
						\item GLP based on Full Scale Regression-based injection coefficients (GLP-Reg-FS) \cite{vivone2018full}
					\end{itemize} & 
					\begin{itemize}
						\item Substitute Wavelet Intensity (SWI) \cite{gonzalez2004fusion}
						\item Additive Wavelet Intensity (AWI) \cite{nunez1999multiresolution}
						\item Additive Wavelet Luminance Proportional (AWLP) \cite{otazu2005introduction, kim2010improved}
						\item Revisited AWLP (AWLP-H) \cite{vivone2019fast}
						\item Weighted Wavelet Intensity (WWI) \cite{zhang2005ihs}
						\item Substitute Wavelet Principal Component (SWPC) \cite{gonzalez2004fusion}
						\item Additive Wavelet Principal Component (AWPC) \cite{gonzalez2005comparison}
						\item GS-Wavelet \cite{dadras2019new}
					\end{itemize} & 
					\begin{itemize}
						\item A Pan-sharpening method using Guided Filter (PGF) \cite{li2021pansharpening}
						\item P + XS \cite{ballester2006variational}
						\item Filter Estimation (FE) \cite{vivone2014pansharpening}
						\item A pan-sharpening method with sparse representation under the framework of wavelet transform (DWTSR) \cite{liu2013practical}
						\item Pan-sharpening with structural consistency and $\ell{1/2}$ gradient prior (L12-norm) \cite{zeng2016pan}
						\item A Variational Pan-Sharpening with Local Gradient Constraints (VP-LGC) \cite{fu2019variational}
					\end{itemize} \\ 
					\hline
				\end{tabular}
			\end{table*}
			
			\subsubsection{Deep Learning-Based Fusion}
			Generating an image with both high spectral and high spatial resolution remains a significant challenge \cite{dian2021recent}. HSI image super-resolution (HSI SR) has emerged as one promising technique to obtain the desired high-resolution hyperspectral image. Over the years, various approaches have been proposed to address this challenge, which can generally be classified into two groups: deep learning-based fusion methods and traditional fusion methods.	Many researchers suggest incorporating varying prior knowledge into traditional models to leverage intrinsic qualities within the context of maximum a posteriori (MAP) estimation. These priors, such as low rankness, self-similarity, and sparsity, have been widely used in applications like remote sensing pansharpening \cite{selva2015hyper, deng2019fusion} and HSI SR \cite{li2018fusing, dian2019hyperspectral}. However, these manual priors are often insufficient in capturing all hidden characteristics, and their optimal parameters require careful tuning for different datasets.
			
			Recently, deep learning-based approaches have gained traction in addressing HSI SR, remote sensing, and many other fields. CNNs have shown promising performance in these tasks \cite{dian2018deep, liu2020remote}. Unlike traditional methods that rely on manual priors, CNNs offer a more comprehensive and flexible approach. However, CNNs have limited receptive fields, restricting their ability to observe global features. Additionally, CNNs require an excessive number of parameters to achieve optimal results, which limits their scalability.
			
			In recent years, the transformer architecture developed by Vaswani et al. \cite{vaswani2017attention} and its variants have gained significant attention. Transformers are particularly powerful in extracting global information from the entire feature map. In \cite{hu2022fusformer}, Jin-Fan Hu et al. proposed Fusformer, which integrates the self-attention mechanism of transformers to address the limited receptive field problem in traditional convolutional operations. This enables transformers to capture a broader range of features, leading to superior image super-resolution results, as demonstrated by Gu and Dong \cite{gu2021interpreting}.
			
			Moreover, the newly developed diffusion model \cite{ho2020denoising, nichol2021improved} has shown outstanding performance in generation and reconstruction tasks. Recently, this model has been investigated in the field of HSI SR, yielding excellent results \cite{saharia2022image}. The diffusion model excels at extracting detailed texture data and global features, making it a highly effective choice for addressing HSI SR problems \cite{zhou2023hyperspectral}. It has also shown remarkable performance in complex spatial-spectral relationships and HSI classification tasks \cite{zhou2023hyperspectral}.
			
			Menghui Jiang and Jie Li \cite{jiang2022deep} presented the first deep residual cycle GAN-based heterogeneous integrated fusion framework. This framework fuses data from multimodalities and simultaneously integrates temporal, spectral, and spatial information. The approach enables multiple types of fusion, including heterogeneous spatiotemporal-spectral fusion, heterogeneous spatio-spectral fusion, and spatiotemporal fusion, while preserving the consistency between the fusion results. Their method effectively addresses bottlenecks such as thick cloud cover and land-cover change.
			
			Additionally, Xueying Li et al. \cite{li2023soil} proposed a deep learning-based feature fusion method to predict soil carbon content. They utilized HSI data and Visible Near-Infrared Reflectance Spectroscopy (VNIR) data, each contributing unique benefits for soil carbon prediction. Two types of multimodal networks were designed: one with artificial features and the other with an attention mechanism. By fusing the information according to the contribution difference of each feature, the attention mechanism-based network improves the predictive power of the fusion. The relative percent deviation for various locations, such as Jiaozhou Bay, Aoshan Bay, and Neilu Bay, increased by 31.16\%, 28.73\%, 43.96\%, and 24.28\%, respectively, when using a fusion network with artificial features, compared to using HSI and VNIR as single modalities.
			
			Table  \ref{Table:DLBasedFusionMethods} presents additional deep learning-based fusion methods, offering an overview of their characteristics and performance indicators. Moreover, to work in HSI domain, some tools and libraries are listed in Table \ref{table:AI_Software_Tools_and_Libraries_for_HSI_Analysis}.
			
			\begin{table*}[ht]
				
				\centering
				\caption{Deep Learning Based Fusion Methods} \label{Table:DLBasedFusionMethods}
				\setlength{\tabcolsep}{0.0025\textwidth} 
				\scriptsize
				\begin{tabular}{|p{2.5cm}|p{1.5cm}|p{5.5cm}|p{2.5cm}|p{5.5cm}|}
					
					\hline
					
					\textbf{Model} & \textbf{Modalities} & \textbf{Mechanism/Technique} & \textbf{Datasets} & \textbf{Key Features/Performance} \\ \hline
					FusAtNet & HSI, LiDAR & To create modality-specific embeddings for classification, the method uses cross-attention to enhance LiDAR-derived spatial data and self-attention to highlight HSI spectral features. & HSI-LiDAR MUUFL Gulfport \cite{gader2013muufl, du2017technical} & The proposed approach combines spectral and spatial characteristics from both modalities using self-attention and cross-attention techniques to increase classification accuracy.\cite{mohla2020fusatnet}. \\ \hline
					AsyFF-Net (Asymmetric feature fusion network) & HSI, SAR & AsyFFNet leverages asymmetric feature fusion through weight-shared residual blocks with batch normalization layers, & HSI-SAR Berlin, HSI-SAR Augsburg \cite{hong2021multimodal}& This outperforms competing methods and results in improved classification performance and efficient feature extraction for multisource remote sensing data.\cite{li2022asymmetric}. \\ \hline
					SS-MAE (Spatial-Spectral Masked Auto-Encoder) & HSI, LiDAR/SAR & Spatial–spectral masked autoencoder is used for reconstructing missing pixels and channels & HSI-SAR Berlin & Its hybrid Transformer-CNN architecture effectively enhances cross-modal feature representation across spatial and spectral domains.\cite{lin2023ss}. \\ \hline
					MHFNet (Multimodal MI-SoftHGR Fusion Network) & Multimodal (e.g., HSI, LiDAR) & The method utilizes the improved SoftHGR module for global handling of multimodal data & Berlin, Augsburg, MUUFL Gulfport & This approach improves recognition capability, reduces resource consumption by 40\%, and enhances classification accuracy \cite{zhang2024mhfnet}. \\ \hline
					MMGLOTS (Multimodal Global-Local Transformer Segmentor) & MSI, HSI, SAR & Global–local transformer and pre-trained Masked image modeling (MIM) model used for fusion and semantic feature extraction & C2Seg-AB, C2Seg-BW \cite{hong2023cross} & Best mean intersection over union (mIoU) and mF1 performance by averaging results across datasets; slight decrease in overall accuracy in C2Seg challenge\cite{liu2024cross}. \\ \hline
					MRSN (Multimodal Remote Sensing Network) & MSI, HSI, SAR & Uses separate backbones for each modality, fused features in stages (MSI+SAR first, then HSI after decoding) & C2Seg-AB, C2Seg-BW \cite{hong2023cross}& Focuses on fully extracting and fusing features from each modality; utilizes late-stage fusion for better performance. \cite{liu2024cross} \\ \hline
					Parallel Transformers & HSI, LiDAR & HSI as primary source with LiDAR branch using cross-attention (CA) & Houston, MUUFL Gulfport & Rigid CA fusion; does not fully consider LiDAR’s spatial-only characteristic, resulting in suboptimal fusion\cite{hu2022hyperspectral} \\ \hline
					SMDN (Semisupervised Multimodal Dual-Path Network) & HSI, LiDAR & It Leverages semisupervised multimodal dual-path network, SMDN simultaneously learns elevation data from LiDAR and spatial-spectral properties from HSI data. & Houston, Trento, MUUFL Gulfport & Encoder-decoder architecture; uses limited labeled data for unsupervised learning; achieves competitive co-classification performance \cite{pu2024multimodal}. \\ \hline
					UMDN (Unsupervised Multimodal Dual-Path Network) & HSI, LiDAR & It utilizes a forced fusion strategy to reconstruct unlabeled samples & Houston, Trento, MUUFL Gulfport & Successfully reconstructs and categorizes unlabeled samples, yielding robust co-classification results \cite{pu2024multimodal}. \\ \hline
					uHNTC (Unsupervised Hybrid Network of Transformer and CNN) & HSI, MSI & It uses 3 subnetworks; SpaDNet for spatial degradations using a dual-branch CNN, SpeDNet for spectral degradations using 1-D convolutions and FeafusFormer for a multilevel cross-feature attention (MCA) mechanism to fuse hierarchical spatial-spectral features.& PaviaU
					dataset \cite{dell2004exploiting}, Chikusei dataset \cite{yokoya2016airborne}, Xiongan dataset
					and real-worlds WorldView-2 & It outperforms 10 SOTA fusion methods on given datasets and real-world WorldView-2 images, it recovers HR-HSI with high accuracy in blind fusion settings and achieves superior cross-interaction by addressing both spatial and spectral degradation \cite{cao2024unsupervised}. \\ \hline
					GAF-Net (Global Attention-based Fusion Network) & Optical, MS, HSI, LiDAR, SAR, audio modalities & It uses global attention learning technique for feature refining by elimination of unnecessary and duplicate data.& Houston 2013 HSI-LiDAR, Augsburg HSI-SAR, Berlin HSI-SAR, Houston 2013 (HSI-MSI) \cite{hong2021multimodal} & GAF-Net shows superior performance across 5 benchmark multimodal datasets.It performs far better than modality-specific classifiers and can handle both local and global contexts in spectral-spatial self-attention \cite{jha2023gaf}.  \\ \hline
				\end{tabular}
			\end{table*}

		\begin{table*}[!htpb]
			\centering
			\caption{AI Software Tools and Libraries for HSI Analysis} \label{table:AI_Software_Tools_and_Libraries_for_HSI_Analysis} 
			\scriptsize
			\setlength{\tabcolsep}{0.0025\textwidth}
			\begin{tabular}{|p{1.5cm}|p{7.25cm}|p{1.5cm}|p{7.25cm}|}
			
					\hline
					\textbf{Name} & \textbf{Description and References} & \textbf{Name} & \textbf{Description and References} \\ \hline
					ENVI & Image processing S/W for spectral analysis, widely used in environmental research and resource exploration \cite{xue2011application}. & eCognition & Object-based image analysis S/W for segmentation, classification, and feature extraction in geospatial data \cite{nussbaum2008ecognition}. \\ \hline
				
					ERDAS IMAGINE & S/W for integrating image processing and GIS, used for satellite image analysis \cite{geosystems2005erdas}. & EnMAP-Box & Open-source QGIS plug-in for processing and visualizing HSI and multispectral data \cite{jakimow2023enmap}. \\ \hline
					
					Quantum GIS & Open-source GIS for spatial data creation, editing, and mapping \cite{graser2015processing}.& AVHYAS & QGIS tool for processing/analyzing HSI images in Python \cite{lyngdoh2021avhyas}. \\ \hline
					
					ATCOR & S/W for atmospheric correction of satellite and airborne imagery \cite{richter2019atmospheric}. & Spectronon & S/W for controlling Resonon systems and analyzing HSI datacubes \cite{abdulridha2019uav}. \\ \hline
					ACRON & S/W for atmospheric correction of HSI and multispectral data using radiative transfer \cite{chutia2016hyperspectral}. &Google Earth Engine & Tool for large-scale analysis of satellite imagery and geospatial datasets to detect changes and trends \cite{amani2020google}. \\ \hline

					SPAM & S/W optimized for analyzing imaging spectrometer data, including segmentation and spectral matching \cite{mazer1988image}. & TNTmips & Integrated geospatial analysis system with GIS, RDBMS, and image processing tools \cite{tancharoen2015molecularly}. \\ \hline
					
					ISIS & Converts raw data into usable products for satellite imagery, mapping, and creating elevation models \cite{gaddis1997overview}. & EnMAP & High-resolution imaging spectroscopy mission for environmental monitoring \cite{storch2023enmap}. \\ \hline
					SIPS & UNIX-based tool for imaging spectrometer data analysis, with visualization and modeling tools \cite{kruse1993spectral}. & OTB & Open-source remote sensing project with image viewer and apps for Python, QGIS, and C++ \cite{grizonnet2017orfeo}. \\ \hline
					
					SPECPR & Interactive program for analyzing one-dimensional arrays in reflectance spectroscopy \cite{clark1993spectrum}. & Spyndex & Python package for spectral indices and expression evaluation compatible with various Python objects \cite{montero2023standardized}. \\ \hline
					
					ORASIS & Identifies endmembers and creates exemplar sets in HSI datasets \cite{bowles2003orasis}. & SpectraSENS & S/W for HSI image acquisition, analysis, and visualization \cite{xiong2018micro}. \\ \hline
					
					ISDAS & Modular system for HSI data analysis using rapid prototyping on SPARC workstations \cite{staenz1998isdas}. & Hylite & Open-source Python package for preprocessing HSI imagery and integrating point-cloud data \cite{thiele2021multi}. \\ \hline
					Hsdar & R package for managing, analyzing, and simulating HSI data, including vegetation index calculation \cite{lehnert2018hyperspectral}. & HIAT & MATLAB-based toolbox for processing HSI and multispectral imagery with classification algorithms \cite{rosario2007matlab}. \\ \hline
					PyHAT & Python tool for spectroscopic data analysis with GUI, supporting Mars and Moon spectrometer data \cite{laura2022introduction}. & OpenHSI & Open-source project providing HSI technology with imaging spectrometer and S/W for reflectance data \cite{mao2022openhsi}. \\ \hline
				\end{tabular}
			\end{table*}

			\subsection{HSI Multimodality Challenges}
			A variety of data modalities are employed in conjunction with traditional hyperspectral multi-band imaging systems to enhance performance in several domains such as super-resolution and remote sensing. These modalities include LiDAR, Thermal Imaging (TI), Optical Imaging (OI), Radar Imaging (RI), Multispectral Imaging (MSI), SAR, and Audio Modality (AM). However, integrating multimodal data comes with several challenges, including issues related to data merging, alignment problems, weakly paired modalities, and differences in sensor properties.
			
			Additionally, different modalities often exhibit varying levels of noise and missing data, which require careful management to maintain data quality. Decisions regarding data preprocessing and weighting must be informed by a deep understanding of the specific characteristics of each modality. A well-defined and efficient data pretreatment pipeline is essential before initiating multimodal model training, as poor preprocessing can lead to increased operational workload. Despite the numerous advantages of multimodal sensing in various applications, the challenges outlined above must be addressed when designing such frameworks \cite{kieu2024multimodal}.
			
			In the field of multimodal satellite imagery segmentation, obtaining high-quality multimodal remote sensing datasets from aerial platforms or satellites presents significant challenges. The modalities that can be utilized in multimodal segmentation may be constrained by factors such as varying satellite revisit intervals and the high cost of acquiring such data \cite{fernando2023toward}. Similarly, for target detection in satellite imagery, the appearance of targets can vary significantly between different sensing techniques, affecting the detection process. Additionally, temporal variations in object appearance, shape, and orientation due to environmental factors or movement further complicate target detection.
			
			Noise in the data is a common problem caused by nonstationary objects, device failures, and environmental influences on the signals acquired during remote sensing operations. The inability of conventional deep learning algorithms to effectively differentiate between clean and noisy data reduces their performance. Despite being costly, inefficient, and challenging to fully detect and eliminate noisy samples, advanced models have made progress in extracting valuable features despite the presence of noise. Early approaches, such as student-teacher networks, have demonstrated effectiveness in managing noisy unimodal data and were later adapted for multiview learning. However, multimodal HSI still faces difficulties in handling noise, as it must balance feature extraction with noise tolerance across several modalities. Label noise, referring to incorrect ground truth annotations provided to the network for model optimization, is another challenge that can significantly decrease model performance \cite{kieu2024multimodal}.
			
			Handling missing data is another key challenge in applying fusion methods \cite{wang2023multi}. One common approach to address this issue is modality reconstruction, where the network predicts the missing data using available information from other modalities. Reconstructing lost signals has become more feasible with deep neural networks, particularly when using fully connected layers and CNNs \cite{dumpala2019audio}, \cite{tran2017missing}, \cite{zhao2021missing}. The core of deep learning-based reconstruction methods often involves an autoencoder, with techniques like the cascaded residual autoencoder \cite{tran2017missing} serving as a strong foundation for more recent and sophisticated methods \cite{zhao2021missing, wang2020multimodal}. However, utilizing the same encoder for different forms of data is challenging, as these modalities often have distinct features. Although useful, reconstruction techniques have limitations, as they typically focus on directly converting one data type to another, making it difficult to experiment with combinations of more than two modalities simultaneously. This one-to-one translation approach limits the extraction of complex information from diverse data sources.
			
			Addressing these challenges is critical for applications such as object detection, change detection, and image segmentation, as overcoming them would improve the robustness and accuracy of models. Unlocking the full potential of remote sensing tasks will require ongoing research and the development of innovative multimodal learning methodologies, particularly in high-impact areas like land analysis and disaster management \cite{kieu2024multimodal}.
			
			\subsection{Integration LLM with Hyperspectral Camera}
			The employment of LLMs like Visual ChatGPT shows a lot of promise in the context of HSI and Remote Sensing. When paired with visual computation, Visual ChatGPT performs exceptionally well in tasks like picture segmentation, line and edge recognition, and verbal image description. These skills aid in better interpretation and information extraction by providing insightful information about the content of hyperspectral images. Although it is still in its early stages, the integration of LLMs and visual models has the potential to revolutionize RSI processing and provide useful and approachable applications in the HSI domain \cite{NLP:HSI:Akewar:2024}.
			Hu et. al. \cite{NLP:HSI:Hu:2023:RSGPTAR} focuses on the development of massive vision language models VLMs for RS data analysis. The authors provide the Remote Sensing Image Captioning dataset RSICap and RSIEval datasets to aid in the training and assessment of large VLMs, emphasizing the necessity for complete datasets that are in line with RS tasks. 
			Tree-GPT, a framework that incorporates LLMs into forestry RS data processing, is proposed by the authors in \cite{NLP:HSI:du:2023:treegptmodularlargelanguage}. Tree-GPT gives LLMs domain expertise and picture comprehension skills to improve forestry data processing.
			The article \cite{NLP:HSI:Huang:2024} demonstrates how transformer-based foundation models have been successfully adapted for HSI classification using cutting-edge techniques including SS-VFMT (spectral-spatial vision foundation model based transformer), SS-VFMT-D, and SS-VLFMT(spectral-spatial vision-language foundation model based transformer). By effectively utilizing spectral-spatial features with few extra learnable parameters, these techniques improve classification performance while maintaining computing efficiency. Patch relationship distillation greatly increases training efficacy, and modules like SpaEM and SpeEM are crucial in fine-tuning the transformer to HSI-specific tasks. Additionally, SS-VLFMT's integration of a language model shows preliminary zero-shot learning (ZSL) capabilities, opening the door for new HSI classification applications without large labeled datasets. By using pretraining on remote sensing datasets to better link fundamental model information with HSI processing, this study lays the groundwork for future developments.

				\section{Compound Annual Growth Rate (CAGR) of HSI System}\label{sec:CAGR}
				The global market for hyperspectral imaging systems is typically assessed using the CAGR. CAGR represents the annual growth rate of an investment over a specific period, assuming that profits are reinvested each year and the growth occurs exponentially. It is one of the most accurate methods for calculating returns on investments that fluctuate over time. The formula to calculate CAGR is given by Equation \eqref{eq:cagr}:
				\begin{equation} \label{eq:cagr}
					CAGR = \left(\left( \frac{EV}{BV} \right)^{\frac{1}{n}} - 1\right) \times 100
				\end{equation}
				Here: \textit{EV} is the ending value of the investment, \textit{BV} is the beginning value of the investment, and \textit{n} is the number of years. CAGR helps quantify the annual growth rate of the hyperspectral imaging systems market over a specific time frame, providing insight into long-term market trends \cite{fernando2021cagr}. \par
\begin{figure}[!htbp]
					\centering
					\includegraphics[width=0.85\linewidth, trim=0.25in 0.20in 0.25in 0.25in, clip, keepaspectratio]{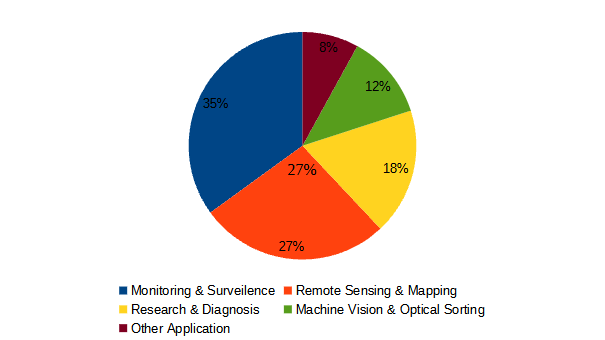}
					\caption{HSI Application Market Share}
					\label{fig:HSIAppsPercent}
				\end{figure}

				The purpose of explaining CAGR here is to make the concept accessible to professionals from technical fields such as computer science, material science, life sciences, and engineering, who may not be familiar with financial metrics. Understanding CAGR is crucial when evaluating global investments in hyperspectral imaging technologies, as it provides a reliable measure of the market's growth trajectory.
				The hyperspectral imaging systems market is primarily driven by monitoring and surveillance applications, according to a global market survey \footnote{https://www.grandviewresearch.com/horizon/outlook/hyperspectral-imaging-systems-market-size/global}, which account for 35\% of the market. This is followed by remote sensing and mapping at 27\%, research and diagnosis at 18\%, machine vision and optical sorting at 10\%, and other applications, which represent 8\% of the market (Figure \ref{fig:HSIAppsPercent}). According to this report, the global hyperspectral imaging systems market achieved a revenue of USD 14,136.6 million in 2023 and is projected to reach USD 28,639.8 million by 2030, reflecting a CAGR of 10.6\% between 2024 and 2030. Among product segments, cameras dominated the market, generating USD 10,153.2 million in 2023. Cameras are not only the largest but also the fastest-growing segment during the forecast period, making them a highly lucrative product category. From the report, it is observed that regionally, North America was the leading market in 2023, accounting for 34.7\% of the global revenue. The U.S. is expected to continue as the top contributor in terms of revenue by 2030. On a country-by-country basis, Kuwait is projected to experience the highest CAGR from 2024 to 2030, with its market expected to reach USD 99.2 million by 2030. \par
\begin{figure}[!htbp]
					\centering
					\includegraphics[width=0.85\linewidth, trim=0.0in 0.0in 0.0in 0.0in, clip, keepaspectratio]{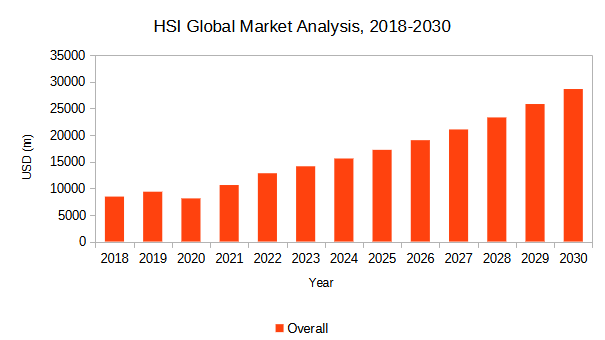}
					\caption{HSI Global Market Analysis 2018-2030}
					\label{fig:HSIGlobalMarketAnalysis2018To2030}
				\end{figure}

				The linear growth can be observed from Figure \ref{fig:HSIGlobalMarketAnalysis2018To2030}. The report is based on the market analysis that covers historical data from 2018 to 2022, with 2023 serving as the base year for estimation. The forecast period extends from 2024 to 2030, and the report provides a comprehensive outlook based on revenue in USD million. The report divides the market into cameras and accessories, with cameras driving the bulk of market growth. These trends highlight the significant expansion opportunities in the hyperspectral imaging systems market, particularly in North America and fast-emerging markets like Kuwait.
				
				
				Table \ref{Table:HSIKeyPlayersProducts} presents a list of key players in the HSI market along with their respective links. This table aims to bridge the information gap that practitioners and researchers often face when identifying leading organizations providing HSI-related solutions. In addition to listing prominent industry players, the table highlights some of their key products, while further details can be explored through the provided links. While this is not an exhaustive list, it includes some of the most significant contributors in the field.

				\begin{table*}
					\centering
					\caption{HSI key players and their products}	\label{Table:HSIKeyPlayersProducts}
					\setlength{\tabcolsep}{0.0025\textwidth}
					\scriptsize
						\begin{tabular}{|p{3.5cm}|p{9cm}|p{5cm}|}\hline 
							\textbf{Key Players in HSI} & \textbf{Products} & \textbf{URL Link} \\ 
							\hline

							Headwall Photonics, Inc. & Hyperspec VNIR, Hyperspec SWIR \cite{Headwall:VNIR:SWIRE:2016}& \href{https://headwallphotonics.com/}{https://headwallphotonics.com} \\ \hline
							
							Specim, Spectral Imaging Ltd. & Specim IQ, FX Series \cite{SpecimIQ:2018},\cite{SpecimFXSeries}& \url{https://www.specim.com/hyperspectral-cameras/}\\ \hline
							
							Resonon Inc. & Pika L, Pika XC2 \cite{Resonon:PIKAXC2},\cite{Resonon:PIKALPIKAXC2}& \url{https://resonon.com} \\ \hline
							
							Corning Incorporated & MicroHSI 410 SHD, Application Development Kit (ADK),Selectable Hyperspectral Airborne Remote-sensing Kit (SHARK) \cite{Corning:microHSI410SHARK}& \href{https://www.corning.com}{https://www.corning.com} \\	\hline 
							
							Telops Inc. & Hyper-Cam MWE, Hyper-Cam LW \cite{Telops:HyperCamLW} \cite{TelopsHyperCamMWE:Sungho:2018},  & \href{https://www.exosens.com/brands/telops}{https://www.exosens.com/brands/telops} \\	\hline 

							Norsk Elektro Optikk AS & HySpex VNIR-1800, HySpex SWIR-384 \cite{NEO:VNIRSWIR:Lubos:2023}\cite{NEO:VNIRSWIREtc}& \href{https://www.neo.no/contact/}{https://www.neo.no/contact/} \\ \hline
							
							BaySpec, Inc. & OCI-F Hyperspectral Imager, Snapscan Hyperspectral Camera \cite{BaySpec:OCIFSeriesCamers}&  \url{https://www.bayspec.com/} \\\hline
							
							Applied Spectral Imaging Inc. & HiBand (G-Band, Q-Band, and R-Band), GenASIs™ \cite{ASI:Hiband:PoseredbyGenasis}& \href{https://www.spectral-imaging.com}{https://www.spectral-imaging.com} \\\hline
							
							XIMEA Corp. & xiSpec, xiSpec2 \cite{xiSpec2factsheet} & \href{https://www.ximea.com}{https://www.ximea.com} \\ \hline
							
							Photon etc. & IR V-EOS, ZEPHIR Camera \cite{ZephIR2DOT5factsheet}&\href{https://www.photonetc.com}{https://www.photonetc.com},  \\ \hline
							
							Surface Optics Corporation & SOC 710-VP, SOC 710-E, SOC 710-SWIR \cite{SOC710Series}& \href{https://surfaceoptics.com/}{https://surfaceoptics.com} \\\hline 
							
							IMEC & SnapScan, On-Chip Filter \cite{imec:onchipfilters} & \url{https://www.imechyperspectral.com/en}\\ \hline
							
							Corescan & Core Imager, HighEye 4K \cite{corescan:HSI:2021}& \href{https://www.corescan.com.au}{https://www.corescan.com.au} \\ \hline
							
							Brimrose Corporation of America &Brimrose Acousto-Optic Tunable Filter (AOTF) Imager, AOTF spectrometers \cite{MWIR:Zhao:2017}\cite{Zhongpeng:AOTF:ma15082826}& \href{https://www.brimrose.com}{https://www.brimrose.com} \\ \hline
							
							Cubert GmbH & ButterflEYE X2, FireflEYE Q285, ULTRIS Q20 \cite{CubertGmbH:HSIDevices}& \href{https://www.cubert-gmbh.de}{https://www.cubert-gmbh.de} \\ \hline
							
							Spectral Evolution & PSR+ Series, SR-6500 \cite{spectralevolutionspectrometers} & \href{https://www.spectralevolution.com}{https://www.spectralevolution.com} \\ \hline
							
							GeoCue Group Inc. & True View, LIDAR Mapping Sensors \cite{GeoCue:TrueView}& \href{https://geocue.com/}{https://geocue.com/s} \\ \hline
							
							Lextel Intelligence Systems & EagleEye VNIR 100E, EagleEye SWIR 100E, \cite{LextelDevices}  & \url{https://www.lextel.com/}\\	\hline
							
							
							DEEPSEA & Camera (NanoSeaCam, SmartSight\textsuperscript{TM} MV100), spectrometers, Services (optical measurements) \cite{DeepSea:SmartSight:Kelly:2023}& \href{https://www.deepsea.com/nano-seacam/}{https://www.deepsea.com/nano-seacam/} \\	\hline
							
							Analytik Ltd&Nano HP VNIR Hyperspectral Imaging Sensors \cite{Analytik:NanoHP:2022}&\url{https://analytik.co.uk/hyperspectral-imaging/}\\	\hline
							
							Infrared Imaging &  SWIR, MWIR, LWIRIRIS camera & \href{https://www.infraredimaging.com}{https://infraredimaging.com} \\	\hline
							
							Leidos & Groundware\textsuperscript{R}, RADARs and Sensors \cite{Leidos:Groundaware}& \href{https://www.leidos.com}{https://www.leidos.com} \\		\hline
							
							Maxar Technologies & SWIR, Satellite Imagery, Optical Imagery, Crow's Nest, Precision3D Data Suite,
							WorldView-3, GeoIQ \cite{Maxar:Capabilities:2021} & \href{https://www.maxar.com}{https://www.maxar.com} \\		\hline
							
							NV5 Global & NV5 HSI, NV5 HSI Pro,ENVI, ENVI® Inform, ENVI® Edge,ENVI® SARscape® etc, INSITE \cite{NV5Devices}  & \href{https://www.nv5.com}{https://www.nv5.com} \\		\hline

							Orbital Sidekick & SIGMA Monitor,SIGMA (Spectral Intelligence Global Monitoring Application) Data, Spectral Intelligence™, GHOSt™ Constellation, Aurora, Aurora+ \cite{OrbitalSidekickDevices}& \href{https://www.orbitalsidekick.com/}{https://www.orbitalsidekick.com/}\\		\hline
							
							Pixxel &hyperspectral imaging satellites, Aurora, Pixxel’s in-house Earth Observation Studio, Firefly \cite{pixxel:Technologies:2024}&\url{https://www.pixxel.space/firefly} \\
							\hline
							
						\end{tabular}
					\end{table*}

					\section{HSI Applications}\label{sec:HSIApplications}
					
					HSI has diverse applications across multiple domains due to its ability to capture rich spectral and spatial information. In food quality and safety assessment, HSI enables non-destructive analysis of chemical and physical properties, aiding in contaminant detection, freshness evaluation, and nutrient assessment \cite{he2015hyperspectral, lorente2012recent, kwak2020realtime, chung2021detection, lee2024detection, saeidan2021detection}. In medical diagnostics, it supports early cancer detection, surgical guidance, and real-time blood oxygen monitoring, improving patient outcomes \cite{martin2006development, leon2020noninvasive, aboughaleb2020hyperspectral, mangotra2023hyperspectral, barberio2021intraoperative, barberio2020hyper, lee2021multimodal, sucher2019hyperspectral, puustinen2023hyperspectral, elsharkawy2024advancements, sicher2018hyperspectral, martinezvega2021oxygen, holmer2016oxygenation, cancio2006hyperspectral}. Precision agriculture benefits from HSI through crop health monitoring, pest detection, and soil nutrient analysis, optimizing resource use and boosting yields \cite{lu2020recent, mahesh2015hyperspectral, terentev2022current, thomas2018benefits, wan2022hyperspectral, peng2019prediction, du2009evaluation, pal2024portable, omia2023remote, yu2022critical, liu2021hyperspectral, qin2023hyperspectral}. In water resource and flood management, HSI is employed for flood mapping, pollution tracking, and drought assessment, facilitating environmental conservation \cite{klemas2015remote, farhadi2022flood, rahman2017state, leung2024water, liu2021uavborne, li2016water, cao2021monitoring, kim2011hyperspectral, gerhards2019challenges, duartecarvajalino2021estimation, kim2015highly}. Similarly, forensic science utilizes HSI for analyzing fingerprints, documents, and biological evidence, enhancing criminal investigations \cite{2018:HSI-ImageAnalysis:Review:IEEEAccess}. HSI also plays an important role in artwork authentication, detecting forgeries, pigment composition, and hidden layers in paintings \cite{huang2022recent}. The most critical use of HSI is seen in defense and security applications, which include surveillance, target detection, and camouflage identification, strengthening national security \cite{yuen2010introduction, 2018:HSI-ImageAnalysis:Review:IEEEAccess, tseng:spectralUnmixing:2000}. In the field of microscopy, HSI aids in analyzing biological samples, etc. The researchers involved in applying HSI technology to different fields of life are organized in Table \ref{Table:ApplicationsPart1}.

					\begin{table*}[htpb]
						\caption{Applications of Hyperspectral Imaging}
						\label{Table:ApplicationsPart1}
						\centering
						\setlength{\tabcolsep}{0.0025\textwidth}  
						\scriptsize
						\begin{tabular}{|p{1.25cm}|p{2.25cm}|p{6.75cm}|p{6cm}|p{1cm}|}
							\hline
							\textbf{Category} & \textbf{Application} & \textbf{Description} & \textbf{Benefits} & \textbf{Ref.} \\\hline
							
							\multirow{5}{*}{\parbox[c][2.5cm][c]{1.3cm}{Food Quality and Safety Assessment}}
							&Detection of Foreign Substances in Foods &Identifies and segregates foreign materials such as stones and plastics in processed foods using unique spectral signatures. &Ensures product safety through non-destructive testing, reducing contamination risks across the supply chain. &\cite{kwak2020realtime, he2015hyperspectral, chung2021detection, lee2024detection, saeidan2021detection}\\
							\cline{2-5}
							&Shelf-Life Prediction for Fresh Fruits and Vegetables &Monitors moisture content, color, and surface texture of fresh products to predict freshness and shelf life, enabling real-time quality control during storage and transport. &Enhances shelf-life prediction accuracy, minimizes waste, and ensures that only high-quality, fresh products reach consumers. &\cite{ktenioudaki2022decision, mohammed2023machine, li2024research, pu2015recent, lara2013monitoring}\\
							\cline{2-5}
							&Evaluation of Meat Freshness and Safety &Detects chemical composition changes in meat to assess freshness, pH levels, and potential microbial contamination. &Enables early detection of spoilage during distribution, ensuring consumer safety and reducing losses.
							&\cite{mladenov2020model, fu2019review, kutsanedzie2019advances, yang2017detection, pu2023recent}\\
							\cline{2-5}
							&Quality Assessment of Cereals and Grains &Evaluates protein content, moisture levels, and presence of mycotoxins in grains, facilitating stringent quality control during grain processing. &Maintains consistent quality through accurate, rapid, non-destructive measurements, ensuring compliance with safety standards in grain production. &\cite{feng2019hyperspectral, aviara2022potential, an2023advances, mahesh2015hyperspectral, sendin2018near, caporaso2018near}\\
							\cline{2-5}
							&Assessment of Sweetness and Texture in Fruits &Measures soluble solid content, firmness, and texture in fruits, which are critical for consumer satisfaction and processing requirements. &Ensures optimal ripeness and flavor, enhances product quality, and supports tailored processing for a variety of fruit-based products, boosting market competitiveness. &\cite{lorente2012recent, rahman2018hyperspectral, sun2017how, pullanagari2021uncertainty, wang2016recent, lu2017innovative}\\
							\hline
							
							\multirow{3}{*}{\parbox[c][2.5cm][c]{1.3cm}{Medical Diagnosis}}
							&Early Cancer Diagnosis &Enables non-invasive cancer diagnosis (e.g., brain, breast, gastric) by analyzing spectral differences between cancerous and healthy tissues for early detection.&Improves treatment success, reduces costs and complications, and boosts patient survival rates through early diagnosis. &\cite{martin2006development, leon2020noninvasive, aboughaleb2020hyperspectral, mangotra2023hyperspectral}\\
							\cline{2-5}
							&Guidance for Surgical Procedures &Provides real-time scanning of the surgical area, distinguishing blood vessels, tumors, and healthy tissue to guide procedures. &Increases surgical accuracy, reduces errors and complications during surgery, and accelerates recovery. &\cite{barberio2021intraoperative, barberio2020hyper, lee2021multimodal, sucher2019hyperspectral, puustinen2023hyperspectral}\\
							\cline{2-5}
							&Monitoring Blood Composition &provides non-invasive measurement of blood oxygen saturation and composition within tissues, useful for ongoing health monitoring. &Enables real-time blood condition monitoring for better patient management and emergency response. &\cite{elsharkawy2024advancements, sicher2018hyperspectral, martinezvega2021oxygen, holmer2016oxygenation, cancio2006hyperspectral}\\
							\hline
							
							\multirow{3}{*}{\parbox[c][2.5cm][c]{1.3cm}{Precision Agriculture}}
							&Monitoring Crop Health Status &Conducts non-invasive monitoring of crop health (e.g., stress, pests, and diseases) to optimize agricultural productivity by detecting specific health conditions through spectral signatures of crops. &Reduces pesticide and fertilizer usage, increases crop yield, and enables early response to pest and disease outbreaks, minimizing potential yield loss. &\cite{lu2020recent, mahesh2015hyperspectral, terentev2022current, thomas2018benefits, wan2022hyperspectral}\\
							\cline{2-5}
							&Soil Fertility and Nutrient Content Analysis &Analyzes organic matter, moisture levels, and nutrients (e.g., nitrogen, calcium) in the soil to assess soil fertility and provide necessary nutrients for crop cultivation. &Enables precise fertilizer application, reducing environmental pollution and enhancing crop quality and productivity. &\cite{lu2020recent, peng2019prediction, du2009evaluation, pal2024portable}\\
							\cline{2-5}
							&Crop Classification and Growth Stage Assessment &Identifies growth stages and classifies the health status of various crops to support efficient crop management. &Allows for real-time monitoring of crop development, enabling optimal management strategies and timing for harvest. &\cite{lu2020recent, omia2023remote, yu2022critical, liu2021hyperspectral, qin2023hyperspectral}\\
							\hline
							
							\multirow{3}{*}{\parbox[c][2.5cm][c]{1.3cm}{Water Resource and Flood Management}}
							&Flood-Affected Area Monitoring, Mapping &Scans areas before and after flooding to map flood-affected regions providing a rapid and accurate identification of flooded areas. &Quickly analyzes large areas; provides accurate information for emergency response; increases efficiency in recovery &\cite{klemas2015remote, farhadi2022flood, rahman2017state}\\
							\cline{2-5}
							&Identification of Water Pollution Sources &Analyzes water color and optical properties to detect pollutants (e.g., runoff, sewage, wastewater) and, with UAV tech, tracks pollution sources in real-time. &Allows for rapid identification of pollution sources, facilitating appropriate remediation actions and minimizing environmental impact from contamination. &\cite{leung2024water, liu2021uavborne, li2016water, cao2021monitoring}\\
							\cline{2-5}
							&Detection of Moisture Stress in Agriculture and its Surroundings &Enables early detection of moisture stress in crops or vegetation, alerting to potential risks like drought. This aids water management for agricultural areas connected to specific water resources. &Allows timely irrigation interventions to maintain agricultural productivity and improve efficiency in water resource management. &\cite{kim2011hyperspectral, gerhards2019challenges, duartecarvajalino2021estimation, kim2015highly}\\
							\hline
							
							\multirow{3}{*}{\parbox[c][2.5cm][c]{1.3cm}{Forensic}}
							&Evidence Analysis &Uses HSI to examine forensic evidence such as bloodstains, fingerprints, and documents without damaging materials. &Enables nondestructive analysis and enhances detection of hidden or minute details for accuracy. &\cite{de2023hyperspectral,melit2021forensic}\\
							\cline{2-5}
							&Trace Analysis &Identifies substances like explosives and gunshot residue at crime scenes through spectral analysis of complex materials. &Preserves forensic evidence while improving trace detection capabilities. &\cite{de2023hyperspectral,melit2021forensic}\\
							\cline{2-5}
							&Postmortem Imaging &Assists in identifying injuries and patterns on bodies that conventional methods may miss. &Increases accuracy of forensic analysis in postmortem scenarios. &\cite{de2023hyperspectral,melit2021forensic}\\
							\hline
							
							\multirow{6}{*}{\parbox[c][6cm][c]{1.3cm}{Document Examination}}
							& Ink Analysis & HSI detects spectral differences in inks that look similar, helping identify fraud, ink age, and origin. & Allows for non-destructive analysis, reducing subjective judgment and improving objectivity & \cite{de2023hyperspectral} \\
							\cline{2-5}
							& Forgery Detection & HSI, along with techniques like PCA, PLS-DA, and MCR-ALS, identifies forgeries like ink-aging, crossed lines, and obliteration. & Enables detection of forgery patterns and document alterations with high accuracy. & \cite{de2023hyperspectral} \\
							\cline{2-5}
							& Pen Discrimination & HSI differentiates between inks from various pen types (e.g., gel, ballpoint) using spectral analysis and clustering techniques like K-means and HCA. & Achieves high discrimination between inks of different colors and types, supporting forensic ink analysis. & \cite{de2023hyperspectral} \\
							\cline{2-5}
							& Chronological Order of Lines & NIR-HSI and Raman HSI analyze the sequence of crossing ink lines, with chemometric techniques such as PCA and MCR-ALS. & Improves accuracy in determining the order of ink applications, crucial for verifying document authenticity. & \cite{de2023hyperspectral} \\
							\cline{2-5}
							& Document Alterations & HSI combined with chemometric tools like PCA and Projection Pursuit detects document alterations, including added text, erasures, and obliterations. & Allows objective assessment of document modifications and enhances the ability to detect subtle changes. & \cite{de2023hyperspectral} \\
							\cline{2-5}
							& Comprehensive Ink Analysis & UV–Vis-HSI and VNIR-HSI differentiate similar inks, even among same-colored pens, with tools like SAM and t-SNE for visualization. & Increases classification accuracy, offers data-rich insights, and efficiently handles high-dimensional datasets. & \cite{de2023hyperspectral} \\
							\hline
							
							\multirow{4}{*}{\parbox[c][4cm][c]{1.3cm}{Artwork Authentication}}
							& Counterfeit Detection & HSI captures spectral data across various wavelengths, distinguishing between materials that look similar to the human eye but have unique spectral fingerprints. & Enables accurate identification of pigments and materials, enhancing verification of historical consistency and authenticity of artworks. & \cite{2018:HSI-ImageAnalysis:Review:IEEEAccess} \\
							\cline{2-5}
							& Identification of Pigments & Analyzes pigment spectral fingerprints to verify consistency with the artwork's creation period, using machine learning tools like SVM for accurate pigment classification. & Detects forgeries by identifying modern materials, which differ spectrally from historical ones, with accuracy rates up to 78\% for pigment identification. & \cite{huang2022recent} \\
							\cline{2-5}
							& Hidden Features Detection & HSI reveals underlying layers, overpainting, or concealed features that are invisible under standard light conditions. & Helps identify restorations, alterations, or hidden details, aiding art conservation and analysis. & \cite{2018:HSI-ImageAnalysis:Review:IEEEAccess} \\
							\cline{2-5}
							\hline
							\multirow{5}{*}{\parbox[c][2.5cm][c]{1.3cm}{Defense and Security}} 
							& Surveillance and Reconnaissance & Hyperspectral satellites and airborne sensors identify subtle environmental changes, detect anomalies, and work under low-light conditions for enhanced surveillance. & Enables monitoring of military activities, camouflage detection, and improved situational awareness across varied terrains. & \cite{yuen2010introduction,2018:HSI-ImageAnalysis:Review:IEEEAccess,tseng:spectralUnmixing:2000}\\
							\cline{2-5} 
							& Target Detection & HSI identifies specific materials (explosives, chemical agents etc.) through spectral signatures in camouflaged or hidden conditions& Vital for identifying concealed weaponry and distinguishing threats in border security and counter-terrorism. & \cite{yuen2010introduction,2018:HSI-ImageAnalysis:Review:IEEEAccess,tseng:spectralUnmixing:2000}\\
							\cline{2-5} 
							& Counter-IED Operations & Analyzes soil disturbances to identify buried improvised explosive devices (IEDs). & Enhances safety by detecting hidden IEDs in field environments, improving response time. & \cite{yuen2010introduction,2018:HSI-ImageAnalysis:Review:IEEEAccess,tseng:spectralUnmixing:2000}\\
							\cline{2-5} 
							& Detection of Enrichment Facilities & Uses spectral imaging to locate concealed or camouflaged nuclear enrichment facilities. & Supports nuclear threat detection by identifying specific structural features distinct from ordinary buildings. & \cite{yuen2010introduction,2018:HSI-ImageAnalysis:Review:IEEEAccess,tseng:spectralUnmixing:2000}\\
							\cline{2-5} 
							& Underwater Threat Detection & Hyperspectral sensors detect submarines and mines in shallow waters. & Increases maritime security by enhancing detection capabilities in naval warfare scenarios. & \cite{yuen2010introduction,2018:HSI-ImageAnalysis:Review:IEEEAccess,tseng:spectralUnmixing:2000}\\
							\hline
							Microscopy &Spectroscopy of lipid layers &Hyperspectral optical microscopy uses hyperspectral imaging integrated with a variety of optical microscopes. &Achieve precise spectral calibration and spatial resolution. & \cite{2017:MultimodalHyperspectralOpticalMicroscopy:ChemicalPhysicsJ}\\
							\hline
						\end{tabular}
					\end{table*}
				
					
\section{Open Research Problems, Challenges, and Recent Trends}\label{sec:ChallengesandTrends}
\subsection{HSI Open Research Problems and Challenges} 
\begin{enumerate}
	
	\item \text{Data Volume and Processing Efficiency:} The rapid growth in HSI data volume presents significant computational challenges, particularly due to the small size of data for detailed tasks. Labeling more data to address the lack of labeled samples is resource-intensive and impractical for many applications. A more feasible alternative is to develop algorithms that can leverage large amounts of unlabeled data alongside limited labeled data \cite{HSI:Challenges:Datta:2022}. Techniques such as unsupervised feature learning, semi-supervised learning, and active learning can reduce the reliance on labeled data while optimizing computational efficiency \cite{HSIVolandProcessing:Judy:2024, zhou2020advances}. Additionally, utilizing efficient computing architectures like MapReduce and cloud computing services can help manage these challenges effectively \cite{HSIVolandProcessing:Cloud:Zebin:2021, HSIVolandProcessing:Cloud:Haut:2024}.
	
	\item \text{Real-time Data Processing:} Real-time processing of HSI data is critical in areas like defense and surveillance. However, limited processing power on mobile and edge devices poses challenges for deploying real-time HSI solutions \cite{tseng:spectralUnmixing:2000}. To address this, cloud computing platforms are increasingly used for hyperspectral data processing in distributed architectures, enabling high-performance, service-oriented computing. The adoption of GPUs, along with deep learning frameworks like \textbf{Apache Spark, TensorFlow, and Caffe}, has facilitated the expansion of real-time hyperspectral analysis. \textbf{Apache Spark Streaming}, for example, supports real-time data processing through \textbf{Spark's DStream} (discretized streams), allowing continuous feature detection, pattern recognition, and anomaly identification as new data arrives. Spark's unified API simplifies batch and stream processing, making it suitable for dynamic HSI analysis \cite{HSIRealTimeDataProcessing:ApachiSpark:Zbakh:2023}. Other frameworks such as \textbf{Kafka} and \textbf{Apache Spark with Kafka or Flink} also provide promising alternatives for real-time processing \cite{HSIRealTimeDataProcessing:ApachiSparkKafka:Vu:2024, HSIRealTimeDataProcessing:ApachiSpark:Shaik:2024, HSIRealTimeDataProcessing:ApachiSparkKafkaFlink:Tosi:2024, HSIRealTimeDataProcessing:ApachiSparkKafkaFlink:Juan:2020}.
	
	The key challenge in consolidating fast computing techniques for real-time hyperspectral data analysis is the high energy consumption of high-performance computing architectures. It works fine for ground operation, but, it limits their use in onboard operations like satellite platforms. Beside that devices such as GPUs consume excessive power, making their integration thoughtful. Additionally, these platforms face radiation tolerance issues. Future advancements in hardware are essential for efficient real-time processing of HSI data in satellite missions \cite{AdvancesinHCI:Ghamisi:2017}.

	\item \text{High Inter-Class Variance and Spectral Variability:} HSI data are often impacted by environmental noise, which reduces image clarity and can lead to inaccurate analysis. Spectral reflectance or absorbance profiles can vary significantly, even for the same material, due to differences in sensor types, wavelength ranges, and spectral resolutions. Additionally, factors such as viewing angle, atmospheric conditions, sensor altitude, and geometric distortions contribute to spectral variability, complicating classification tasks \cite{zhou2020advances}. Advanced noise reduction techniques, domain adaptation methods, and robust spectral preprocessing can help mitigate these effects \cite{MitigateEnvironmentalEffectsonHSI:DING:2024}\cite{HSIVarainceVariabilityIssue:Sarpong2024}.
	
	\item \text{Miniaturization of Sensors:} Miniaturizing hyperspectral sensors for portability and mobile device integration often compromises sensor performance, affecting the quality of captured data \cite{lee2022compact}. However, combining advanced optical design, on-chip spectral analysis, novel materials, and efficient signal processing offers promising solutions. These innovations aim to address the trade-offs between miniaturization and data quality, enabling high-performance sensors for mobile and portable applications. Table \ref{Table:SolutionsOfMiniaturization} provides an overview of such advancements.
	\begin{table}[!htpb]
		\centering
		\caption{Solutions of Miniaturization Problem}	\label{Table:SolutionsOfMiniaturization}
		\setlength{\tabcolsep}{0.0025\textwidth}  
		\scriptsize
		\begin{tabular}{|p{1.75cm}|p{5.75cm}|p{0.70cm}|}
			\hline
			\textbf{Technique} & \textbf{Description} & \textbf{Ref.} \\
			\hline
			Miniaturized Optics & Compact optical systems like micro-lenses and diffractive elements help focus light efficiently on smaller sensors. &\cite{Diffraction:Spectrometers:Chen:2024,AdvancedInMiniComSpctrmtrs:Xue:2024,MimiatutrisedOpitcs:Ilchenko:2024}\\
			\hline
			On-Chip Spectrometers & MEMS and CMOS-based spectrometers reduce size by integrating spectral selection on a single chip. &\cite{CMOSMEMSensors:Kang:2019}\cite{OnchipHSISensors:BavoDelaure:2024} \\
			\hline
			Advanced Sensor Materials & Quantum dots and photonic crystals enable broader spectral coverage while maintaining compact size. & \cite{QuantumDotEnabled:HSISensors:Meng:2024} \\
			\hline
			Pixel-Level Integration & Higher pixel density on CMOS/InGaAs sensors improves data capture in a smaller form factor. & \cite{InGaAsInP:ImageingSensors:Tang:2024}\cite{InGaAs:ImageingSensors:Geum:2024} \\
			\hline
			Signal Processing & On-chip processing and data compression help mitigate reduced sensor performance and memory limitations. &\cite{OnChipCompression:2018:Keymeulen}\cite{ParallelDatCompress:Chatziantoniou:2024} \\
			\hline
			MME Techniques with Nanoscale Structures&Micro-Mechanical-Electro (MME) techniques combined with nanoscale structures enable the creation of compact, advanced devices that enhance analytical methods and capabilities.&\cite{NanoMiniaturization:Thakur:2022}\\ \hline
			Software-Based Calibration & Calibration algorithms replace physical systems, improving sensor performance in portable applications. &\cite{SoftwarebasedCalibration:Shen:2025} \\
			\hline
			Compact Power Management & Low-power electronics and solid-state storage enable operation of miniaturized sensors without excessive power drain.&\cite{Jain2024,CompactPowerMgt:MuhammadSaqib:2024} \\
			\hline
			3D Integration Technology & Stacked sensor layers reduce footprint and integrate multiple components efficiently. &\cite{3Dintegration:Guan:2023}\\
			\hline
			Hybrid Imaging Approaches & Combining multiple sensing technologies on a single chip optimizes performance and space usage. &\cite{hybridAppraochforMiniaturization:Wang:2024} \\
			\hline
			Miniaturized Light Sources & Small LEDs and laser diodes reduce size and power consumption for effective illumination in hyperspectral systems. &\cite{LightwieghtLEDs:Islam:2017} \\
			\hline
		\end{tabular}
	\end{table}
	
	\item \text{Challenges in Feature Learning for Hyperspectral Data:} Traditional feature learning methods, such as dimensionality reduction and spatial feature extraction, often rely on assumptions about data characteristics, complicating task-specific optimization. Deep learning approaches, however, adaptively learn features within the context of the overall analysis task. Despite this, many methods, like autoencoders, require hyperspectral data to be flattened, which results in the loss of spatial information. While convolutional autoencoders and hierarchical models (e.g., DBNs, RBMs) have shown promise, challenges remain in preserving multi-scale spatial-spectral correlations, ensuring robustness in extracted features, and achieving high-resolution reconstructions in unsupervised settings \cite{prasad2020hyperspectral}. Deep learning-based methods, like convolutional autoencoders and hierarchical models, help to address these challenges by preserving spatial-spectral correlations and optimizing multi-scale feature learning \cite{HSIVolandProcessing:Judy:2024, HSISingleModelMultiModelDeepLearning:Shivam:2024}.
	
\item \text{Model Problem in Bio-Geophysical Data Processing:} The main challenge in processing bio-geophysical data lies in managing large, complex datasets with semi-automatic techniques that are accurate, robust, reliable, and efficient. While many bio-geophysical retrieval methods have been developed, few have been widely adopted for operational use. Machine learning has shown promising results in estimating climate variables, such as the leaf area index (LAI) and Gross Primary Production (GPP), using neural networks, kernel methods, random forests, and support vector methods. However, these techniques are often applied without adapting to the specific spatial and temporal structures of the data. There is a need for algorithms that naturally incorporate data characteristics, such as customized kernel functions or convolutional networks, and that can seamlessly fuse multi-sensor data without relying on simplified re-sampling methods. Developing methods that can handle these complexities while scaling to large datasets remains a key challenge \cite{prasad2020hyperspectral}.
	
\item {AI-based HSI Legal, Ethical, Social, and Regulatory Challenges:} AI-based applications, including Hyperspectral Imaging (HSI), face a range of legal, ethical, social, and regulatory challenges, such as data privacy, algorithmic bias, decision transparency, and compliance with industry standards and regulations. Addressing these concerns is critical for the responsible development and deployment of AI technologies, including HSI, across various sectors \cite{AI:Ethics:Bhatti2023}. While AI has the potential to revolutionize industries by automating tasks that traditionally require human intelligence, benefiting sectors such as healthcare, self-driving cars, and household management, concerns about misuse and unintended consequences have prompted efforts to establish standards for trustworthy AI. In 2024, numerous states introduced AI-related bills, with many enacting legislation to address these challenges and ensure the safe application of AI technologies \cite{NCSL:AIPolicyLawsInfoDocPage:2024}. However, much more legislation and regulation are needed to ensure the responsible behavior of AI solutions, including HSI. Countries worldwide are developing AI governance policies and legislation to balance innovation and risk regulation, with some focusing on specific use cases, national strategies, or ethics policies. These efforts are unfolding globally, addressing the transformative nature of AI and its regulatory challenges. These can be tracked using Global AI Law and Policy Tracker \cite{GlobalAILawsTracker:2024:IAPP}.
\end{enumerate}

					\subsection{Recent Trends and HSI}
					\begin{enumerate}
						\item \text{Digital Twin:} Digital twins facilitate the real-time simulation and monitoring of physical assets, thereby enabling predictive maintenance, optimization, and resilience in a range of sectors, including manufacturing and autonomous systems. Significant developments have been made in the integration of real-time data from sensors and the enhancement of simulation fidelity through the application of AI in analytical processes \cite{yu2023autonomous}.
						\item \text{AI-Driven Edge Computing for Real-Time Analytics:} The utilization of AI at the edge (in proximity to data sources) serves to diminish latency and bandwidth usage, which is of paramount importance for applications that necessitate immediate data processing, such as autonomous systems and IoT-enabled infrastructures. The implementation of optimized processing algorithms serves to enhance the accuracy and speed of analysis, thereby rendering AI-driven edge computing a suitable option for on-site data interpretation \cite{wang2024digital}.
						\item \text{Sparse and DL Techniques for Hyperspectral Unmixing:} This approach employs advanced noise filtering algorithms and machine learning models to extract meaningful information from hyperspectral data while preserving its integrity and minimizing the impact of noise \cite{prasad2020hyperspectral}.
						\item \text{Sensor Technology for High-Resolution Data Capture:} Modern sensor designs aim to balance portability with high resolution, supporting miniaturized yet powerful sensing capabilities for diverse applications, including hyperspectral imaging and autonomous navigation. Advances include compact sensors that maintain fidelity in data capture while conserving energy, suitable for mobile devices and autonomous vehicles \cite{campolo2024edge}.
						\item \text{Autonomous Driving Systems:} Autonomous driving requires precise real-time data from various sensor types, including LiDAR, RADAR, and camera-based systems. Integrating these inputs with robust AI algorithms enables accurate navigation and environmental perception. Innovations in sensor integration and AI-driven processing enhance safety and reliability in autonomous vehicles \cite{wang2024digital}.
					\end{enumerate}
					
					\section{Conclusion and {Future Research Plans}}\label{sec:ConclusionFuturework}
					
					HSI has emerged as a transformative technology, offering unparalleled spatial and spectral insights across a broad range of applications. The advancements in spectral resolution, sensor miniaturization, and computational methodologies have significantly expanded its utility in domains such as healthcare, agriculture, defense, and industrial automation. The advent in the field of AI has augmented HSI and increased its precision and accuracy in all these areas. This study has provided a comprehensive exploration of HSI, covering its fundamental principles, image processing techniques, state-of-the-art sensors, datasets, and AI-driven analytical tools, applications, and the use of deep learning in HSI, and its growth rate. One of the key innovations highlighted in this work is the integration of hyperspectral cameras with LLMs, forming the "high-brain LLM" framework. This synergy enhances real-time decision-making for critical applications like low-visibility crash detection and face anti-spoofing, demonstrating the potential of AI-powered HSI in generating actionable insights. Furthermore, the study underscores the rapid growth of the HSI industry, as reflected in its increasing market adoption and investment trends. Despite its progress, HSI still faces challenges in data fusion, quality assessment, and spectral unmixing, necessitating further research into deep learning-enhanced methodologies. The integration of multimodal HSI with AI-driven analytics offers exciting prospects for future advancements, promising more accurate and efficient hyperspectral data interpretation. By addressing these challenges and exploring emerging trends, this work serves as a valuable resource for researchers and industry professionals, paving the way for the continued evolution and adoption of HSI technologies.\par
					
\begin{figure}[!htbp]
						\centering
						\includegraphics[width=\linewidth, keepaspectratio, trim=1.5cm 1.25cm 1.5cm 1.5cm, clip]{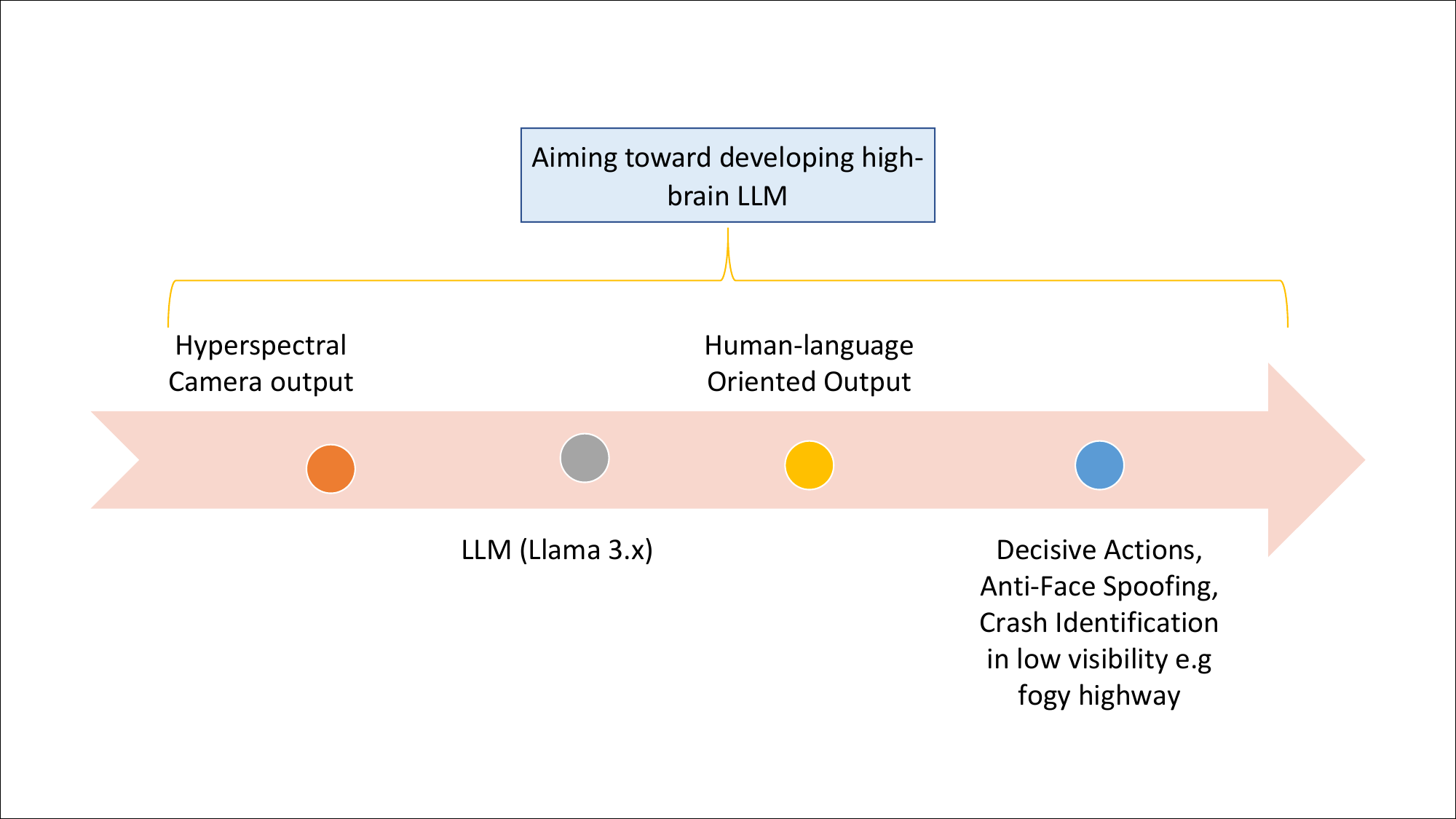}
						\caption{Methodology for High-brain LLM}
						\label{fig:highbrainLLM}
					\end{figure}

					The integration of HSI with AI, particularly LLMs, presents a transformative direction for ongoing research. This synergy enables AI-powered LLMs to perceive beyond human vision, allowing them to make informed decisions in complex real-world scenarios, such as face anti-spoofing and low-visibility crash identification (e.g., fog, smog). This concept elevates traditional LLMs into "high-brain LLMs," enhancing their ability to interpret hyperspectral data and generate actionable insights, as illustrated in Figure \ref{fig:highbrainLLM}.
					One promising application is in face anti-spoofing, where HSI can capture spectral signatures that differentiate between real and fake faces. By processing these features, LLMs can provide human-readable alerts, improving authentication systems and preventing unauthorized access attempts. Similarly, in low-visibility crash detection, hyperspectral imaging can identify obstacles, such as crashed or rolled-over vehicles, that may be obscured by adverse weather conditions. LLMs can then generate context-aware alerts, assisting autonomous and human-driven vehicles in making safer navigation decisions.
					Future research will focus on optimizing these frameworks by improving the efficiency of hyperspectral data processing, enhancing the real-time inference capabilities of LLMs, and developing lightweight models suitable for deployment in edge devices. Additionally, exploring the fusion of HSI with other sensing modalities, such as LiDAR and thermal imaging, could further strengthen these applications.
					\section*{Acknowledgments}
					This work was supported by Institute of Information \& communications Technology Planning \& Evaluation (IITP) grant funded by the Korea government(MSIT) (IITP-2025-RS-2021-II210118, Development of decentralized consensus composition technology for large-scale nodes) and This work was supported by the IITP(Institute of Information \& Communications Technology Planning \& Evaluation)-ITRC(Information Technology Research Center) grant funded by the Korea government(Ministry of Science and ICT)(IITP-2025-RS-2021-II211835).
					
					\section*{Abbreviations}\label{sec:abbreviations}
										
					\begin{table}[htbp]
						\scriptsize
						\setlength{\tabcolsep}{0.005\textwidth}  
						\begin{tabular}{p{4.25cm}p{4.25cm}}
							HSI:Hyperspectral Imaging & MRA:Multiresolution Analysis \\
							LLMS:Large Language Models & RADAR:Radio Detection and Ranging \\
							LiDAR:Light Detection and Ranging & RMSE:Root Mean Square Error \\
							SAR:Synthetic Aperture Radar & SAM:Spectral Angle Mapper \\
							CNN:Convolutional Neural Network & ICA:Independent Component Analysis \\
							LSTM:Long Short-Term Memory & YOLO:You Only Look Once \\
							GRU:Gated Recurrent Unit & ViTs:Vision Transformers \\
							RNN:Recurrent Neural Network & NIR:Near-Infrared \\
							GANs:Generative Adversarial Networks & CASI:Compact Airborne Spectral Imager \\
							DBNs:Deep Belief Networks & SWIR:Shortwave Infrared \\
							MIR:Mid-Infrared \newline MWIR:Midwave Infrared & AISA:Airborne Imaging Spectrometer for Applications \\
							MCR:Multivariate Curve Resolution & PCA:Principal Component Analysis \\
							AVIRIS:Airborne Visible/Infrared Imaging Spectrometer & NASA:National Aeronautics and Space Administration \\
							UAV:Unmanned Aerial Vehicle & DNNs:Deep Neural Networks \\
							PPI:Pixel Purity Index & CAGR:Compound Annual Growth Rate \\
							RGB:Red, Green, Blue \newline DL:Deep Learning & VNIR:Visible Near Infrared Reflectance Spectroscopy \\
							\multicolumn{2}{l}{ESA’s PROBA-CHRIS:European Space Agency’s PRoject for Atmospheric } \\
							\multicolumn{2}{l}{CHaracterization and Remote Imaging Spectrometer} \\
						\end{tabular}
					\end{table}
	
	
					\bibliography{References}
					\bibliographystyle{ieeetr}
					
				\end{document}